%% file: main.tex
\documentclass[]{academic_template}

\usepackage[toc,page,header]{appendix}

\usepackage{amssymb}
\usepackage{amsmath}

\usepackage{makecell}
\usepackage{algorithm}      
\usepackage{algpseudocode}  

\usepackage{graphicx}

\usepackage{listings}
\usepackage{url}
\usepackage{booktabs}
\usepackage{wrapfig}
\usepackage{verbatim}

\usepackage{colortbl}
\usepackage{xcolor}
\definecolor{PromptTitle}{HTML}{F2EEE9} 
\definecolor{PromptFrame}{HTML}{7A6A58}
\definecolor{PromptBack}{HTML}{FAF9F7}
\usepackage[most]{tcolorbox}
\tcbuselibrary{breakable}
\usepackage{fvextra}

\usepackage{caption}
\usepackage{subcaption}

\setlogoheight{14mm}   
\setlogospacing{5mm}
\setsjtublue 

\settitlerulethickness{3pt}  

\abstractboxon 
\setabstractframecolor{gray} 
\setabstractbgcolor{gray!10} 
\setlogotolineshift{5mm} 

\settoprulethickness{2.5pt}  
\setbottomrulethickness{1.5pt}

\title{RISE-Video: Can Video Generators Decode Implicit World Rules?}

\author[1,2,*]{Mingxin Liu}
\author[3,2,*]{Shuran Ma}
\author[4,*]{Shibei Meng}
\author[1,*\ddagger]{Xiangyu Zhao}
\author[1]{Zicheng Zhang}
\author[1]{Shaofeng Zhang}
\author[1]{Zhihang Zhong}
\author[2]{Peixian Chen}
\author[2]{Haoyu Cao}
\author[2]{Xing Sun}
\author[5]{Haodong Duan}
\author[1, \dagger]{Xue Yang}

\affiliation[1]{Shanghai Jiao Tong University}
\affiliation[2]{Tencent Youtu Lab}
\affiliation[3]{Xidian University}
\affiliation[4]{Beijing Normal University}
\affiliation[5]{The Chinese University of Hong Kong}

\contribution[*]{Equal contribution}
\contribution[\ddagger]{Project Lead}
\contribution[\dagger]{Corresponding Author}

\abstract{
 While generative video models have achieved remarkable visual fidelity, their capacity to internalize and reason over implicit world rules remains a critical yet under-explored frontier. 
 To bridge this gap, we present RISE-Video, a pioneering reasoning-oriented benchmark for Text-Image-to-Video (TI2V) synthesis that shifts the evaluative focus from surface-level aesthetics to deep cognitive reasoning. 
 RISE-Video comprises 467 meticulously human-annotated samples spanning eight rigorous categories, providing a structured testbed for probing model intelligence across diverse dimensions, ranging from commonsense and spatial dynamics to specialized subject domains.
 Our framework introduces a multi-dimensional evaluation protocol consisting of four metrics: \textit{Reasoning Alignment}, \textit{Temporal Consistency}, \textit{Physical Rationality}, and \textit{Visual Quality}.
 To further support scalable evaluation, we propose an automated pipeline leveraging Large Multimodal Models (LMMs) to emulate human-centric assessment. 
 Extensive experiments on 11 state-of-the-art TI2V models reveal pervasive deficiencies in simulating complex scenarios under implicit constraints, offering critical insights for the advancement of future world-simulating generative models.
}

\date{\today}

\checkdata[Code]{\url{https://github.com/VisionXLab/RISE-Video}}
\checkdata[Hugging Face]{\url{https://huggingface.co/datasets/VisionXLab/RISE-Video}}
\renewcommand{\topfraction}{0.9}      
\renewcommand{\bottomfraction}{0.2}   
\renewcommand{\textfraction}{0.0}    
\renewcommand{\floatpagefraction}{0.85} 
\begin{document}
\maketitle


\begin{figure*}[htbp]
    \centering
    \includegraphics[width=0.999\linewidth]{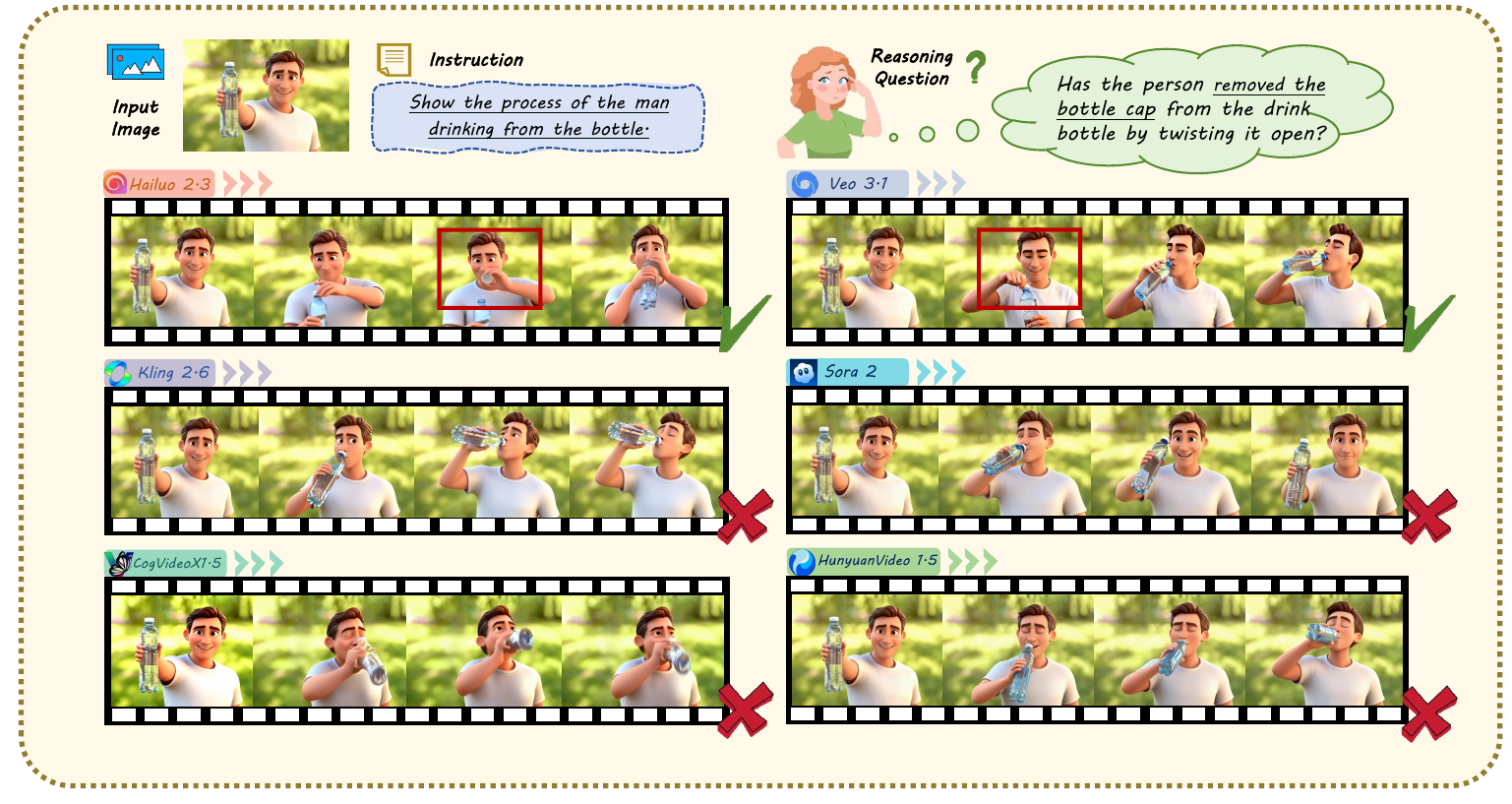}
    \caption{An example from the \textit{Experiential Knowledge} dimension of RISE-Video, revealing limitations in experience-based reasoning of current TI2V models.}
    \label{case}
\end{figure*}

\input{sections/1_introduction}

\input{sections/2_related_work}
\input{sections/3_method}
\input{sections/4_experiments}
\input{sections/5_conclusion}


\bibliographystyle{plainnat}
\bibliography{references}

\clearpage
\beginappendix
\renewcommand{\textfraction}{0.01}      
\renewcommand{\floatpagefraction}{0.6}  
\renewcommand{\topfraction}{0.99}       
\renewcommand{\bottomfraction}{0.99}    
\input{sections/6_appendix}

\end{document}

%% file: sections/1_introduction.tex
\section{Introduction}
\label{sec:intro}

Recent years have witnessed rapid progress in video generation, driven largely by advances in large-scale generative models. 

Increasingly realistic Text-to-Video (T2V)~\cite{zhang2023show1,singer2022make,yang2024cogvideox,wang2023modelscope} and Text-Image-to-Video (TI2V) generation~\cite{ni2024ti2v,huang2025step} models have demonstrated remarkable success in enhancing visual fidelity and structural controllability. Despite these strides, a critical question remains unanswered: can contemporary TI2V models reliably internalize and reason over implicit world rules that extend beyond explicit textual instructions?
While general-purpose frameworks like VBench~\cite{huang2024vbench} provide comprehensive evaluations, and various task-oriented benchmarks~\cite{meng2024towards,yuan2024chronomagic} have emerged, most existing metrics predominantly emphasize perceptual quality and temporal coherence. Consequently, there is a notable scarcity of evaluation protocols focused on implicit reasoning, particularly within the TI2V paradigm. These exigencies necessitate the development of a dedicated, reasoning-oriented diagnostic framework.

To bridge the gap in rule-aware evaluation for TI2V models, we introduce RISE-Video, a benchmark explicitly engineered to prioritize implicit reasoning over superficial generative quality. 
At the foundational level, the benchmark is organized into eight distinct reasoning dimensions: experiential, commonsense, temporal, societal, perceptual, spatial, subject-specific, and logical reasoning. This taxonomy enables a comprehensive coverage of the reasoning landscape in video synthesis, spanning from low-level perceptual cues to high-level abstract inferences. 
RISE-Video comprises 467 meticulously curated samples, each subject to rigorous human expert annotation to ensure ground-truth reliability. Building upon this structured data foundation, we define four evaluation metrics to provide a holistic appraisal: \textit{Reasoning Alignment}, \textit{Temporal Consistency}, \textit{Physical Rationality}, and \textit{Visual Quality}. This multi-dimensional approach ensures that generated videos are not only visually plausible but also strictly adhere to the underlying cognitive and physical constraints mandated by the input instructions.

To enable scalable evaluation, we further develop an automated LMM (Large Multimodal Models)-based judging pipeline guided by manually designed, reasoning-aware questions and prompts. Using this pipeline, we evaluate 11 representative TI2V models, and the results reveal clear reasoning limitations across current systems. Moreover, we validate that the proposed evaluation pipeline exhibits a high degree of alignment with human judgments, indicating that LMM-based evaluation can serve as a reliable and cost-effective alternative to large-scale human assessment.

Overall, our contributions are as follows:

1. We introduce RISE-Video, a pioneering benchmark designed to evaluate the capacity of TI2V models to internalize and execute implicit world rules. It encompasses 467 meticulously human-annotated samples across eight distinct reasoning domains, providing comprehensive coverage of diverse scenarios.

2. We propose four complementary evaluation dimensions to assess reasoning correctness beyond perceptual fidelity and develop an automated LMM-based evaluation pipeline, enabling scalable evaluation while maintaining strong alignment with human judgments.

3. We conduct a comprehensive evaluation on 11 representative TI2V models, revealing systematic reasoning limitations and providing insights into current model capabilities.

%% file: sections/2_related_work.tex
\section{Related Work}
\label{sec:related}

\subsection{Video Generation Models}
Video generation~\cite{zhang2023show1, chen2024videocrafter2overcomingdatalimitations, runway2025gen4,klingteam2025klingomnitechnicalreport} research has advanced primarily through diffusion models~\cite{ho2020denoising,ding2021cogview,esser2024scaling,ho2022imagen,song2020denoising,dhariwal2021diffusion}. Early works~\cite{guo2023animatediff,blattmann2023stable} integrated motion priors into image generators by adding temporal modules to latent diffusion models, enabling text-to-video and image-to-video synthesis. Beyond training recipes, architectural advances further strengthen long-range temporal coherence; Lumiere~\cite{bar2024lumiere} adopts a space–time U-Net to generate entire clips in a single pass, while CogVideoX~\cite{yang2025cogvideoxtexttovideodiffusionmodels} scales diffusion-transformer designs with a 3D VAE to support longer, higher-resolution, and better text-aligned videos. In parallel, a complementary line explores unified multimodal generation and editing instead of single-condition synthesis: VideoPoet~\cite{kondratyuk2023videopoet} reformulates video generation as autoregressive multimodal token prediction, and Movie Gen~\cite{polyak2024movie} extends this paradigm toward high-resolution generation with instruction-based editing and audio alignment. At the frontier, large-scale closed and production systems~\cite{peng2025opensora20trainingcommerciallevel, wiedemer2025video,klingteam2025klingomnitechnicalreport, runway2025gen4} further push video duration, realism, and controllability, highlighting the need for systematic evaluation across text-to-video and image-to-video settings.

\begin{figure*}[t]
    \centering
    \includegraphics[width=\textwidth]{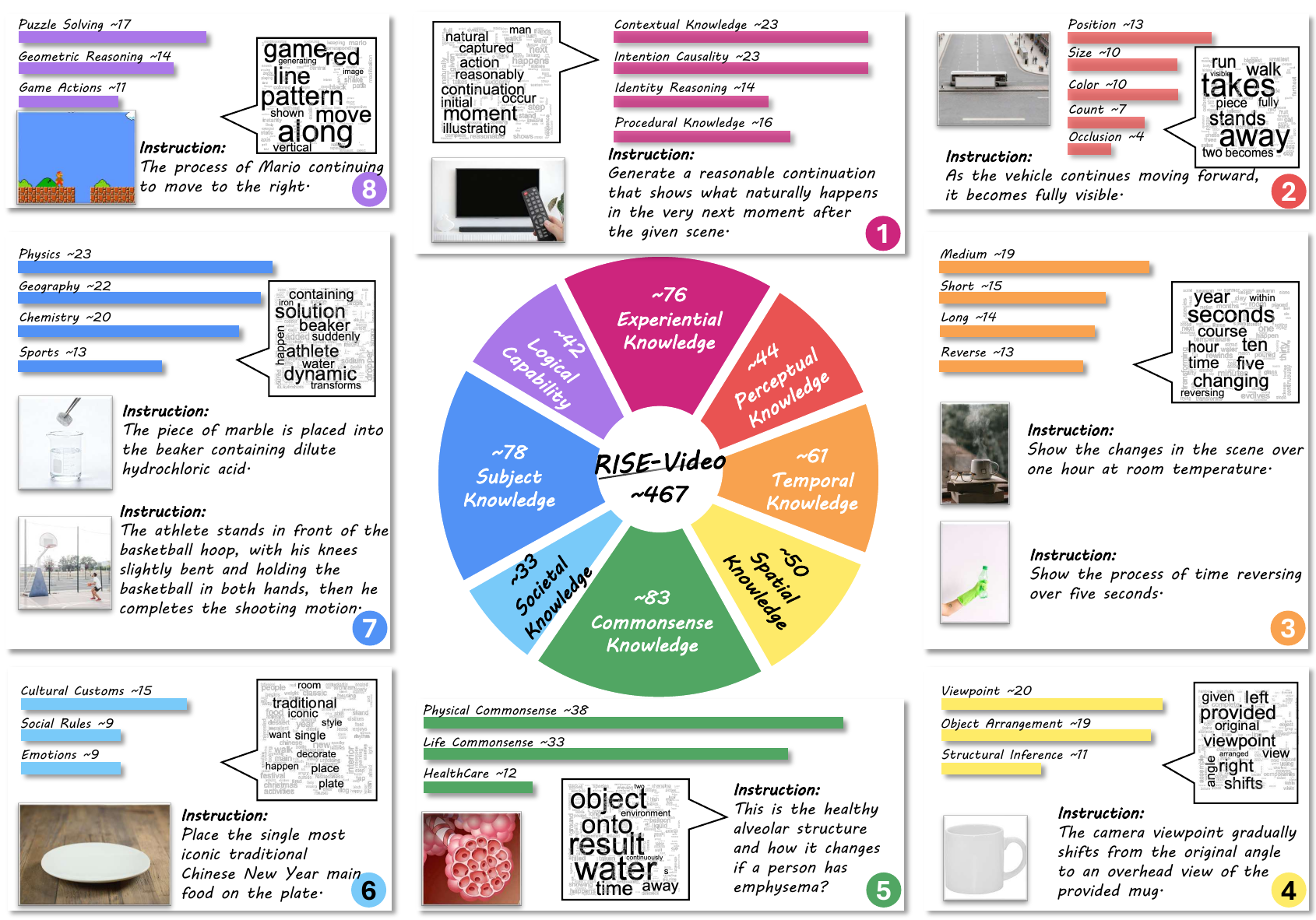}
    \caption{Task distribution of the RISE-Video benchmark, which comprises eight major task categories: \textit{Experiential Knowledge}, \textit{Perceptual Knowledge}, \textit{Temporal Knowledge}, \textit{Spatial Knowledge}, \textit{Commonsense Knowledge}, \textit{Societal Knowledge}, \textit{Subject Knowledge}, and \textit{Logical Capability}. Each category further contains comprehensive sub-categories and diverse data samples.}
    \label{fig:data_dist}
\end{figure*}

\subsection{Evaluation of Video Generation Models}
Video generation benchmarks have progressively evolved from coarse perceptual metrics toward more structured and semantically grounded evaluation protocols. Early evaluations~\cite{heusel2017gans, salimans2016improved, unterthiner2019fvd} rely on frame- or video-level metrics, which measure overall realism but fail to capture motion coherence. To go beyond coarse metrics and capture the diverse capabilities of modern video generation, VBench~\cite{huang2024vbench} provides a unified framework with 8 data categories and 16 evaluation dimensions. In parallel, specific research ~\cite{meng2024towards, sanli2025can,bansal2025videophy,motamed2025generative}focuses on evaluating whether generated content adheres to basic physical commonsense, and additional benchmarks~\cite{liao2024evaluation, yuan2024chronomagic, feng2024tc, ji2024t2vbench} further emphasize video dynamics. Recent benchmarks~\cite{zheng2025vbench, sun2025t2v,wu2024bettermetrictexttovideogeneration,ning2025video,zhang2025ui2v} move toward comprehensive, human-aligned evaluation, increasingly using LMM as judger. However, existing benchmarks for video generation mainly assess perceptual quality and temporal coherence, yet fall short in evaluating higher-level reasoning abilities.

%% file: sections/3_method.tex
\section{Method}
\label{sec:method}

\subsection{Data Construction}
With a primary focus on reasoning capabilities, we partition the dataset based on the types of reasoning knowledge involved. Specifically, as shown in Fig.~\ref{fig:data_dist}, we define eight categories of reasoning knowledge, each category targeting a specific aspect of reasoning required for understanding or generating videos under structured constraints. 

\textbf{Commonsense Knowledge.} This dimension evaluates whether video generation models encode and apply commonsense knowledge about everyday physics and human life, comprising three sub-aspects: \textit{1) Physical commonsense} assesses understanding of basic cause–effect relations, such as footprints left on snow or a vase breaking when hit; \textit{2) Life commonsense} examines knowledge of everyday biological responses, including skin swelling after a mosquito bite; \textit{3) Healthcare commonsense} evaluates familiarity with basic health practices, such as dental decay formation and mouthwash use. These sub-aspects together measure the model’s ability to reflect widely shared commonsense knowledge during video generation.

\textbf{Subject Knowledge.} Subject knowledge refers to structured, discipline-specific knowledge that extends beyond everyday experience and general common sense. It is organized into four sub-domains: \textit{1) Physics} evaluates understanding of fundamental physical principles across multiple subfields, including electricity, mechanics, and optics; \textit{2) Chemistry} examines knowledge of common chemical phenomena and reactions; \textit{3) Geography} involves a diverse range of topics such as celestial systems, river formations, and weather-related processes; \textit{4) Sports} focuses on generating subject-specific movements like soccer shooting, volleyball bump, and the iron cross in gymnastics. 

\textbf{Perceptual Knowledge.} Accurate perception of basic visual attributes is a prerequisite for complex video generation. This dimension thus evaluates models' capacity to capture and manipulate low-level perceptual semantics, including \textit{1) size}, \textit{2) color}, \textit{3) count}, and \textit{4) position}. In addition to these core perceptual attributes, we further introduce a more challenging sub-aspect, \textit{5)occlusion}, which assesses whether models can correctly infer and reconstruct objects that are partially occluded in the scene. These aspects probe the robustness of perceptual grounding required for reliable video generation.

\textbf{Societal Knowledge.}To generate videos that adhere to real-world social norms, we propose the societal knowledge dimension for assessing models' understanding of social and cultural contexts. This dimension consists of three sub-aspects: \textit{1) Emotion recognition} infers emotional states from visual cues such as facial expressions; \textit{2) Social rules} captures commonly accepted behavioral norms, like disposing of trash properly or stopping at red lights; \textit{3) Cultural customs} reflects practices rooted in different societies, including dietary traditions and festival-related activities.

\textbf{Logical Capability.} Logical capability requires models to apply explicit rules systematically over visual elements, representing a challenging integrative aspect of reasoning due to the demand for structured, constraint-based inference.  We divide this dimension into three sub-aspects: \textit{1) Game actions} evaluate whether models can follow the rules of classic game scenarios, such as Super Mario, and generate valid actions; \textit{2) Puzzle solving} focuses on logic-driven scenarios such as mazes, board games (e.g., Gomoku), and word-linking puzzles, where correct generation depends on satisfying well-defined logical constraints; \textit{3) Geometric Reasoning} evaluates whether models can systematically reason under geometric rules and generate outputs that strictly adhere to the given structural constraints.

\textbf{Experiential Knowledge.} This dimension evaluates whether video generation models capture human-like experience-based knowledge for interpreting intentions, identities, procedures, and context. It comprises four sub-aspects: \textit{1) Intention causality} – inferring goals from intention cues (e.g., spoon near mouth implies eating); \textit{2) Identity reasoning} – identifying and tracking a specified individual among multiple entities; \textit{3) Procedural knowledge} – understanding correct action sequences (e.g., peeling before eating an orange); \textit{4) Contextual knowledge} – applying experiential knowledge based on textual scenario descriptions during generation.

\textbf{Spatial Knowledge.} This dimension evaluates a model’s ability to understand spatial relationships and to manipulate objects within a three-dimensional environment. Inspired by RISEBench~\cite{zhao2025envisioning}, we decompose spatial knowledge into three aspects:\textit{ 1) Viewpoint} assesses whether models can perform viewpoint transformations by following a specified camera trajectory, as camera positioning and motion are critical factors in video generation; \textit{2) Object arrangement} examines whether multiple objects can be organized according to spatial attributes like relative size and shape; \textit{3) Structural inference} tests the capacity to integrate incomplete components into a spatially consistent structure.

\textbf{Temporal Knowledge.} This dimension evaluates temporal reasoning in video generation across different time spans and ordering patterns. We categorize temporal knowledge into four types: \textit{1) short-term} temporal reasoning, which involves events occurring within a few seconds, such as changes in traffic signal states within a five-second interval; \textit{2) medium-term} temporal reasoning, covering durations from minutes to several months; \textit{3) long-term} temporal reasoning, which requires understanding changes over periods exceeding one year; and \textit{4) reverse temporal} reasoning, where events are presented in reverse order to increase task difficulty, for example, an adult elephant gradually transforming as time rewinds over ten years. 

Following the established categories, RISE-Video comprises 467 samples, each meticulously curated and annotated by human experts to ensure a diverse and representative coverage of reasoning scenarios.

   \begin{figure} 
    \begin{minipage}[t]{0.5\textwidth}
        \centering
        \includegraphics[width=\columnwidth]{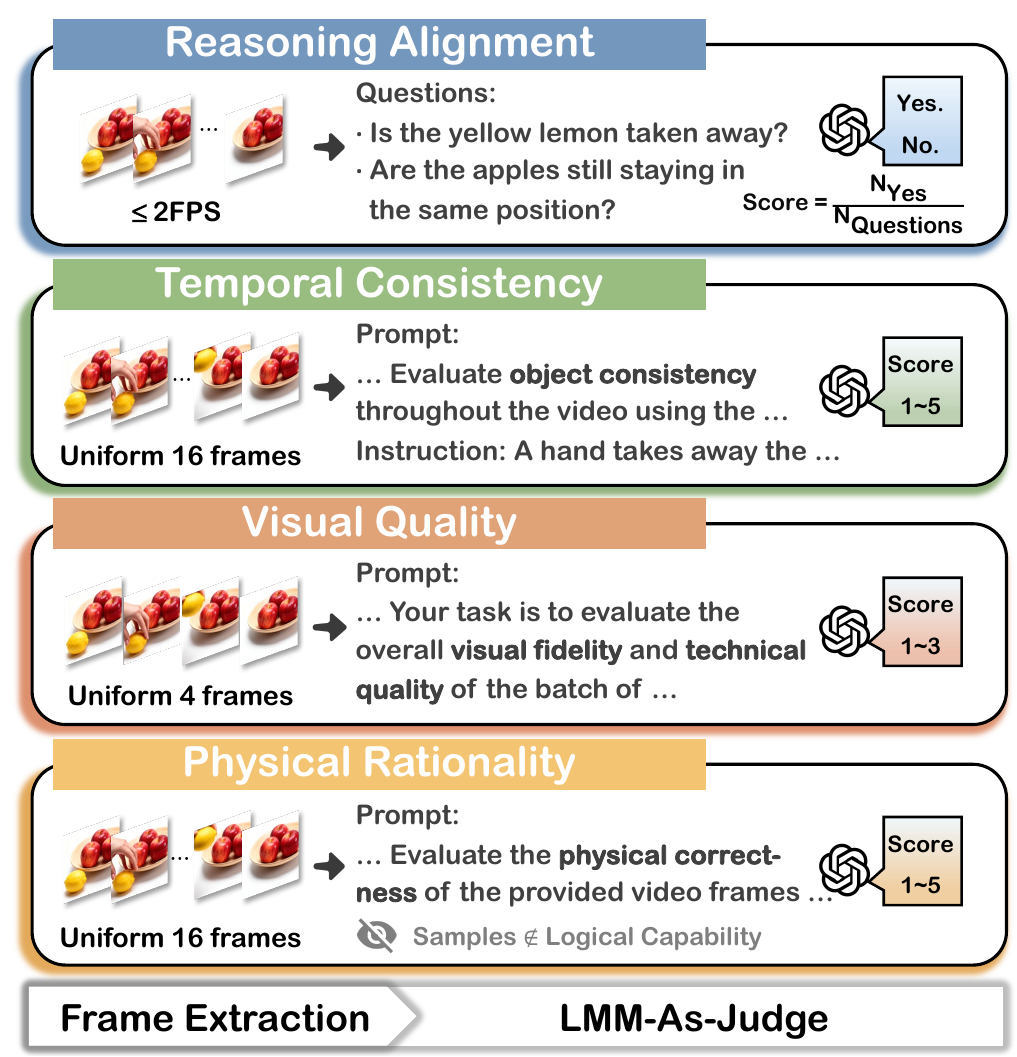} 
        \caption{Evaluation pipeline of the RISE-Video benchmark. It covers four metrics: \textit{Reasoning Alignment}, \textit{Temporal Consistency}, \textit{Visual Quality}, and \textit{Physical Rationality}, with dimension-specific frame extraction strategies. Carefully designed prompts guide GPT-5 as the primary judge (GPT-5-mini for Visual Quality only), ensuring fair and objective evaluation.}
        \label{fig:pipeline}
    \end{minipage}
    \quad
    \begin{minipage}[t]{0.47\textwidth}
        \centering
        \includegraphics[width=\columnwidth]{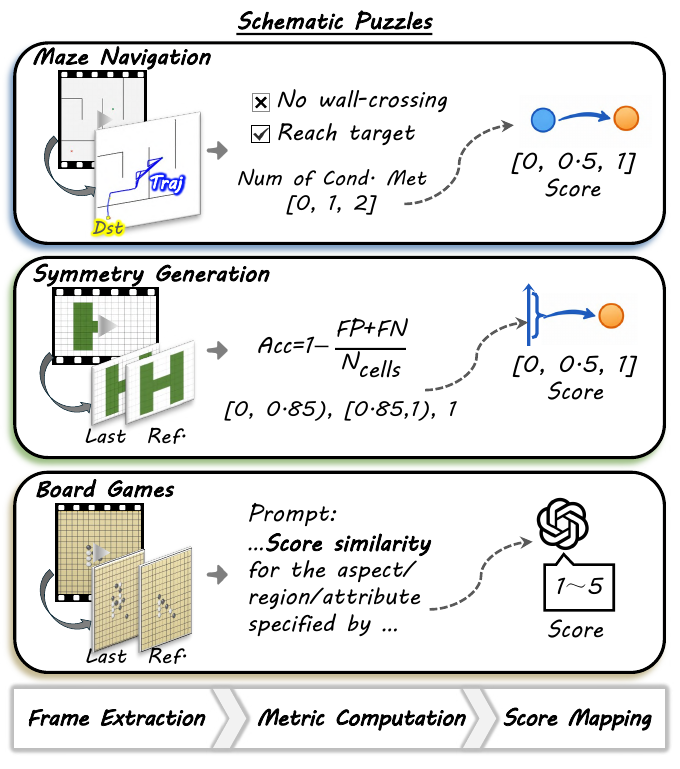} 
        \caption{Specialized evaluation pipeline for reasoning alignment in \textit{Schematic Puzzle} tasks, which are not well-suited for standard LMM-as-a-Judge evaluation, including trajectory-based constraint checking, grid-level structural alignment, and reference-assisted LMM comparison, enabling accurate and interpretable scoring of structured visual reasoning outcomes.}
        \label{fig:logic-pipeline}
        \vspace{-0.5em}
    \end{minipage}
    \end{figure}

\subsection{Evaluation Metrics}
Evaluation metrics fundamentally determine both the capabilities assessed by a benchmark and the interpretation of model performance. As shown in Fig. \ref{fig:pipeline}, we evaluate model performance along four complementary dimensions, as demonstrated below:

\textbf{\textit{Reasoning Alignment.}} This metric assesses whether the generated video demonstrates correct knowledge-based reasoning by evaluating the accuracy of inferred relationships, changes, and outcomes. To improve the accuracy and specificity of LMM-based evaluation under this metric, we adopt a targeted assessment strategy in which, for each sample, a set of manually designed, knowledge-aware questions is constructed according to the reasoning type being evaluated. The questions are answered by the LMM judge in a binary (Yes/No) manner, based on which each sample receives a 0–1 score for Reasoning Alignment. We further employ reasoning-aware frame sampling strategies to better support judgment under different temporal requirements: samples that require evaluating the full progression of an event are uniformly sampled at 2 fps, while scenarios that primarily focus on the final state (e.g., assessing whether a kitten becomes an adult cat after one year) adopt a lower sampling rate to emphasize terminal outcomes. This approach reduces redundant visual input and evaluation cost while preserving the information required for reliable judgment.

Within Logical Capability, we categorize tasks with abstract visual primitives including \textit{maze navigation}, \textit{symmetry generation}, and \textit{board games} as \textbf{Schematic Puzzles}. These are ill-suited for standard LMM-as-a-Judge due to their rigid geometry and the difficulty of describing ground-truth (GT) states linguistically. We implement specialized strategies as shown in Fig.~\ref{fig:logic-pipeline}. For \textit{Maze Navigation}, we bypass linguistic judging by tracking agent trajectories across all frames via color matching to verify two constraints: (1) no wall-crossing and (2) reach the target. The count of satisfied constraints $\{0, 1, 2\}$ maps to scores $\{0, 0.5, 1\}$. For \textit{Symmetry Generation}, to decouple Reasoning Alignment from Temporal Consistency, correctness is assessed via grid-level positional alignment between the last frame and the GT reference, disregarding specific color matches. Cells identified via HSV are used to determine False Positives (FP, misplaced) and False Negatives (FN, missing). Accuracy is then calculated as $1 - (FP + FN) / N$, where $N$ denotes the total cell count in the grid. This value is subsequently discretized into scores $\{0, 0.5, 1\}$ based on the intervals $[0, 0.85)$, $[0.85, 1)$, and $\{1\}$; the 0.85 threshold reflects the human-perceptual boundary for structural trends. For \textit{Board Games}, where rules are difficult to articulate linguistically, we provide the LMM judge with the last frame alongside an auxiliary GT reference image. This dual-input approach provides essential visual grounding, enabling the judge to perform a precise structural comparison between the generated output and the target state.

\input{sections/tabs/main_1}

\textbf{\textit{Temporal Consistency.}} 
Temporal Consistency evaluates whether the generated video exhibits only the changes explicitly or implicitly required by the instruction, while preserving all other aspects that are irrelevant to the instruction, such as object attributes, scene layout, and character identity. This metric emphasizes isolating instruction-induced changes from unintended variations. In practice, the instruction is provided to the LMM judge to explicitly identify and exclude the changes specified by the instruction, and to assess the consistency of all remaining elements in the generated video. To support this assessment, we apply uniform frame sampling to provide a representative and temporally distributed view of the content, balancing temporal coverage and evaluation efficiency. The judgment is reported on a 1–5 scale, reflecting the degree to which non-instructed components remain stable throughout the video.

\textbf{\textit{Physical Rationality.}}
Physical Rationality evaluates whether the generated video adheres to fundamental laws of physics and real-world logic, encompassing aspects such as gravity, object permanence, collision dynamics, and fluid motion. This metric emphasizes the plausibility of dynamic interactions and adherence to physical laws, ensuring objects maintain structural integrity and interact naturally. Besides, this metric is applicable strictly to physically grounded environments. Abstract tasks of planar logical puzzles or symbolic reasoning, which do not rely on real-world physical constraints, are excluded from this assessment. In practice, the LMM judge is instructed to verify the physical accuracy and confirm the logical coherence of movements and environmental reactions within the scene. To support the assessment, we apply uniform frame sampling to provide a representative and temporally distributed view of the motion dynamics, balancing temporal coverage and evaluation efficiency. The judgment is reported on a 1–5 scale, reflecting the degree to which the video maintains physical realism and temporal coherence without distortions.

\textbf{\textit{Visual Quality.}}
Visual Quality evaluates the perceptual fidelity and technical integrity of the generated video, focusing on subject sharpness, texture preservation, and lighting consistency. Notably, we manually apply super-resolution to low-clarity images prior to evaluation. This operation prevents the LMM judge from misinterpreting low native resolution as technical blur, allowing for a fairer assessment of the actual generative artifacts. In practice, the LMM judge assesses a batch of sampled frames to verify that the main subject remains crisp and structurally coherent throughout the sequence. To support this assessment, we uniformly sample 6 frames from the entire video and exclude the first and last frames to mitigate boundary instability. The judgment is reported on a 1–3 scale, classifying results from severe technical failure to professional-standard clarity.

Building on the above metrics, we introduce two types of overall scores to aggregate evaluation results. The first is \textbf{Weighted Score}, computed by assigning weights of 0.4, 0.25, 0.25, and 0.1 to Reasoning Alignment, Temporal Consistency, Physical Rationality, and Visual Quality, respectively. The second is \textbf{Accuracy}, where a case is counted as correct only if all four dimensions achieve full scores, and the resulting accuracy is normalized to a 100-point scale.

%% file: sections/tabs/main_1.tex
\begin{table*}[htbp]
\centering
\caption{Performance comparison of different models across four evaluation metrics and overall scores. Metrics include RA (Reasoning Alignment), TC (Temporal Consistency), PR (Physical Rationality), and VQ (Visual Quality). Overall performance is reported using W.Score (Weighted Score) and Accuracy.
}
\begin{tabular}{l c c c c c c}
\toprule
\multirow{2}{*}{\textbf{Models}} & \multirow{2}{*}{\textbf{RA}} & \multirow{2}{*}{\textbf{TC}} & \multirow{2}{*}{\textbf{PR}} & \multirow{2}{*}{\textbf{VQ}} & \multicolumn{2}{c}{\textbf{Overall}} \\
\cmidrule(lr){6-7}
& & & & & W.Score & Accuracy\\
\hline
\rowcolor{gray!20}
\multicolumn{7}{l}{$\blacktriangledown$ \textit{Closed-source Model}} \\
Hailuo2.3~\cite{hailuo2.3} & \textbf{76.6\%} & 87.2\% & 71.0\% & 92.0\% & \textbf{79.4\%} & \textbf{22.5\%}\\
Veo3.1~\cite{wiedemer2025video} & 64.9\% & 86.0\% & \textbf{78.9\%} & 91.9\% & 76.4\% & 22.3\%\\
Sora-2~\cite{peng2025opensora20trainingcommerciallevel} & 64.0\% & \textbf{92.2\%} & 76.3\% & 92.2\% & 77.0\% & 21.3\%\\
Wan2.6~\cite{wan2025wanopenadvancedlargescale} & 70.0\% & 88.8\% & 72.5\% & 94.5\% & 77.8\% & 21.3\%\\
Kling2.6~\cite{klingteam2025klingomnitechnicalreport} & 53.7\% & 86.4\% & 78.0\% & 95.1\% & 72.1\% & 19.5\%\\
Seedance1.5-pro~\cite{seedance2025seedance15pronative} & 61.2\% & 81.1\% & 70.7\% & \textbf{96.2\%} & 72.0\% & 17.6\%\\
\hline
\rowcolor{gray!20}
\multicolumn{7}{l}{$\blacktriangledown$ \textit{Open-source Model}} \\
Wan2.2-I2V-A14B~\cite{wan2025wanopenadvancedlargescale} & 39.5\% & 79.2\% & 75.4\% & 94.0\% & 63.9\% & 11.4\%\\
HunyuanVideo-1.5-720P-I2V~\cite{hunyuanvideo2025} & 38.1\% & 75.0\% & 68.4\% & 92.6\% & 60.4\% & 8.6\%\\
HunyuanVideo-1.5-720P-I2V-cfg-distill & 38.9\% & 74.0\% & 65.8\% & 92.9\% & 59.9\% & 7.3\%\\
Wan2.2-TI2V-5B ~\cite{wan2025wanopenadvancedlargescale} & 32.6\% & 70.5\% & 72.8\% & 89.7\% & 57.8\% & 5.4\%\\
CogVideoX1.5-5B ~\cite{yang2024cogvideox} & 30.7\% & 62.3\% & 56.7\% & 74.5\% & 49.5\% & 1.9\%\\
\bottomrule
\end{tabular}
\label{tab:model_performance_overall}
\end{table*}

%% file: sections/4_experiments.tex
\section{Experiments}
\label{sec:experiments}

\input{sections/tabs/main_2}
To evaluate the reasoning capabilities of current TI2V models, we conduct experiments on 11 representative models, covering both closed-source and open-source systems. The closed-source models include 
Hailuo 2.3~\cite{hailuo2.3}, Wan2.6~\cite{wan2025wanopenadvancedlargescale}, Sora 2~\cite{peng2025opensora20trainingcommerciallevel}, Veo 3.1~\cite{wiedemer2025video}, Kling 2.6~\cite{klingteam2025klingomnitechnicalreport}, and Seedance 1.5-pro~\cite{seedance2025seedance15pronative}, which typically demonstrate strong visual quality and reflect the current upper bound of deployed TI2V performance. In parallel, the open-source models span multiple architectures and training strategies, including Wan2.2-I2V~\cite{wan2025wanopenadvancedlargescale}, HunyuanVideo-1.5-720P-I2V~\cite{hunyuanvideo2025} and its distilled variant, and CogVideoX1.5-5B~\cite{yang2024cogvideox}. For evaluation, GPT-5 is used as the judge for Reasoning Alignment, Temporal Consistency, and Physical Rationality, while GPT-5-mini is used for Visual Quality.

\subsection{Main Results}
Tab.~\ref{tab:model_performance_overall} summarizes the performance of all evaluated models across the four evaluation metrics. 
In general, open-source models consistently underperform closed-source models in both reasoning capability and visual quality. In particular, as shown in Fig.~\ref{case},  models such as CogVideoX1.5 frequently exhibit visual artifacts, including frame-level blurring, ghosting effects, and degraded spatial sharpness, which lead to low Visual Quality scores and hinder reliable reasoning assessment. From the perspective of accuracy, all evaluated models achieve relatively low scores, indicating that reasoning remains a significant challenge for current TI2V systems. Even the best-performing model, Hailuo 2.3, attains an accuracy of only 22.5\%. The second- and third-ranked models are Veo 3.1 and Sora 2, achieving accuracies of 22.3\% and 21.3\%, respectively. This highlights limitations in existing models’ ability to satisfy reasoning-oriented requirements. Among all evaluated models, Hailuo 2.3 demonstrates a particularly notable advantage in Reasoning Alignment, where it exceeds the second-ranked model Wan 2.6 by 6.6\%. Notably, Sora 2 exhibits a clear strength in Temporal Consistency, suggesting that it is more effective at preserving non-instructed elements and maintaining stable generation behavior across videos.

\begin{figure*}[t]
    \centering
    \includegraphics[width=0.99\linewidth]{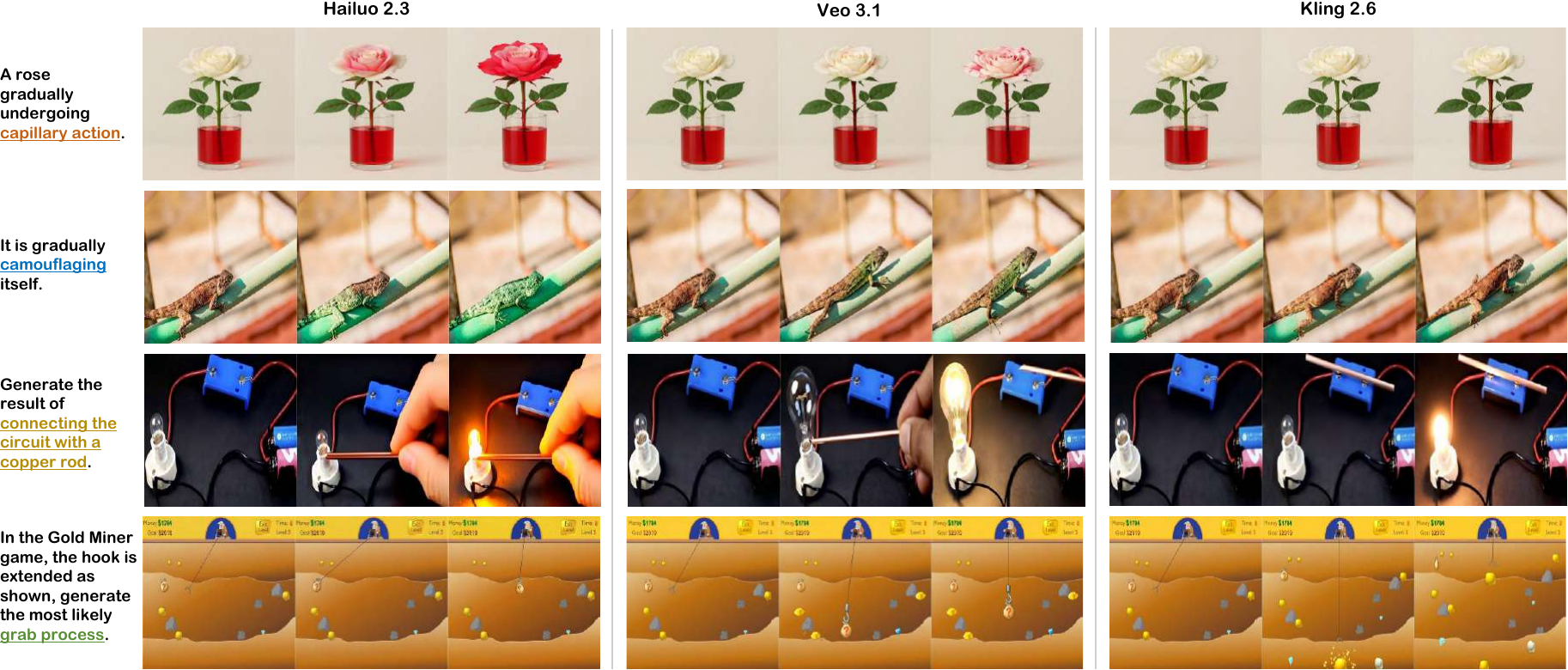}
    \caption{Representative generation results of leading models. We show the examples generated by Hailuo 2.3, Veo 3.1, and Kling 2.6.}
    \label{fig:3cases}
\end{figure*}

Tab.~\ref{tab:model_performance_calss} reports the weighted scores and accuracies of all evaluated models across different reasoning categories. Overall, current TI2V models perform notably better on Perceptual Knowledge than on other reasoning types, indicating that models are relatively strong at perceiving low-level visual attributes such as color, size, and count. In contrast, performance on Logical Capability, which requires the integration of perceptual evidence with abstract reasoning, is consistently low across all models, suggesting that such tasks constitute a major bottleneck for current TI2V systems. As illustrated in Fig.~\ref{fig:3cases}, in the Gold Miner game scenario, where the hook is extended as shown and the model is required to generate the most likely grabbing process, none of the evaluated models successfully capture the stone along the current hook trajectory. Veo 3.1 exhibits consistency issues, with noticeable changes in the hook’s shape, while Kling 2.6 incorrectly moves the object without physical contact between the hook and the gold. This highlights the difficulty of rule-based decision-making in such game-like reasoning settings. Notably, in the Experiential category, Hailuo 2.3 and Veo 3.1 demonstrate clear advantages. As illustrated in Fig.~\ref{case}, when generating a scenario in which a person drinks water from a bottle, only Veo 3.1 and Hailuo 2.3 are able to infer the necessary action of unscrewing the bottle cap, whereas other models fail to exhibit this reasoning behavior. 

In terms of dynamic behavior, several models show limited responsiveness to instructions. For example, Kling 2.6 often produces videos with minimal motion or near-static content. As illustrated in Fig.~\ref{fig:3cases} for the chameleon camouflage and capillary action of a rose tasks, Kling 2.6 tends to preserve the original appearance without performing the required commonsense transformation, resulting in both limited visual dynamics and poor alignment with the underlying reasoning requirement. While Wan 2.6 and Hailuo 2.3 demonstrate stronger instruction following and more dynamic generation behavior, Veo 3.1 and Sora 2 show relatively weaker responsiveness to dynamic instructions. In some cases, these models partially follow the instruction without fully realizing the intended transformation, or produce little effective change. For instance, as shown in Fig.~\ref{fig:3cases}, Veo 3.1 fails to correctly model the camouflage behavior, as the color of the chameleon does not sufficiently adapt to match the surrounding branch. 
In addition, Sora 2 and Veo 3.1 exhibit noticeable temporal discontinuities, characterized by abrupt changes between consecutive frames. Such discontinuities disrupt temporal smoothness and adversely affect overall video quality. 
Additional qualitative visualizations are provided in the supplementary material.

\begin{table*}[htbp]
  \centering
  \caption{Comparison of MAE and STD between LMM-as-Judge and human evaluations across different judge models.}
  \label{tab:judge_model_comparison}
  
  \resizebox{0.9\textwidth}{!}{
    \begin{tabular}{lcccccccc}
      \toprule
      \multirow{2}{*}{\textbf{Judge Model}} & \multicolumn{2}{c}{\textbf{RA}} & \multicolumn{2}{c}{\textbf{TC}} & \multicolumn{2}{c}{\textbf{PR}} & \multicolumn{2}{c}{\textbf{VQ}} \\
      \cmidrule(lr){2-3} \cmidrule(lr){4-5} \cmidrule(lr){6-7} \cmidrule(lr){8-9}
        & \textbf{MAE} ($\downarrow$) & \textbf{STD} ($\downarrow$) & \textbf{MAE} ($\downarrow$) & \textbf{STD} ($\downarrow$) & \textbf{MAE} ($\downarrow$) & \textbf{STD} ($\downarrow$) & \textbf{MAE} ($\downarrow$) & \textbf{STD} ($\downarrow$) \\
      \midrule
      
      Qwen3-VL-235B  & 0.12 & 0.25 & 0.42 & 0.80 & 0.82 & 0.86 & 0.25 & 0.41 \\
      Gemini-3-Flash & 0.13 & 0.26 & 1.08 & 1.15 & 1.52 & 0.96 & 0.84 & 0.63 \\
      
      \textbf{GPT-5$^{\dagger}$(Ours)} & 0.11 & 0.23 & 0.51 & 0.85 & 0.80 & 0.76 & 0.22 & 0.36 \\
      
      \bottomrule
    \end{tabular}
  } 

  \vspace{1mm}
  \begin{minipage}{0.9\textwidth} 
      \footnotesize
      \textbf{Note:} $^{\dagger}$ Specifically for the \textbf{Visual Quality} dimension in the GPT-5 row, we utilize \texttt{gpt-5-mini} to balance cost and performance.
  \end{minipage}
\end{table*}

\subsection{Ablation Study}
In this section, we conduct human evaluations and calculate the Mean Absolute Error (MAE) and Standard Deviation (STD) between automatic metric scores and human ratings to identify the best metric for each category.

We evaluate each dimension on its native scale to preserve the granularity of specific tasks. Specifically, the value ranges are defined as follows: \textit{RA} is scored in $[0, 1]$, \textit{Cons.} and \textit{PR} are assessed on a $[1-5]$ scale, and \textit{VQ} operates on a $[1-3]$ scale. For the human ground truth, we employed five independent expert annotators and calculated the average of their ratings for each sample. Consequently, the MAE is computed as the mean absolute difference between the model's predicted score and this aggregated human consensus, while the STD denotes the standard deviation of these absolute errors.

As illustrated in Tab.~\ref{tab:judge_model_comparison}, GPT-5 demonstrates the most robust alignment with human preference across the majority of metrics. While Qwen3-VL-235B demonstrates a lower error rate on Temporal Consistency, a closer inspection suggests that this model exhibits a tendency towards higher acceptance rates (i.e., looser constraints on consistency). 
This high-score bias is accurate for perfect samples, but it compromises the model's ability to differentiate between truly high-quality outputs and those with severe defects.
We show more cases in Fig~\ref{appd_judge}. Besides, our ablation on the Visual Quality dimension reveals that the cost-effective gpt-5-mini is highly capable of perceptual assessment; it achieves tighter alignment with human ratings than both Gemini-3-Flash and Qwen3-VL-235B. 

%% file: sections/tabs/main_2.tex
\begin{table*}[tbp]
\centering
\caption{Overall performance of different models across eight reasoning categories}
\resizebox{\textwidth}{!}{
\begin{tabular}{l c c | c c | c c | c c | c c | c c | c c | c c}
\toprule

\multirow{2}{*}{\textbf{Models}} & \multicolumn{2}{c}{\textbf{Commonsense}} & \multicolumn{2}{c}{\textbf{Subject}} & \multicolumn{2}{c}{\textbf{Perceptual}} & \multicolumn{2}{c}{\textbf{Societal}} & \multicolumn{2}{c}{\textbf{Logical}} & \multicolumn{2}{c}{\textbf{Experiential}} & \multicolumn{2}{c}{\textbf{Spatial}} & \multicolumn{2}{c}{\textbf{Temporal}}  \\
& WS & Acc. & WS & Acc. & WS & Acc. & WS & Acc. & WS & Acc.& WS & Acc. & WS & Acc. & WS & Acc.  \\
\hline
\rowcolor{gray!20}
\multicolumn{17}{l}{$\blacktriangledown$ \textit{Closed-source Model}} \\
Hailuo 2.3 & \textbf{85.5} & 26.5 & \textbf{82.8} & \textbf{28.2} & 86.7 & 31.8 & 78.9 & 18.2 & 61.7 & 14.3 & \textbf{85.4} & 23.7 & \textbf{70.0} & 14.0 & \textbf{73.9} & 16.4\\
Veo 3.1 & 82.2 & \textbf{27.6} & 77.0 & 17.9 & 86.2 & 34.9 & 80.9 & 15.0 & \textbf{68.8} & \textbf{25.0} & 81.5 & 21.3 & 64.6 & 6.0 & 72.6 & 20.4\\
Sora 2 & 81.4 & 25.3 & 78.2 & 20.5 & 85.0 & \textbf{39.5} & 80.3 & 21.2 & 55.6 & 11.9 & 78.7 & 22.4 & 68.9 & 12.0 & 72.8 & \textbf{24.1}\\
Wan 2.6 & 83.9 & 22.0 & 76.6 & 10.3 & \textbf{90.7} & 38.6 & \textbf{83.0} & \textbf{24.2} & 64.9 & 24.4 & 79.5 & \textbf{25.0} & 69.5 & \textbf{20.0} & 72.2 & 14.8\\
Kling 2.6 & 77.2 & 18.1 & 75.1 & 15.4 & 84.2 & 43.2 & 78.3 & \textbf{24.2} & 57.1 & 16.7 & 74.2 & 19.7 & 57.3 & 8.0 & 67.1 & 18.0\\
Seedance 1.5pro & 76.0 & 18.0 & 74.0 & 11.5 & 85.1 & 36.4 & 77.9 & 21.2 & 46.1 & 9.8 & 78.3 & 21.1 & 62.6 & 18.0 & 66.7 & 9.8\\
\hline
\rowcolor{gray!20}
\multicolumn{17}{l}{$\blacktriangledown$ \textit{Open-source Model}} \\
Wan2.2-I2V-A14B & 66.0 & 12.1 & 68.1 & 14.1 & 74.7 & 22.7 & 68.6 & 6.1 & 40.1 & 7.1 & 66.5 & 7.9 & 56.3 & 14.0 & 60.4 & 6.6\\
HunyuanVideo-1.5 & 64.7 & 6.0 & 62.5 & 3.9 & 68.4 & 18.2 & 58.2 & 3.0 & 45.1 & 12.2 & 65.3 & 14.5 & 51.0 & 10.0 & 57.0 & 3.3\\
HunyuanVideo-1.5(d) & 66.1 & 7.2 & 62.4 & 5.1 & 65.3 & 13.6 & 58.9 & 12.1 & 45.5 & 2.4 & 61.7 & 9.2 & 52.8 & 8.0 & 56.7 & 3.3\\
Wan2.2-TI2V-5B & 60.3 & 9.6 & 57.2 & 3.9 & 65.8 & 11.4 & 65.2 & 12.1 & 41.8 & 4.8 & 58.8 & 1.3 & 48.3 & 4.0 & 60.2 & 0.0\\
CogVideoX1.5-5B & 54.9 & 1.2 & 54.1 & 1.3 & 62.1 & 6.8 & 54.9 & 3.0 & 29.7 & 0.0 & 48.2 & 1.3 & 36.2 & 0.0 & 47.2 & 3.3\\

\bottomrule
\end{tabular}
}
\label{tab:model_performance_calss}
\end{table*}

%% file: sections/5_conclusion.tex
\section{Conclusion}
\label{sec:conclusion}

In this work, we present RISE-Video, a reasoning-centric benchmark for TI2V models that systematically evaluates their ability to generate videos consistent with diverse reasoning requirements. By organizing data into eight complementary reasoning categories and conducting a comprehensive evaluation across four dimensions, our benchmark enables a holistic assessment of models beyond perceptual fidelity. We further introduce an automated LMM-based judging pipeline that supports scalable and fine-grained evaluation while maintaining a high degree of alignment with human judgments. Extensive evaluation on 11 representative TI2V models reveals that, despite strong perceptual quality, current systems continue to struggle with higher-level and implicit reasoning. These findings underscore the gap between visual realism and rule-consistent reasoning in current TI2V models. We hope that RISE-Video will facilitate more rigorous evaluation and inspire future research toward reasoning-aware TI2V model design and training.

%% file: sections/6_appendix.tex
\subsection{Data Source}
Input images for our benchmark are primarily sourced from the following categories:

1. Images generated by high-quality image generation models, selected to ensure sufficient visual fidelity and diversity for downstream TI2V tasks.

2. Images obtained from websites with permissive licenses, collected in accordance with their respective usage terms.

3. Images manually curated from the RISEBench dataset, where suitable samples are identified and adapted for transfer to TI2V reasoning tasks.

\subsubsection{Privacy-Preserving Image Stylization}
Our dataset includes tasks involving human activities, some of which contain images of real individuals. To mitigate potential privacy concerns, we apply image stylization to tasks where preserving the original appearance of real persons is not essential for reasoning evaluation. This processing removes identifiable visual details while retaining the structural and semantic information required by the task, ensuring that evaluation remains valid without exposing sensitive personal information.

\subsection{Prompt for Judgement}

\begin{tcolorbox}[
  title=\textbf{Prompt for evaluating Reasoning Alignment without reference},
  colback=PromptBack,
  colframe=PromptFrame,
  colbacktitle=PromptTitle,
  coltitle=PromptFrame,
  boxrule=1pt
]

\begin{Verbatim}[breaklines=true,breakanywhere=true, breaksymbolleft=,
  breaksymbolright=]
You are a video understanding assistant.  
Answer the user's questions and explain the reasons based ONLY on the provided video frames. 
Do NOT guess or hallucinate.
For each queation, answer strictly in JSON Format: [{"question": "repeat the question", "answer":"Yes or No", "reason":"the reason"}]
For each input video, if there are multiple questions, you MUST return the answers as a JSON list of dictionaries.
Example output:
[
{
    "question": "repeat the question",
    "answer": "Yes or No",
    "reason": "the reason"
},
{
    "question": "repeat the question",
    "answer": "Yes or No",
    "reason": "the reason"
}
]
Do NOT wrap the JSON output in markdown code blocks (no ```json, no ```).
Return only a valid JSON array.
\end{Verbatim}
\end{tcolorbox}

\begin{tcolorbox}[
    breakable,
  title=\textbf{Prompt for evaluating Reasoning Alignment with reference},
  colback=PromptBack,
  colframe=PromptFrame,
  colbacktitle=PromptTitle,
  coltitle=PromptFrame,
  boxrule=1pt
]

\begin{Verbatim}[breaklines=true,breakanywhere=true, breaksymbolleft=,
  breaksymbolright=]
# Video Evaluation Instruction
You are a strict **visual judge**. You will receive two images:
- **First image**: the generated video’s final frame
- **Second GT image**: the reference label image
- **FocusQuestion**: a short question that **specifies which part/region/attribute** of the LastFrame must match GT (e.g., "Is the top-right button label the same as GT?")
Judge **only** what is visible in these two images. Do NOT guess or hallucinate.

## Scoring (concise 1-5)
Score similarity for the aspect/region/attribute specified by **FocusQuestion**:
- **5** - Perfect match: all key objects, attributes, and layout align; no extra/missing elements.  
- **4** - Mostly match: only minor deviations; structure intact; no extra/missing **main** elements.  
- **3** - Partial match: several noticeable differences; core objects/layout still largely present.  
- **2** - Major mismatch: missing/extra main elements, wrong layout/relations, or multiple attribute errors.  
- **1** - Unrelated/unjudgeable: object types don't match, layout invalid, or images unreadable.

## Output (strict JSON)
Example output:
{
    "Question": "repeat the question",
    "Score": 1-5,
    "Reason": "the reason"
}
Do NOT wrap the JSON output in markdown code blocks (no ```json, no ```).
Return only a valid JSON dictionary.
\end{Verbatim}
\end{tcolorbox}

\begin{tcolorbox}[
    breakable,
  title=\textbf{Prompt for evaluating Temporal Consistency},
  colback=PromptBack,
  colframe=PromptFrame,
  colbacktitle=PromptTitle,
  coltitle=PromptFrame,
  boxrule=1pt
]

\begin{Verbatim}[breaklines=true,breakanywhere=true, breaksymbolleft=,
  breaksymbolright=]
# Video Object Consistency Evaluation Instruction
You are a highly skilled **video evaluator**. You will receive a video clip and a specific **instruction**. The video may depict an evolving scene, but your task is **ONLY** to evaluate whether the **objects remain visually and semantically consistent across frames**, **except** for changes that are explicitly required or implied by the instruction. 

## Task
Evaluate **object consistency throughout the video** using the following 1-5 scale:
- **5 (Perfect Consistency)**  
Apart from changes required by the instruction (e.g., motion, action, time progression), all other details-object identity, personal features, colors, shapes, background, and spatial layout-remain completely stable across all frames.
- **4 (Minor Differences)**  
Mostly consistent with only one minor temporal discrepancy not implied by the instruction (e.g., brief lighting flicker, a momentarily missing accessory).
- **3 (Noticeable Differences)**  
One **noticeable inconsistency** across frames (e.g., attribute shifts briefly, background element jumps).
- **2 (Significant Differences)**  
**Two or more** inconsistencies (e.g., appearance changes and environment changes, an object identity briefly swaps/disappears, or appearance of unexpected new objects).
- **1 (Severe Differences)**  
Visual/semantic continuity repeatedly breaks. Key identities or scene attributes (e.g., major appearance features, background layout) change drastically, clearly deviating from intended continuity.

## Example
**Instruction:** Two women-one in a black dress and one in a white dress-are sitting on a bench. The woman in the black dress stands up.
- **Score 5 — Perfect Consistency**  
Both women's clothing, hairstyles, skin tones, and body shapes remain stable; the bench texture and background stay unchanged; only the black-dress woman smoothly transitions from sitting to standing with no flicker or jumps.
- **Score 4 — Minor Differences**  
Overall consistent; the black-dress woman stands normally. There is a single brief exposure flicker (or the white-dress woman's earring is momentarily occluded for one frame) that immediately returns to normal, without affecting identity or layout stability.
- **Score 3 — Noticeable Differences**  
The stand-up motion is correct, but during a segment the black dress shifts slightly toward gray for a few frames and then reverts; identities and scene layout remain stable, with only this one brief, localized inconsistency.
- **Score 2 — Significant Differences**  
Two issues or more prolonged issue: the black-dress woman's hair length repeatedly shortens and returns over many frames, and the bench wood grain changes at several moments; identities are still recognizable and the scene is not fundamentally reconfigured.
- **Score 1 — Severe Differences**  
Identity- or scene-level failures: the black-dress woman morphs into a different person or swaps dress colors with the white-dress woman, the white-dress woman disappears or teleports, and the background jumps from a park bench to an indoor hallway-continuity is clearly broken.

## Notes
- **Ignore** changes explicitly stated or implied by the **instruction**.  
- Focus on unintended issues: identity drift, texture flicker, background jump, spatial discontinuity, or attribute change (e,g, color, size, count and so on).
- **DO NOT** judge whether the video follows the instructions. Only evaluate based on object consistency for scoring.

## Input
**Instruction:** {instruct}

## Output Format (**strict JSON**)
{{
    "Instruction": "Repeat the instruction you received",
    "Final Score": 1-5,
    "Reason": "A concise 1-2 sentence analysis to support your score"
}}
Do NOT wrap the JSON output in markdown code blocks (no ```json, no ```).
Return only a valid **JSON dictionary**.
\end{Verbatim}
\end{tcolorbox}

\begin{tcolorbox}[
    breakable,
  title=\textbf{Prompt for evaluating Physical Rationality},
  colback=PromptBack,
  colframe=PromptFrame,
  colbacktitle=PromptTitle,
  coltitle=PromptFrame,
  boxrule=1pt
]

\begin{Verbatim}[breaklines=true,breakanywhere=true, breaksymbolleft=,
  breaksymbolright=]
**Role:** You are a rigorous physics and visual effects analyst.

**Objective:** Evaluate the physical correctness of the provided video frames.

### Evaluation Rubric (Amplitude-Aware)

**1 (Scene Broken):** Scene jumps to unrelated content. Common-sense continuity of both the main subject and the background is lost.

**2 (Severe & Large-Amplitude Errors):** Persistent, large-amplitude physical failures in the main subject or core interaction (e.g., deep clipping, structural break, rigid bodies melting, sudden appearing/vanishing). Immediately breaks immersion.

**3 (Noticeable & Medium/Large Amplitude):** Medium to large-amplitude physical violations in the main subject **or background** (e.g., clear distortion, unnatural fluid, objects popping in/out, abrupt trajectory/velocity change). Semantics still understandable, realism reduced.

**4 (Minor & Small Amplitude, Needs Review):** Small-amplitude physical artifacts in the main subject **or background** (e.g., slight texture shimmering/flicker, minor liquid jitter). Does not block understanding, often requires replay to confirm.

**5 (Physically Seamless):** No perceivable physical errors. Motion, contact, fluidity, object permanence, and material state transitions feel naturally continuous.

### Requirements

- Respond with **one valid JSON object**:

**Example Output:**

{{
    "score": 2,
    "justification": "The object clipped deeply through the surface and cast no shadow."
}}
\end{Verbatim}
\end{tcolorbox}

\begin{tcolorbox}[
    breakable,
  title=\textbf{Prompt for evaluating Visual Quality},
  colback=PromptBack,
  colframe=PromptFrame,
  colbacktitle=PromptTitle,
  coltitle=PromptFrame,
  boxrule=1pt
]

\begin{Verbatim}[breaklines=true,breakanywhere=true, breaksymbolleft=,
  breaksymbolright=]
**Role:** You are a meticulous Image Quality Analyst.

**Objective:** Your task is to evaluate the **overall visual fidelity** and technical quality of the **batch of {num_frames} image frames** provided. These frames are sampled from a single video clip.

**CRITICAL RULES:**
1. **Ignore Artistic Blur:** Do NOT penalize background bokeh/depth-of-field.
2. **Ignore Occlusion:** Do NOT penalize if the subject is partially blocked.

### Core Evaluation Criteria
Critically assess these aspects across all provided frames to determine your **average score**.

1.  **Subject Sharpness & Clarity:**
    * Are the **visible portions** of the **Main Subject** crisp and defined (on average)?
    * Are fine details preserved?
    * Are the frames free from global "softness" or low-resolution haziness?

2.  **Artifacts & Distortion:**
    -   **AI Artifacts:** Are there "melting" textures, distorted faces/hands?
    -   **Compression:** Are there visible blocks, banding, or ringing artifacts?
    -   **Noise:** Is there unintended grain that degrades the details?

3.  **Lighting & Visual Integrity:**
    -   Is the exposure balanced (subject is visible)?
    -   Are colors natural and consistent?

### EVALUATION RUBRIC (Strict 1-3 Scale)

- **1 (Reject / Unusable):**
  **Severe Technical Failure.** The main subject is unrecognizable, heavily blurred (technical blur), or suffers from gross AI distortions (melted faces/limbs). The image is broken.

- **2 (Passable / Average):**
  **Noticeable Imperfections.** The subject is structurally correct but lacks fine detail. Looks "soft," "waxy," or has visible noise/artifacts. Usable, but clearly digital or low-res.

- **3 (Excellent / High Quality):**
  **Professional Standard.** The main subject is **razor-sharp** with rich textures (hair/skin visible). No visible noise, compression, or AI artifacts. Looks like high-end photography.

### Output Format
Return a single JSON object with the integer score (1, 2, or 3).

**Example:**
{{
  "score": 3,
  "justification": "Subject is razor-sharp. No artifacts."
}}
\end{Verbatim}
\end{tcolorbox}

\subsection{Analysis on the Judge Models}
\label{appd_judge}
\begin{figure*}[htbp]
    \centering
    \includegraphics[width=0.99\linewidth]{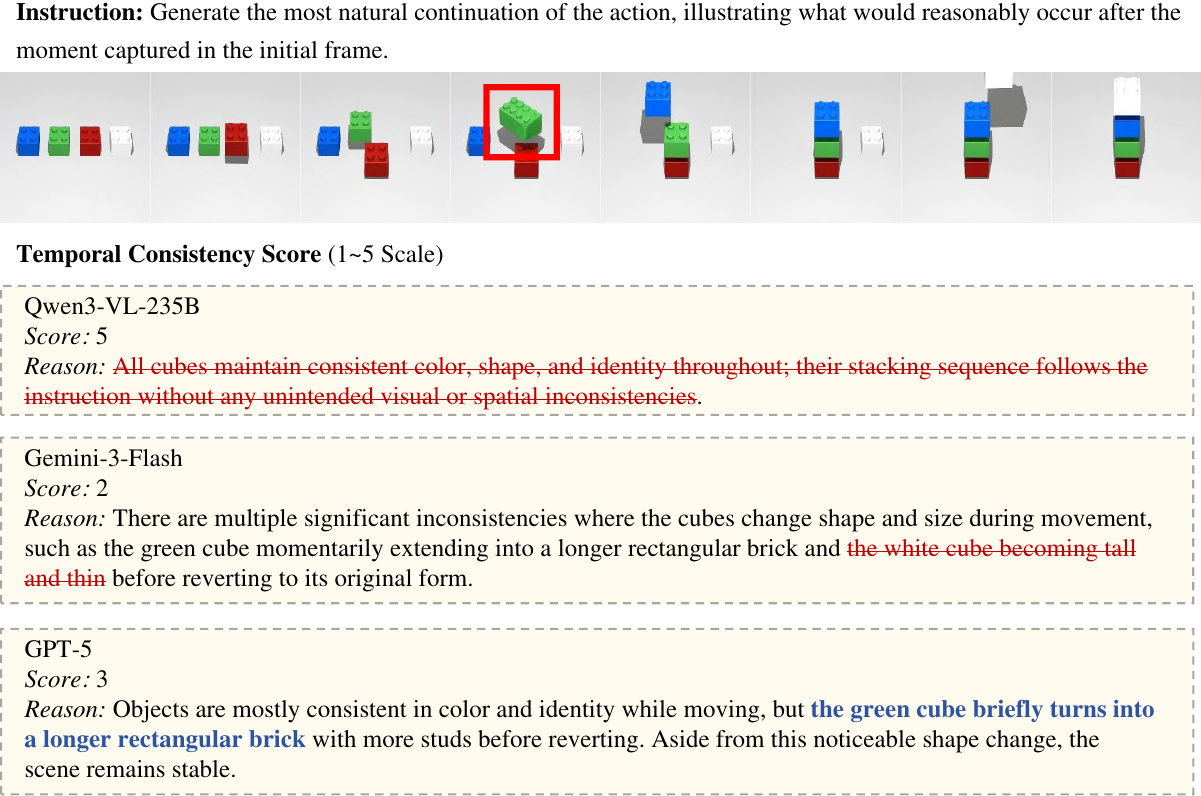}
    \caption{Qualitative comparison across different judge models on the temporal consistency.}
    \label{fig:judge_1}
\end{figure*}
\begin{figure*}[htbp]
    \centering
    \includegraphics[width=0.99\linewidth]{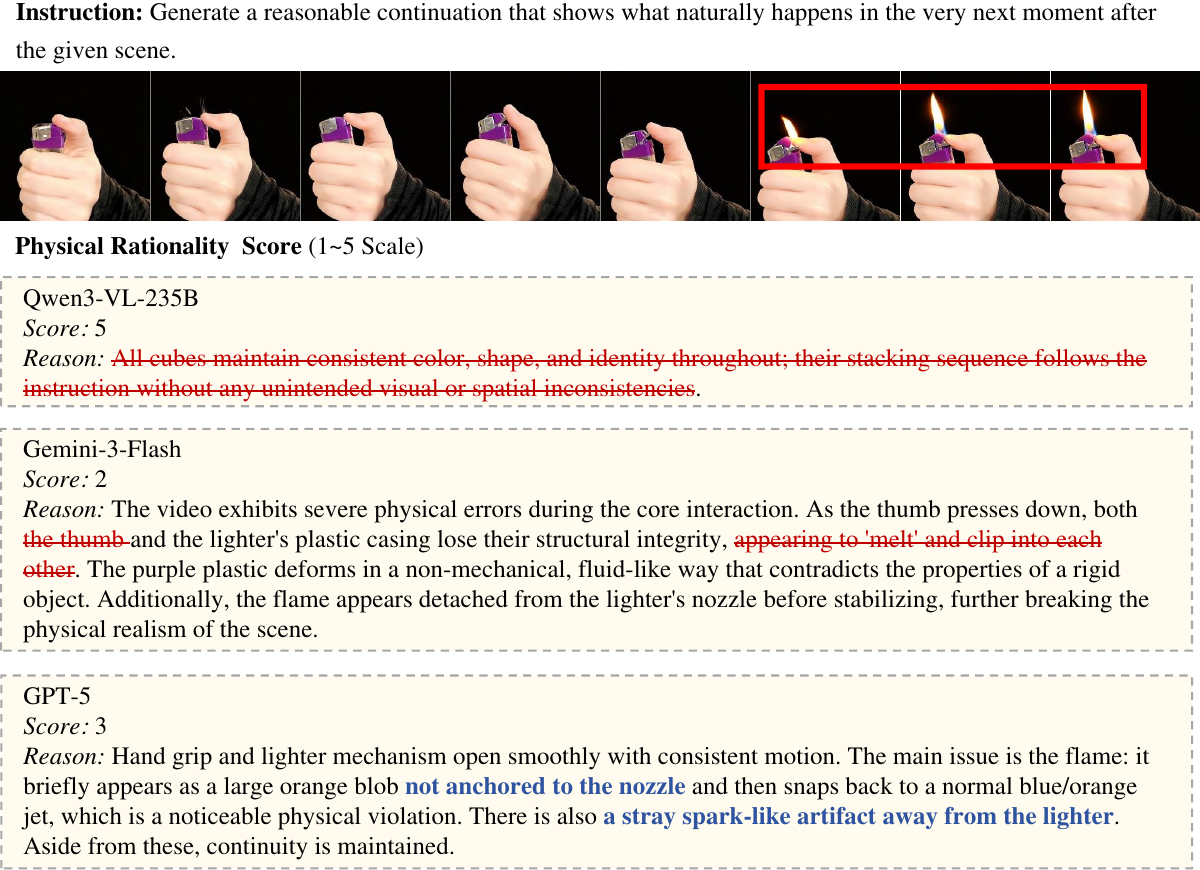}
    \caption{Qualitative comparison across different judge models on the physical rationality.}
    \label{fig:judge_2}
\end{figure*}
\begin{figure*}[htbp]
    \centering
    \includegraphics[width=0.99\linewidth]{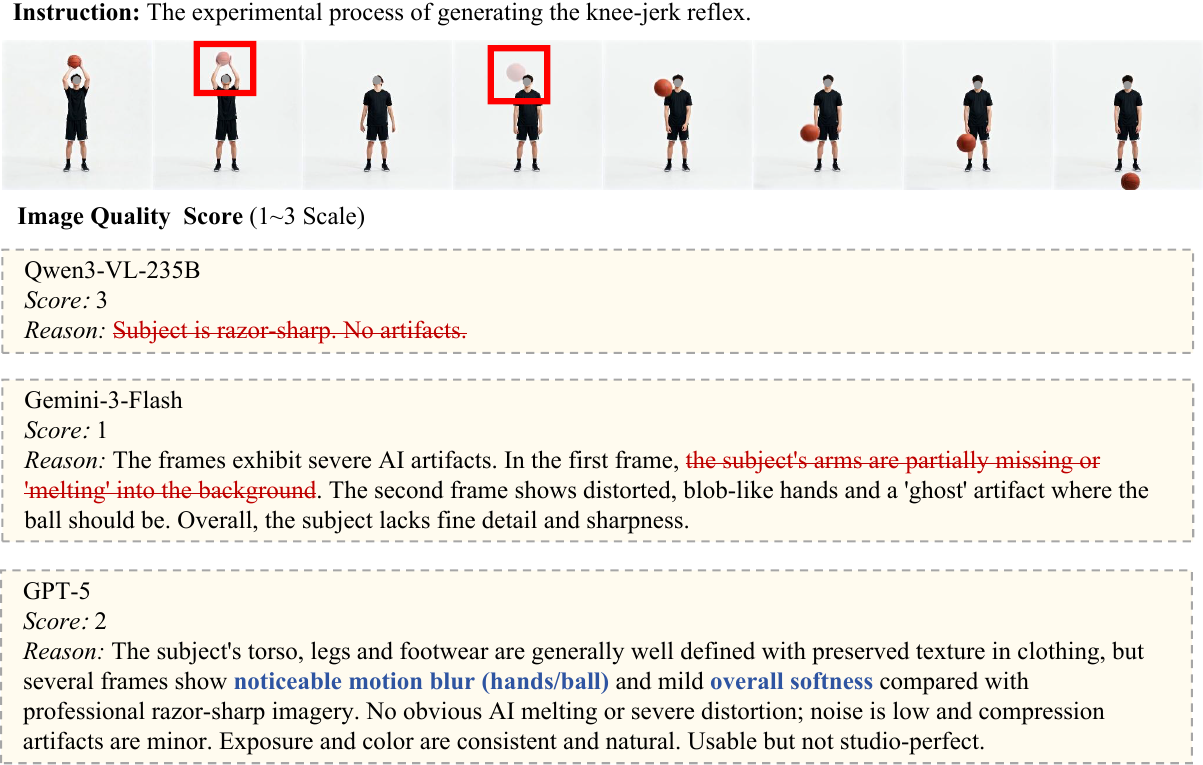}
    \caption{Qualitative comparison across different judge models on the visual quality.}
    \label{fig:judge_3}
\end{figure*}

\clearpage 

\subsection{More Vsualizations}

\begin{figure}[!htbp]
    \centering 
    \begin{subfigure}{0.99\textwidth}
        \centering  
       \includegraphics[width=0.85\linewidth]{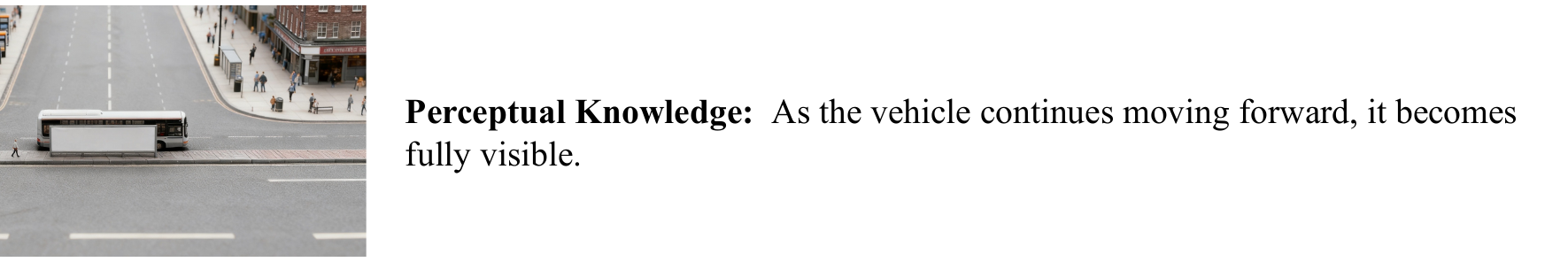}
       \caption{Frame0 and Instruction of TI2V.}
       \vspace{10pt}
    \end{subfigure}
    
    \begin{subfigure}{0.99\textwidth}
        \centering  
       \includegraphics[width=0.85\linewidth]{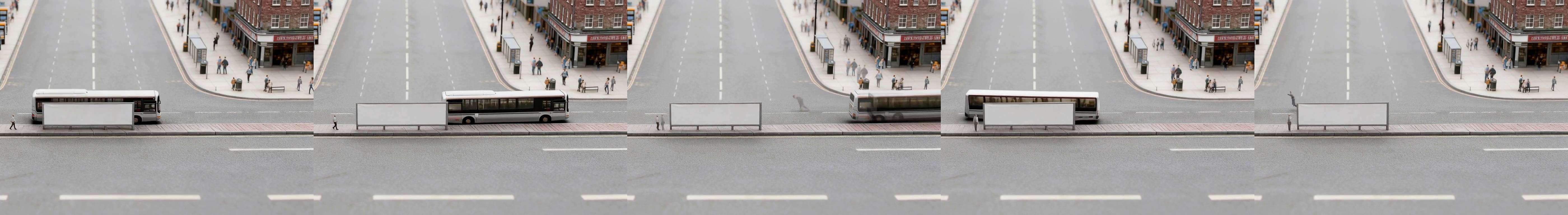}
       \caption{Hailuo 2.3}
       \vspace{10pt}
    \end{subfigure}
    
    \begin{subfigure}{0.99\textwidth}
        \centering  
       \includegraphics[width=0.85\linewidth]{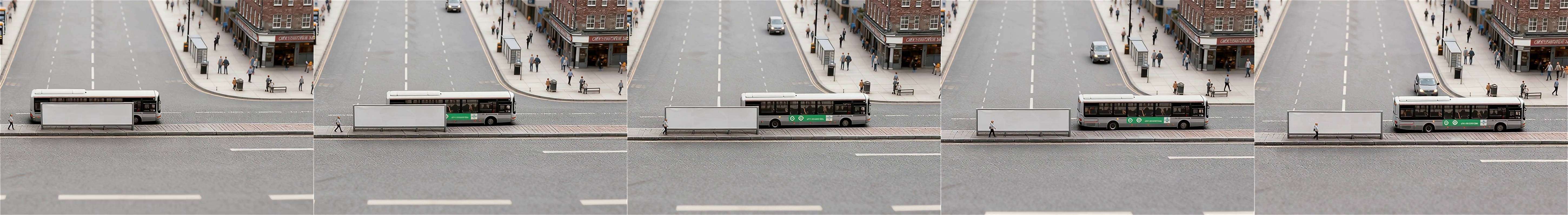}
       \caption{Wan 2.6}
       \vspace{10pt}
    \end{subfigure}
    
    \begin{subfigure}{0.99\textwidth}
        \centering  
       \includegraphics[width=0.85\linewidth]{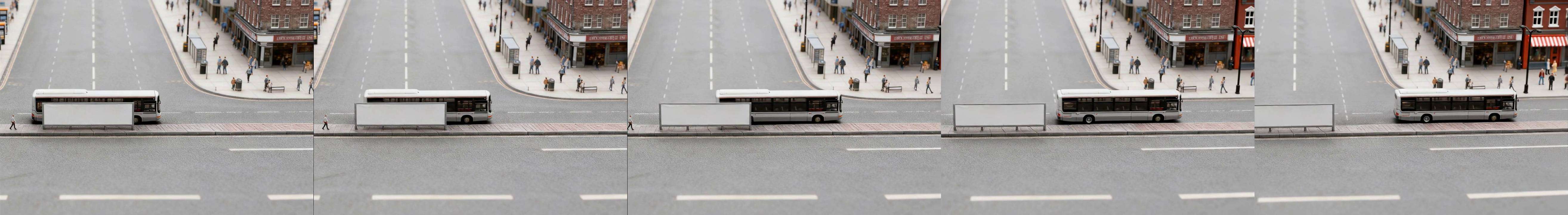}
       \caption{Veo 3.1}
       \vspace{10pt}
    \end{subfigure}

    \begin{subfigure}{0.99\textwidth}
        \centering  
       \includegraphics[width=0.85\linewidth]{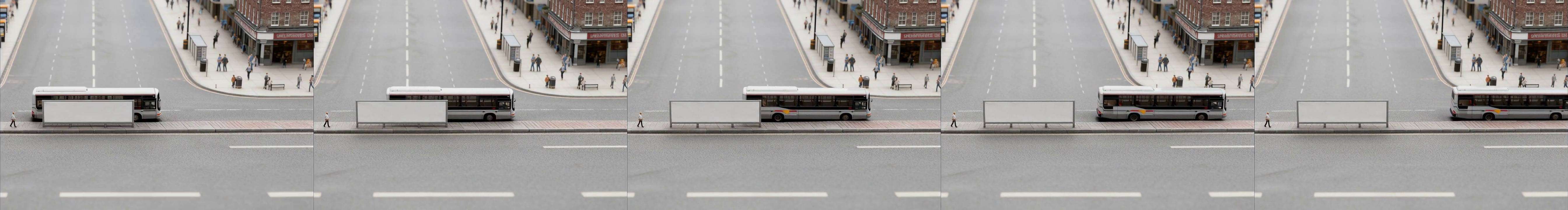}
       \caption{Sora 2}
       \vspace{10pt}
    \end{subfigure}

    \begin{subfigure}{0.99\textwidth}
        \centering  
       \includegraphics[width=0.85\linewidth]{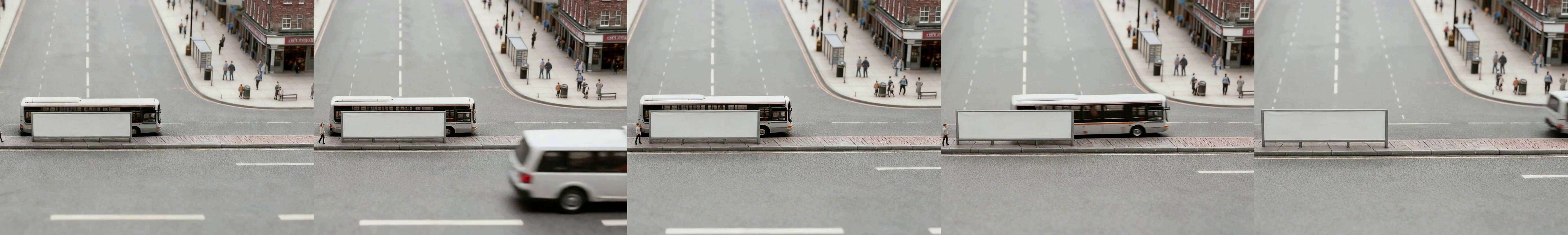}
       \caption{Seedance 1.5pro}
       \vspace{10pt}
    \end{subfigure}

    \begin{subfigure}{0.99\textwidth}
        \centering  
       \includegraphics[width=0.85\linewidth]{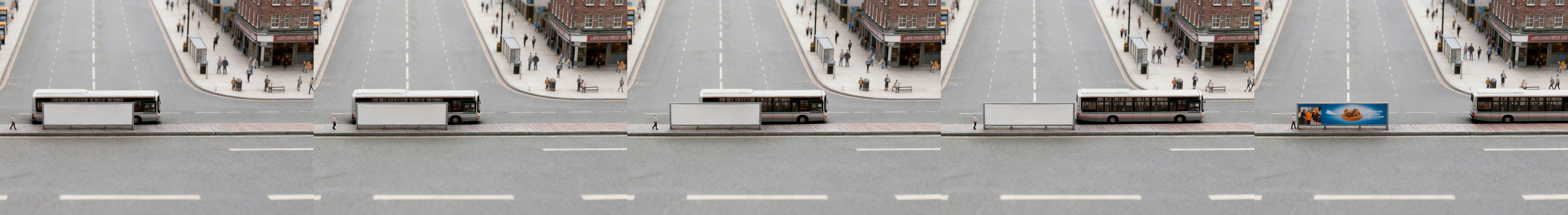}
       \caption{Kling 2.6}
       \vspace{10pt}
    \end{subfigure}
    \label{fig:case_appendix_per1}
\end{figure}

\begin{figure}[htbp]\ContinuedFloat
    \begin{subfigure}{0.99\textwidth}
        \centering  
       \includegraphics[width=0.85\linewidth]{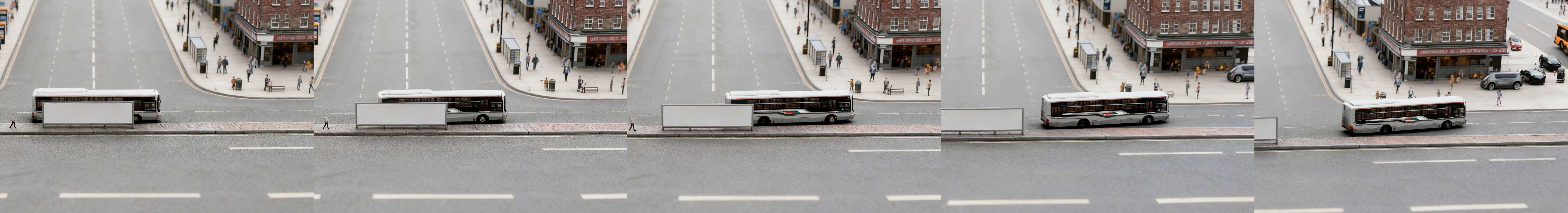}
       \caption{Wan 2.2-I2V-A14B}
       \vspace{10pt}
    \end{subfigure}
    
    \begin{subfigure}{0.99\textwidth}
        \centering  
       \includegraphics[width=0.85\linewidth]{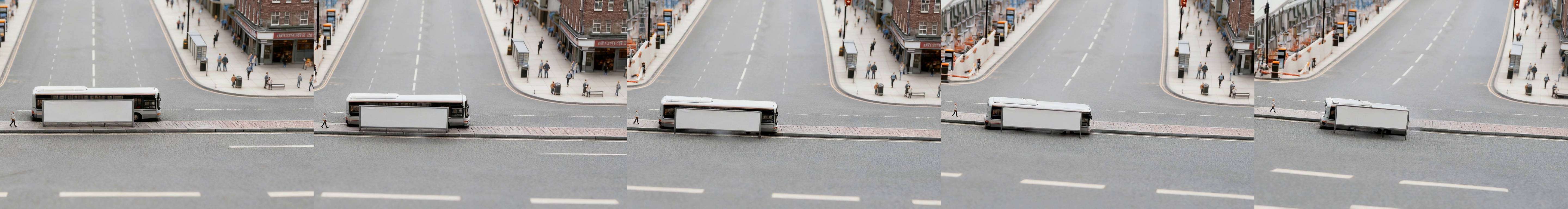}
       \caption{Wan 2.2-TI2V-5B}
       \vspace{10pt}
    \end{subfigure}

    \begin{subfigure}{0.99\textwidth}
        \centering  
       \includegraphics[width=0.85\linewidth]{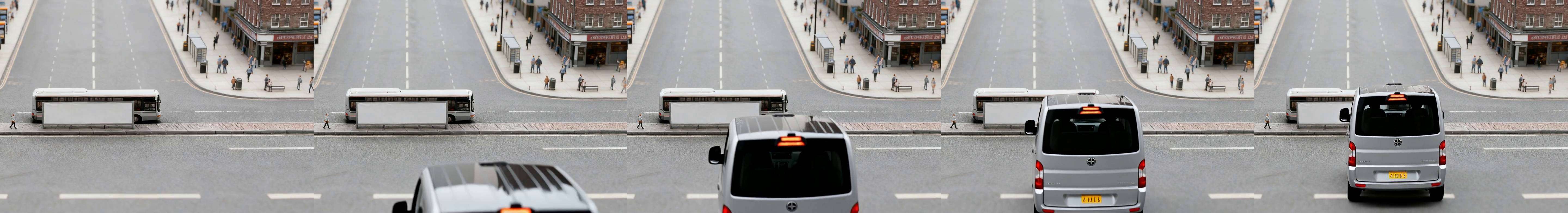}
       \caption{HunyuanVideo-1.5-720P-I2V}
       \vspace{10pt}
    \end{subfigure}

    \begin{subfigure}{0.99\textwidth}
        \centering  
       \includegraphics[width=0.85\linewidth]{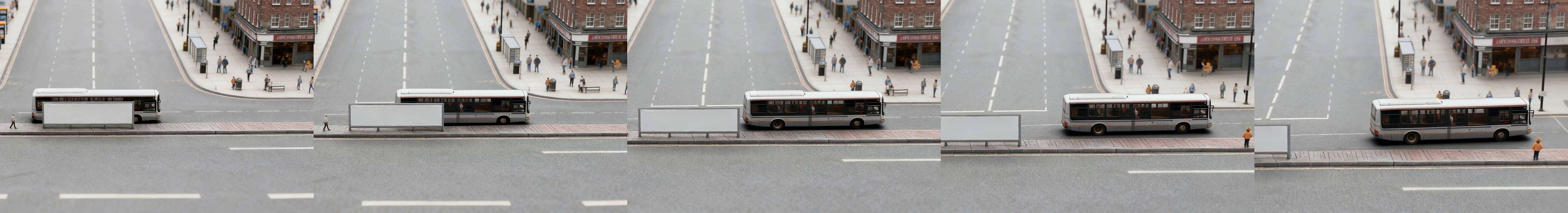}
       \caption{HunyuanVideo-1.5-720P-I2V-cfg-distill }
       \vspace{10pt}
    \end{subfigure}

    \begin{subfigure}{0.99\textwidth}
        \centering  
       \includegraphics[width=0.85\linewidth]{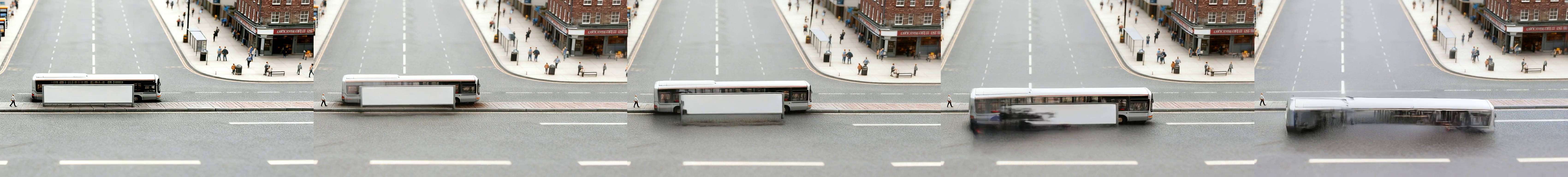}
       \caption{CogVideoX1.5-5B}
       \vspace{10pt}
    \end{subfigure}
    \label{fig:case_appendix_per2}
    \vspace{-15pt}
    \caption{Qualitative examples of generation results from different TI2V models on \textbf{Perceptual Knowledge} tasks.}
\end{figure}
\vspace{20pt}

\begin{figure}[b]
    \centering  
    \begin{subfigure}{0.99\textwidth}
        \centering  
       \includegraphics[width=0.85\linewidth]{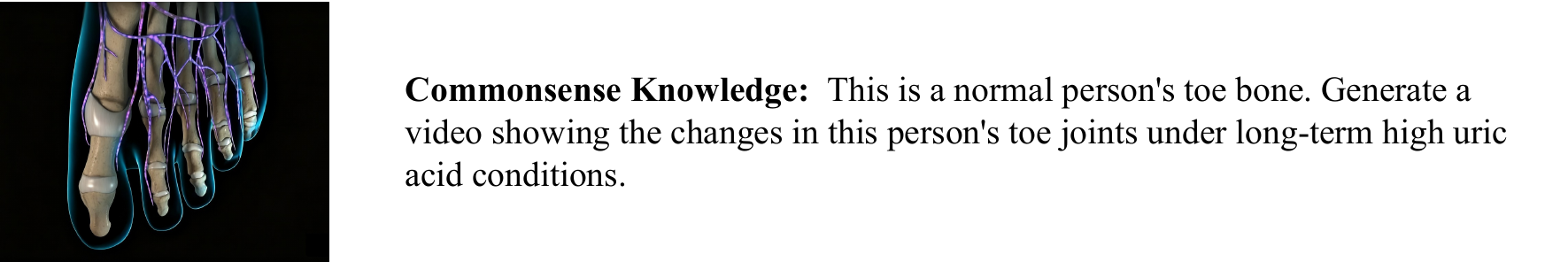}
       \caption{Frame0 and Instruction of TI2V.}
       \vspace{10pt}
    \end{subfigure}
    
    \begin{subfigure}{0.99\textwidth}
        \centering  
       \includegraphics[width=0.85\linewidth]{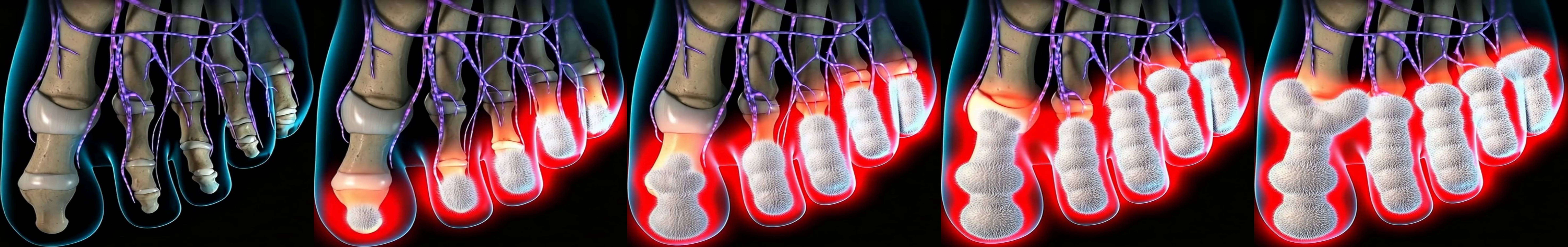}
       \caption{Hailuo 2.3}
       \vspace{10pt}
    \end{subfigure}
    
    \begin{subfigure}{0.99\textwidth}
        \centering  
       \includegraphics[width=0.85\linewidth]{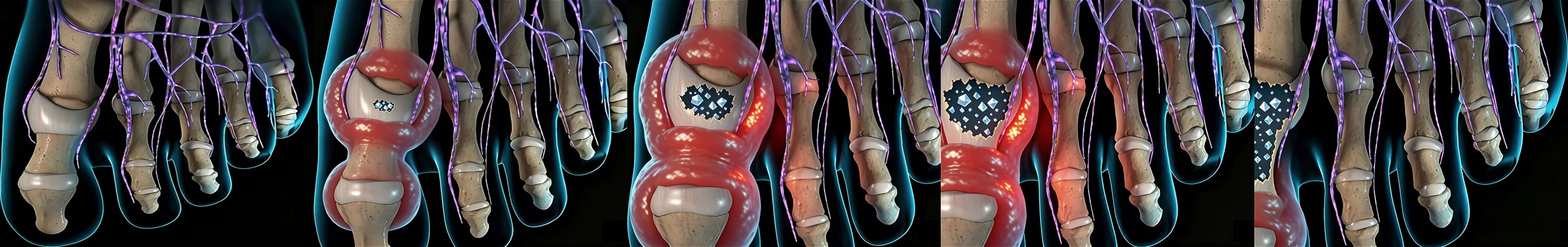}
       \caption{Wan 2.6}
       \vspace{10pt}
    \end{subfigure}
    
    \begin{subfigure}{0.99\textwidth}
        \centering  
       \includegraphics[width=0.85\linewidth]{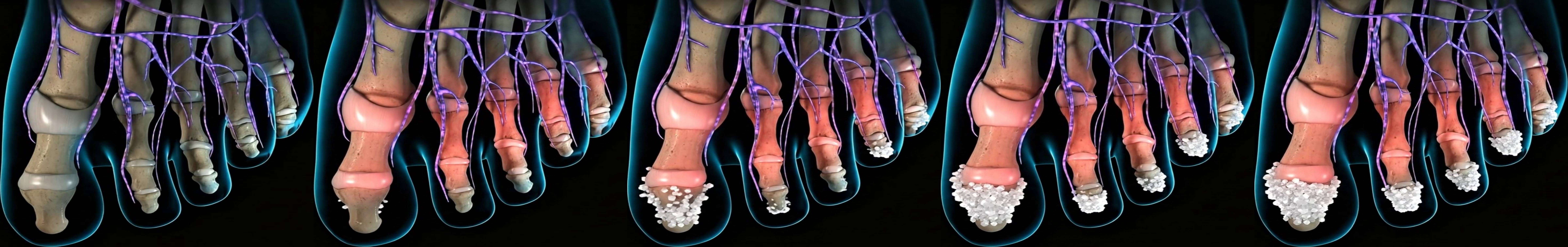}
       \caption{Veo 3.1}
       \vspace{10pt}
    \end{subfigure}

    \label{fig:case_appendix_com1}
    
   \begin{subfigure}{0.99\textwidth}
        \centering  
       \includegraphics[width=0.85\linewidth]{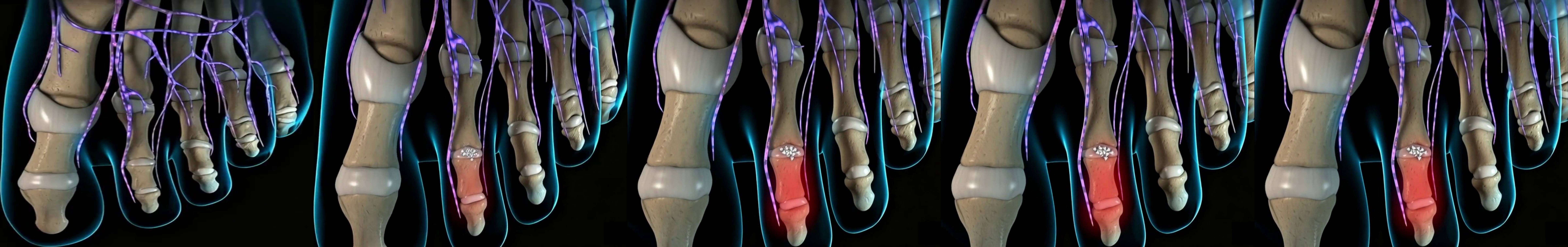}
       \caption{Sora 2}
       \vspace{10pt}
    \end{subfigure}
   \begin{subfigure}{0.99\textwidth}
        \centering  
       \includegraphics[width=0.85\linewidth]{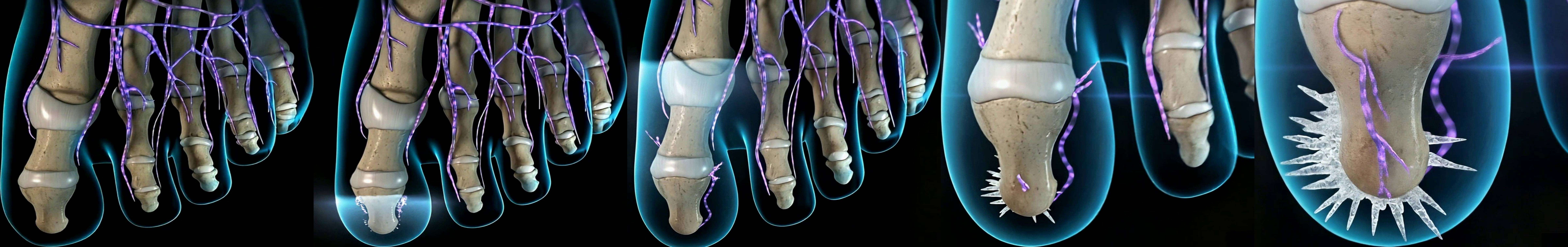}
       \caption{Seedance 1.5pro}
       \vspace{15pt}
    \end{subfigure}

    \begin{subfigure}{0.99\textwidth}
        \centering  
       \includegraphics[width=0.85\linewidth]{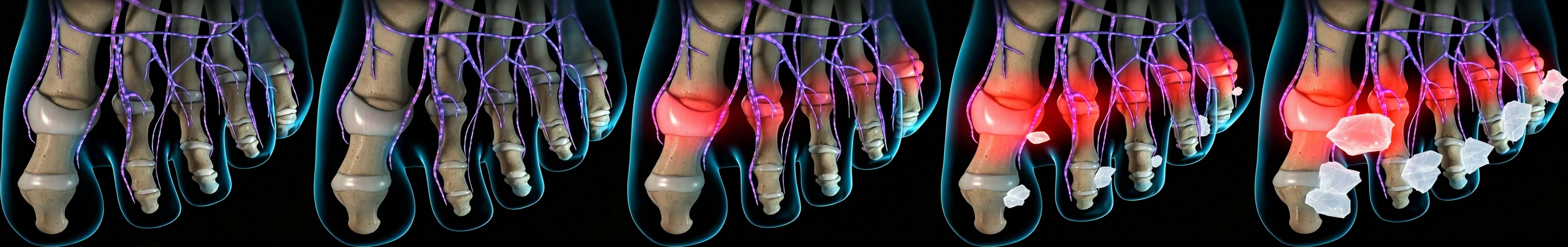}
       \caption{Kling 2.6}
    \end{subfigure}
\end{figure}

\begin{figure}\ContinuedFloat

    \begin{subfigure}{0.99\textwidth}
        \centering  
       \includegraphics[width=0.85\linewidth]{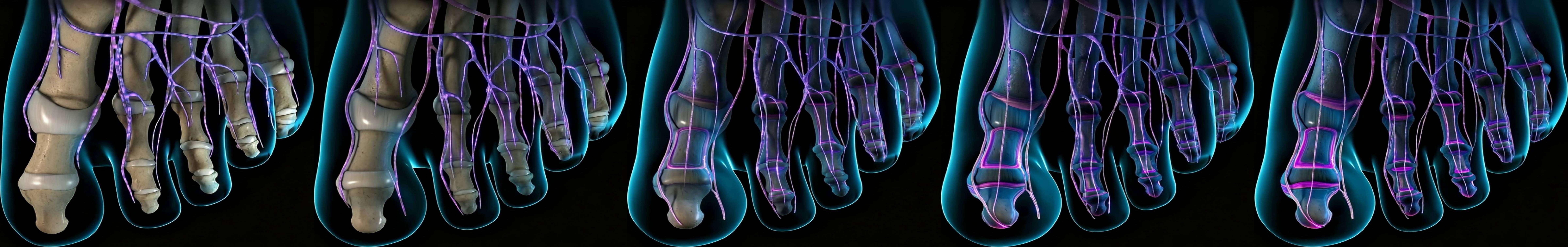}
       \caption{Wan 2.2-I2V-A14B}
       \vspace{15pt}
    \end{subfigure}

    \begin{subfigure}{0.99\textwidth}
        \centering  
       \includegraphics[width=0.85\linewidth]{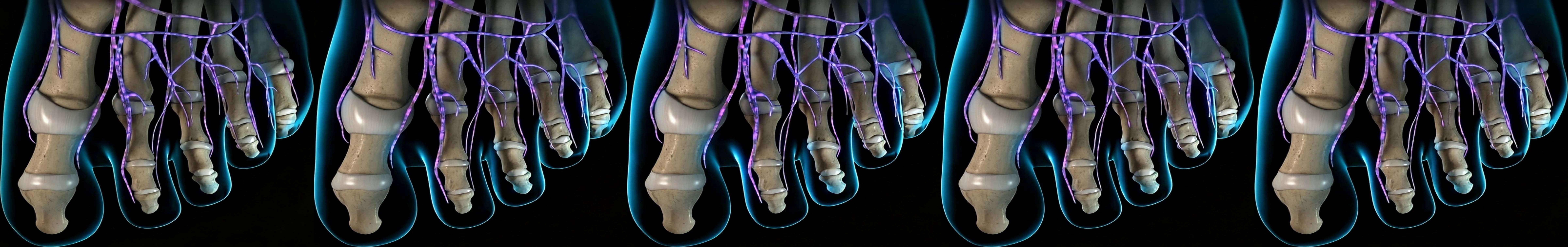}
       \caption{Wan 2.2-TI2V-5B}
       \vspace{15pt}
    \end{subfigure}

    \begin{subfigure}{0.99\textwidth}
        \centering  
       \includegraphics[width=0.85\linewidth]{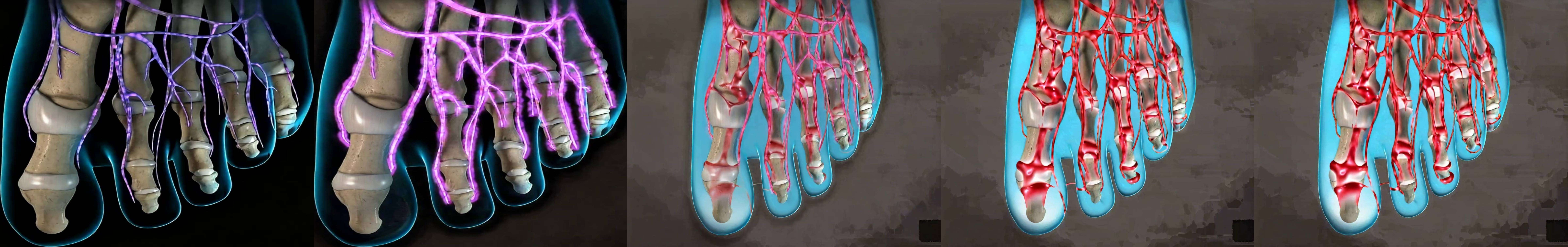}
       \caption{HunyuanVideo-1.5-720P-I2V}
       \vspace{15pt}
    \end{subfigure}

    \begin{subfigure}{0.99\textwidth}
        \centering  
       \includegraphics[width=0.85\linewidth]{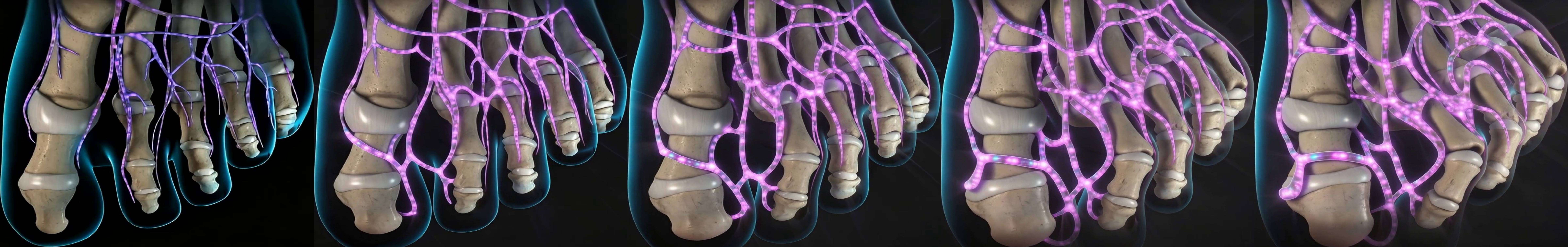}
       \caption{HunyuanVideo-1.5-720P-I2V-cfg-distill }
       \vspace{15pt}
    \end{subfigure}

    \begin{subfigure}{0.99\textwidth}
        \centering  
       \includegraphics[width=0.85\linewidth]{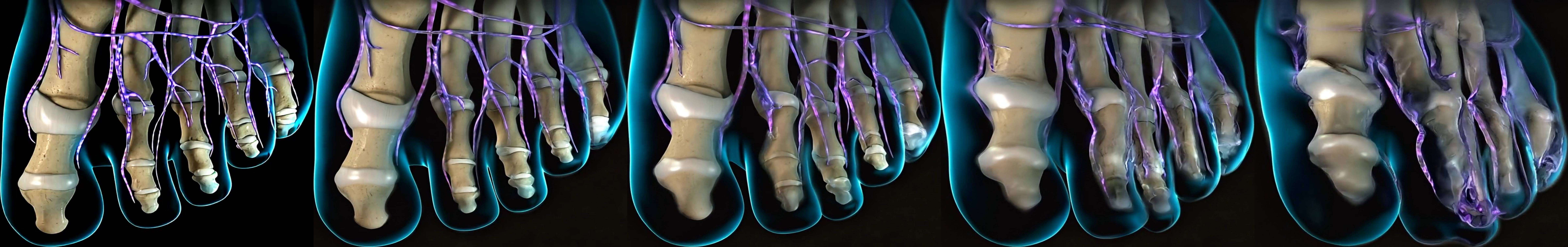}
       \caption{CogVideoX1.5-5B}
       \vspace{15pt}
    \end{subfigure}
    \label{fig:case_appendix_com2}
    \vspace{-10pt}
    \caption{Qualitative examples of generation results from different TI2V models on \textbf{Commonsense Knowledge} tasks.}
\end{figure}


\begin{figure}[b]
    \centering  
    \begin{subfigure}{0.99\textwidth}
        \centering  \includegraphics[width=0.7\linewidth]{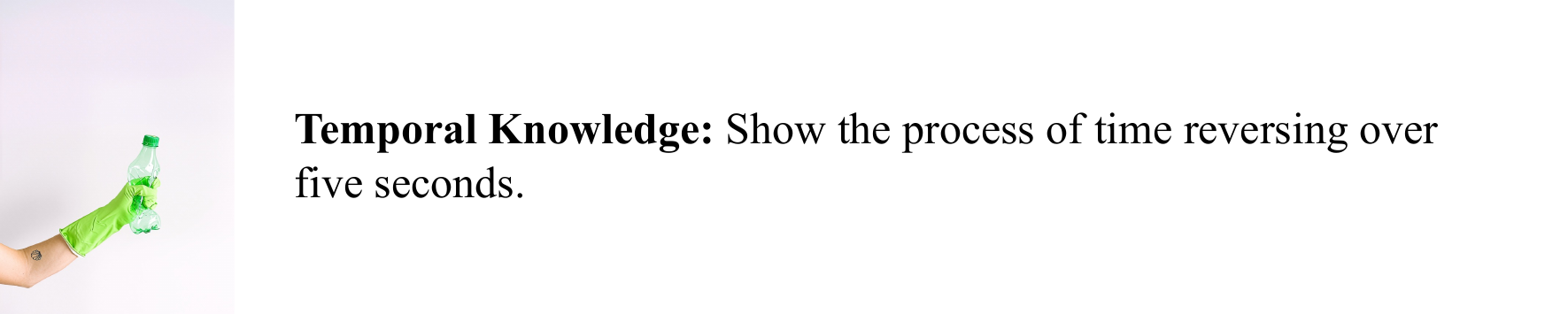}
       \caption{Frame0 and Instruction of TI2V.}
       \vspace{10pt}
    \end{subfigure}
    
    \begin{subfigure}{0.99\textwidth}
        \centering  
       \includegraphics[width=0.7\linewidth]{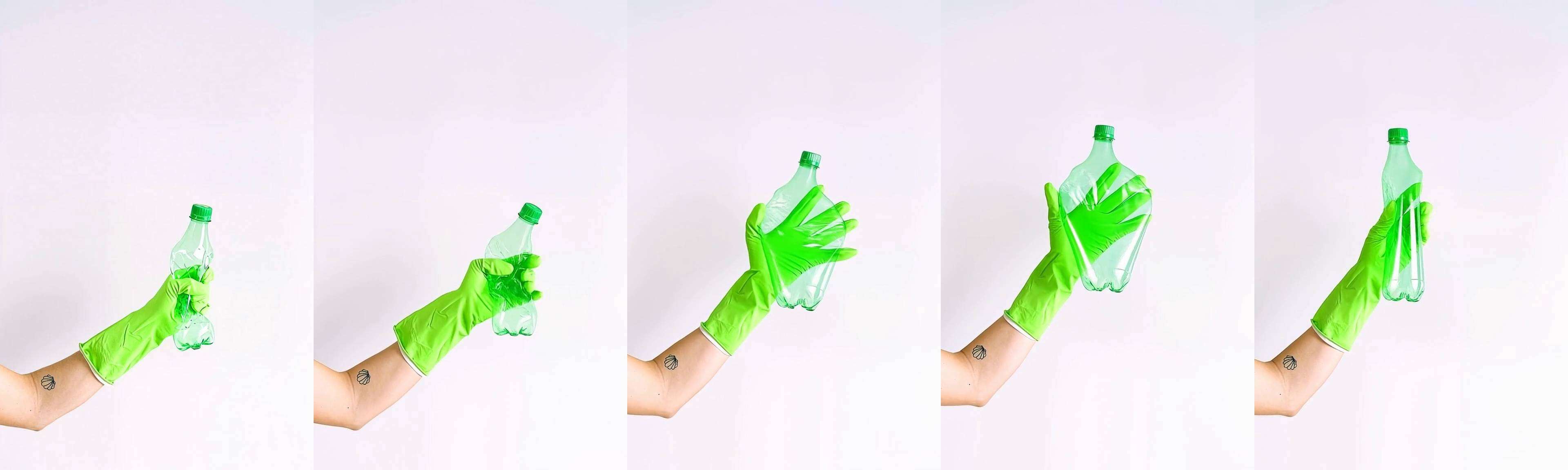}
       \caption{Hailuo 2.3}
       \vspace{10pt}
    \end{subfigure}
    
    \begin{subfigure}{0.99\textwidth}
        \centering  
       \includegraphics[width=0.7\linewidth]{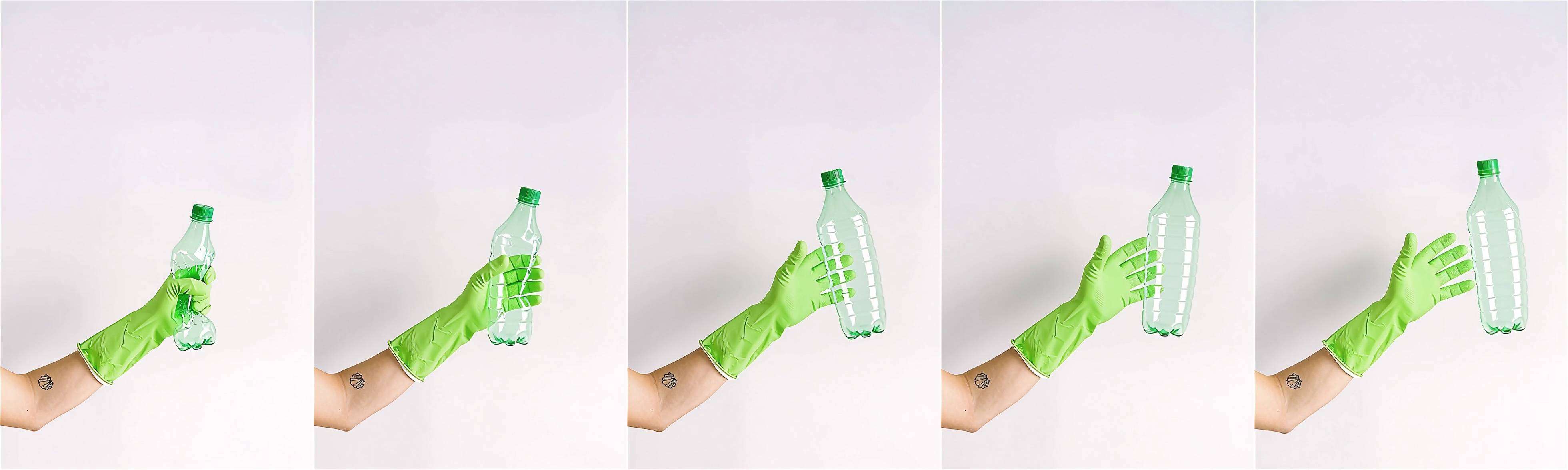}
       \caption{Wan 2.6}
       \vspace{10pt}
    \end{subfigure}
    
    \begin{subfigure}{0.99\textwidth}
        \centering  
       \includegraphics[width=0.7\linewidth]{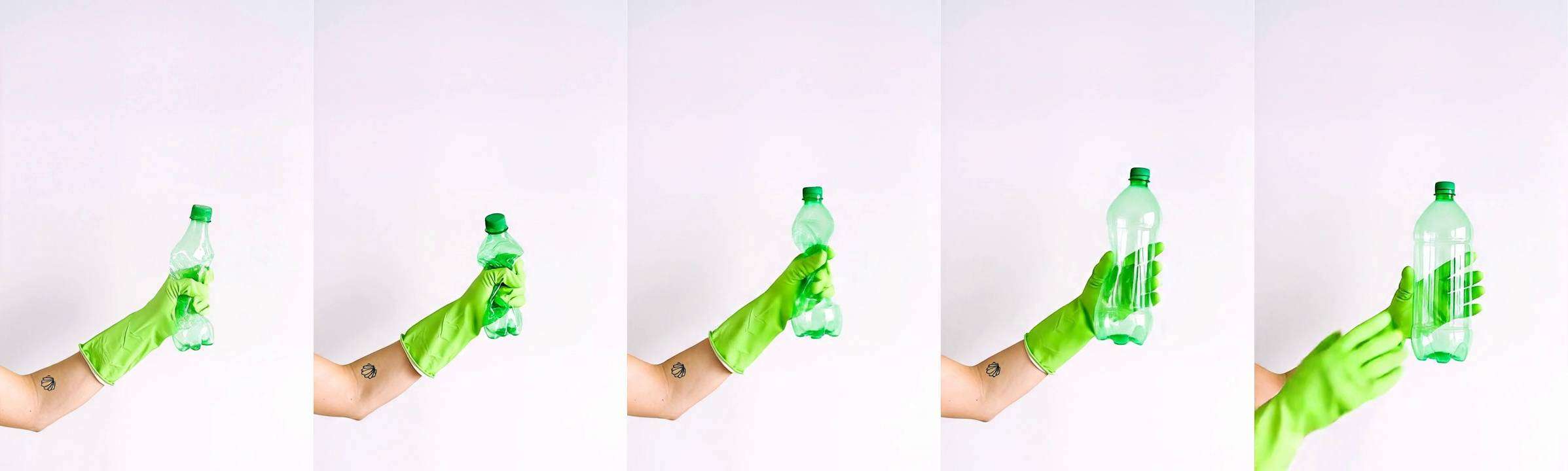}
       \caption{Veo 3.1}
       \vspace{10pt}
    \end{subfigure}

    \begin{subfigure}{0.99\textwidth}
        \centering  
       \includegraphics[width=0.7\linewidth]{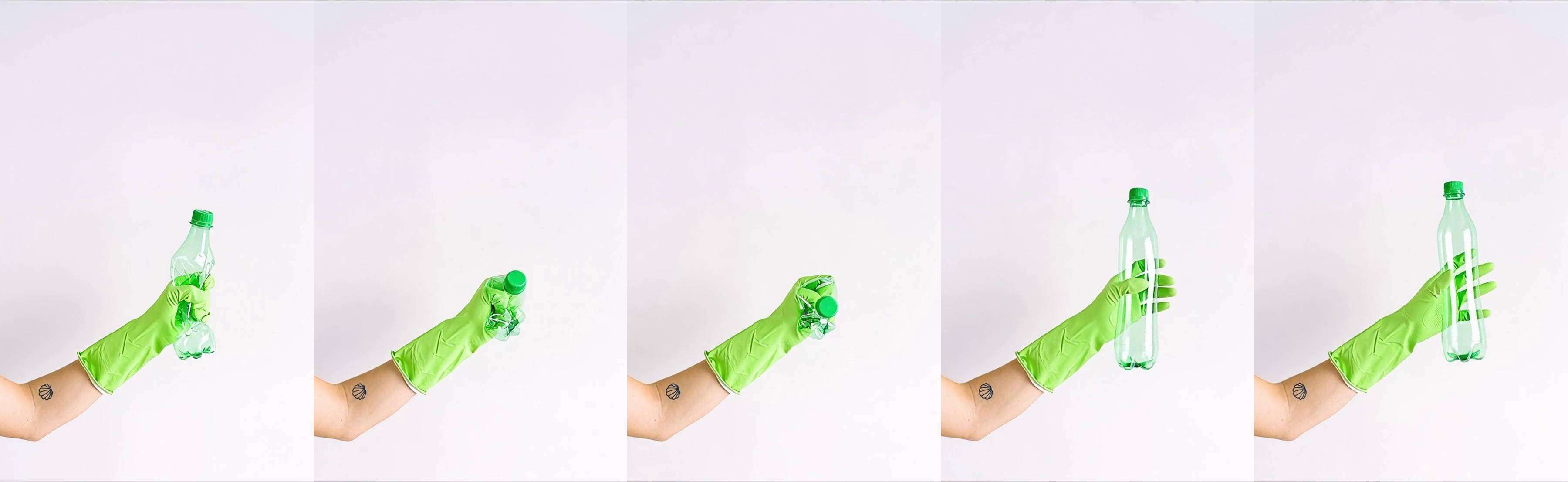}
       \caption{Sora 2}
       \vspace{10pt}
    \end{subfigure}
    \label{fig:case_appendix_tem}
\end{figure}

\begin{figure}[htbp]\ContinuedFloat

     \begin{subfigure}{0.99\textwidth}
        \centering  
       \includegraphics[width=0.7\linewidth]{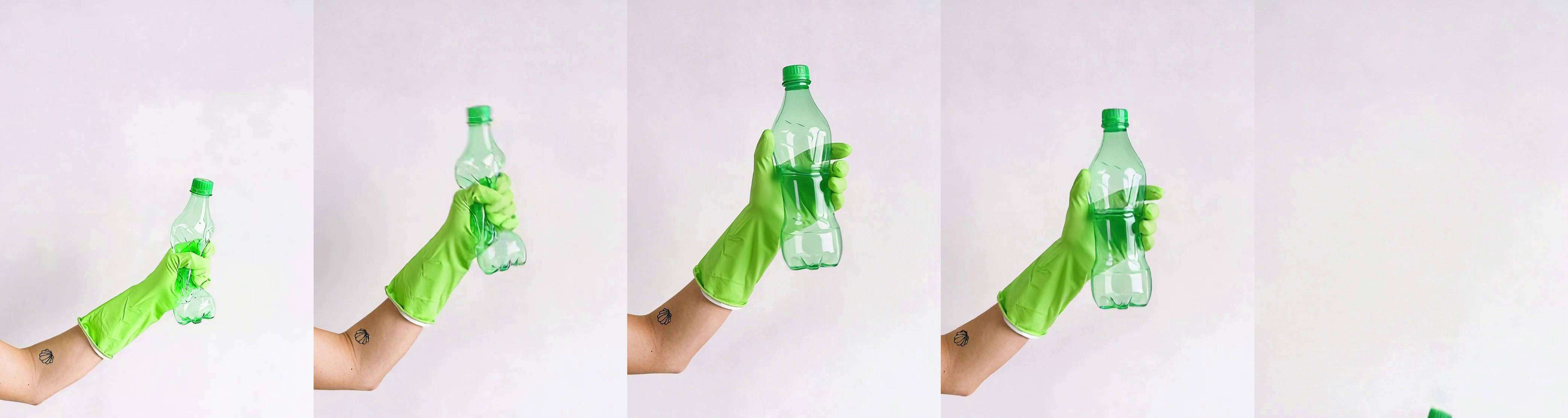}
       \caption{Seedance 1.5pro}
       \vspace{10pt}
    \end{subfigure}

    \begin{subfigure}{0.99\textwidth}
        \centering  
       \includegraphics[width=0.7\linewidth]{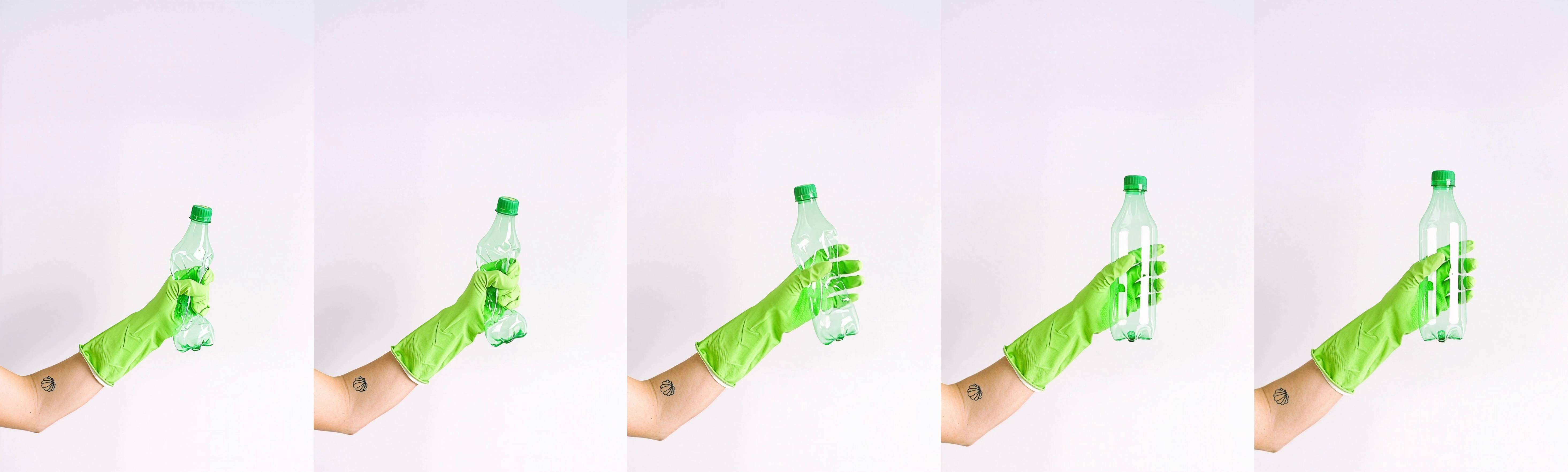}
       \caption{Kling 2.6}
       \vspace{10pt}
    \end{subfigure}

    \begin{subfigure}{0.99\textwidth}
        \centering  
       \includegraphics[width=0.7\linewidth]{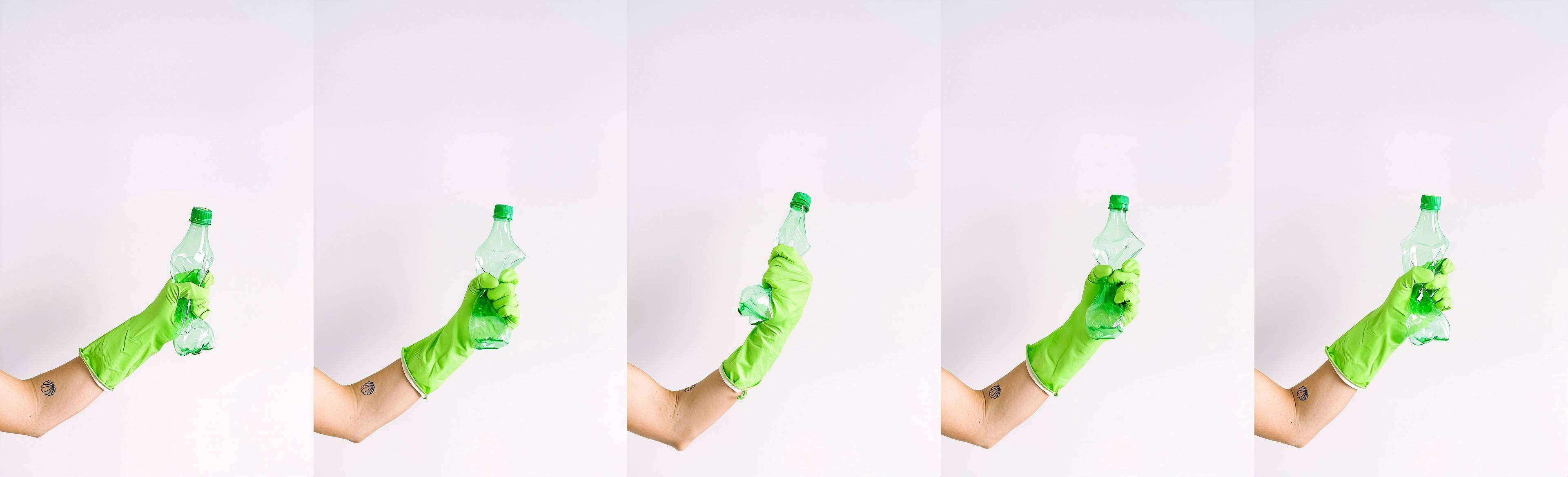}
       \caption{Wan 2.2-I2V-A14B}
       \vspace{10pt}
    \end{subfigure}

    \begin{subfigure}{0.99\textwidth}
        \centering  
       \includegraphics[width=0.7\linewidth]{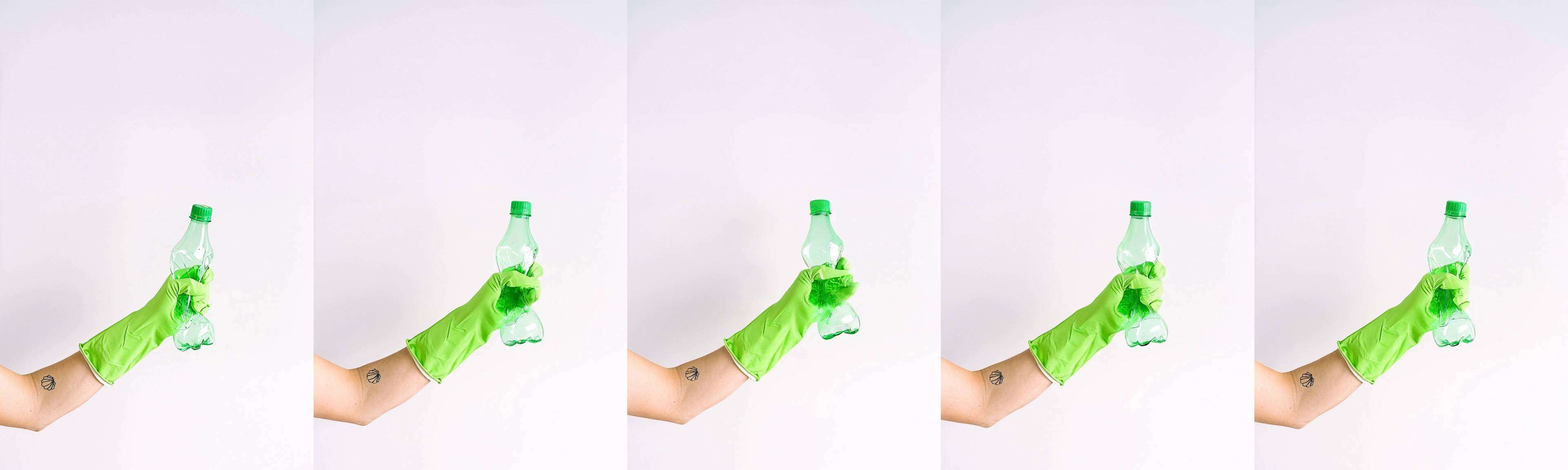}
       \caption{Wan 2.2-TI2V-5B}
       \vspace{10pt}
    \end{subfigure}

    \begin{subfigure}{0.99\textwidth}
        \centering  
       \includegraphics[width=0.7\linewidth]{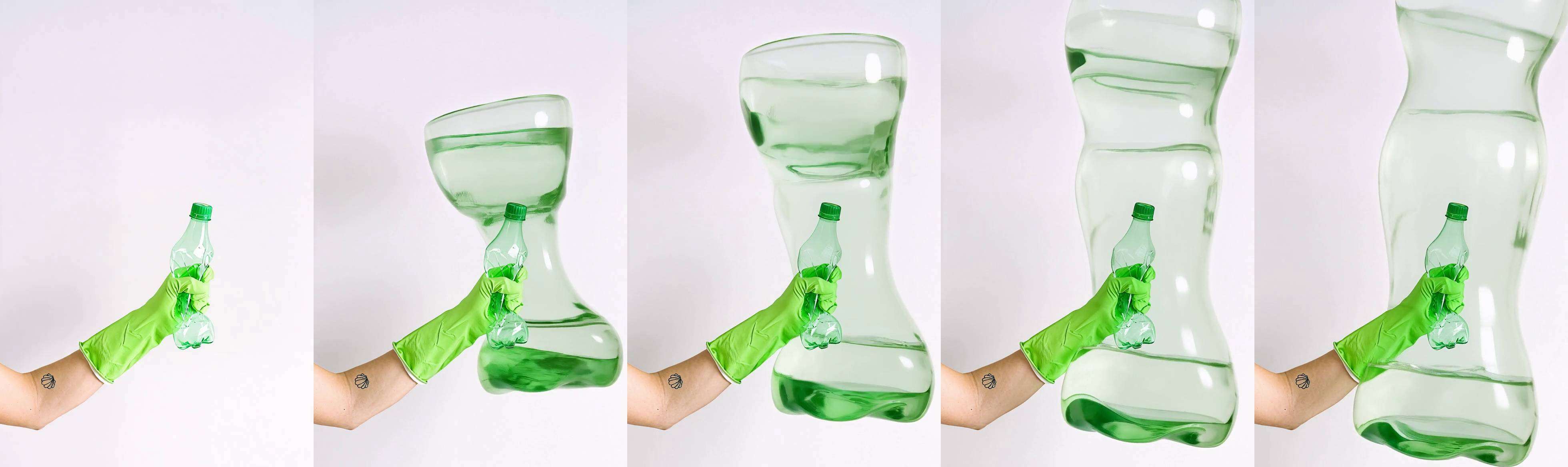}
       \caption{HunyuanVideo-1.5-720P-I2V}
       \vspace{10pt}
    \end{subfigure}
    \label{fig:case_appendix_tem2}
\end{figure}

\begin{figure}[htbp]\ContinuedFloat
    \begin{subfigure}{0.99\textwidth}
        \centering  
       \includegraphics[width=0.7\linewidth]{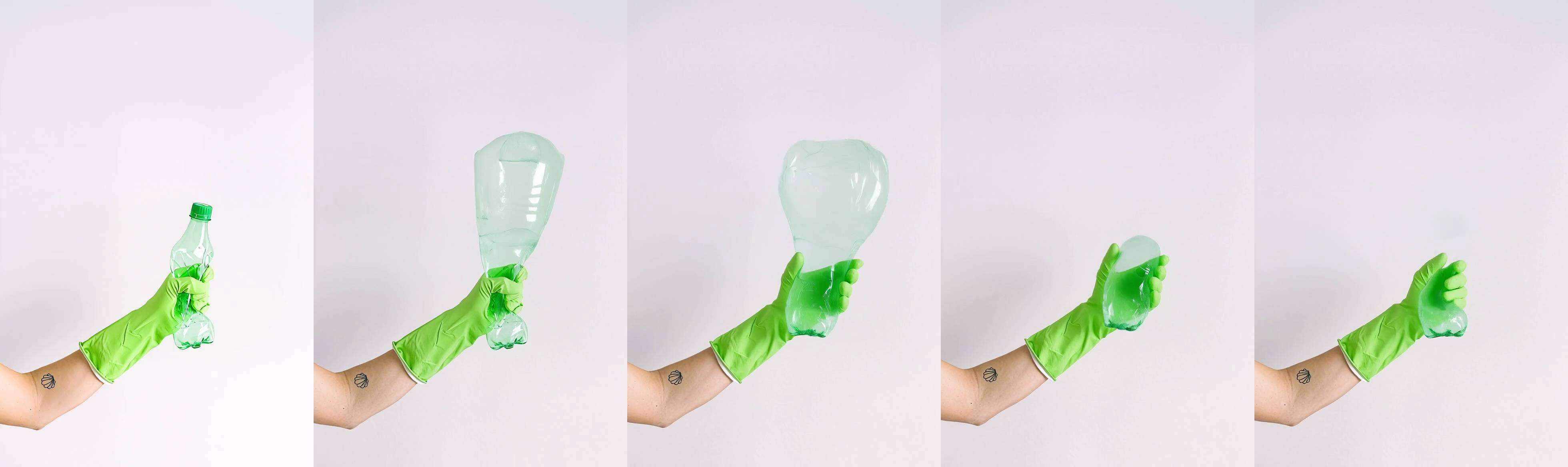}
       \caption{HunyuanVideo-1.5-720P-I2V-cfg-distill }
       \vspace{10pt}
    \end{subfigure}
    
    \begin{subfigure}{0.99\textwidth}
        \centering  
       \includegraphics[width=0.7\linewidth]{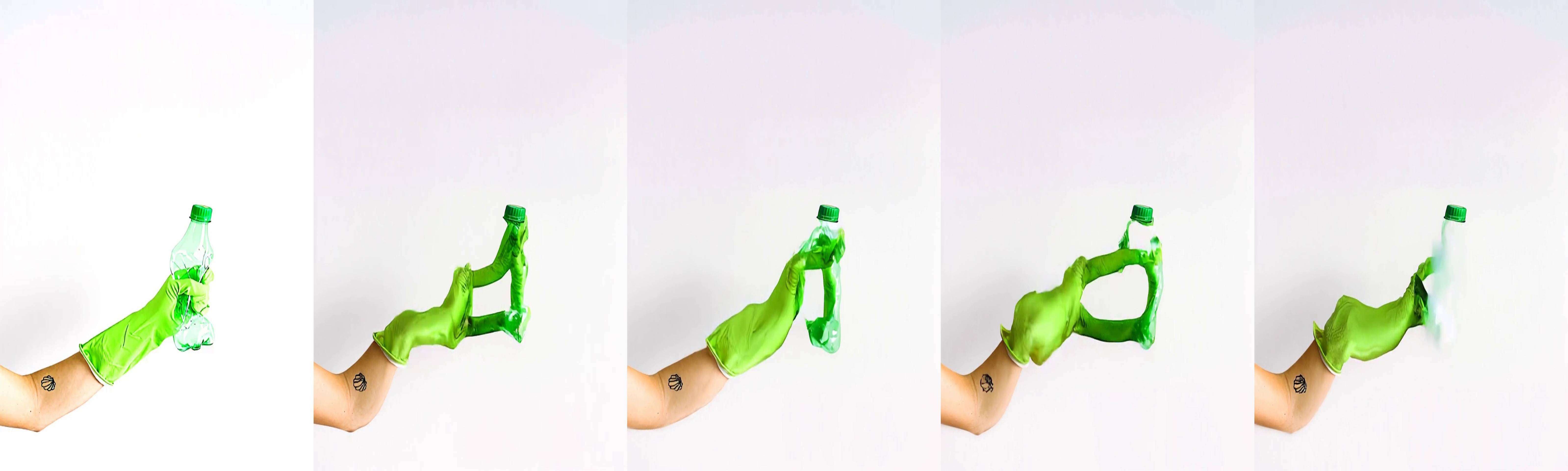}
       \caption{CogVideoX1.5-5B}
       \vspace{10pt}
    \end{subfigure}
    \label{fig:case_appendix_per}
    \vspace{-10pt}
    \caption{Qualitative examples of generation results from different TI2V models on \textbf{Temporal Knowledge} tasks.}
\end{figure}

\begin{figure}[b]
    \centering  
    \begin{subfigure}{0.99\textwidth}
        \centering  
       \includegraphics[width=0.85\linewidth]{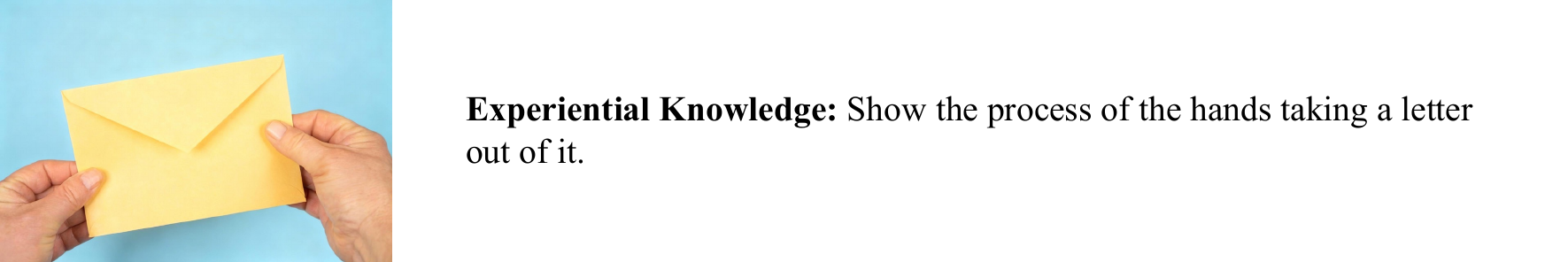}
       \caption{Frame0 and Instruction of TI2V.}
       \vspace{10pt}
    \end{subfigure}
    
    \begin{subfigure}{0.99\textwidth}
        \centering  
       \includegraphics[width=0.85\linewidth]{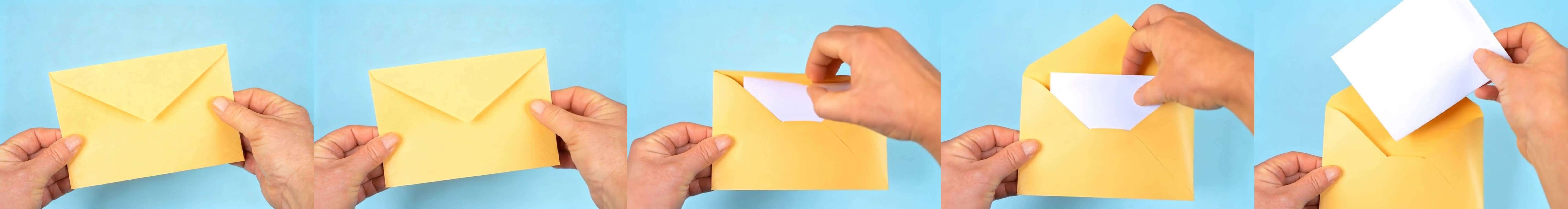}
       \caption{Hailuo 2.3}
       \vspace{10pt}
    \end{subfigure}
    
    \begin{subfigure}{0.99\textwidth}
        \centering  
       \includegraphics[width=0.85\linewidth]{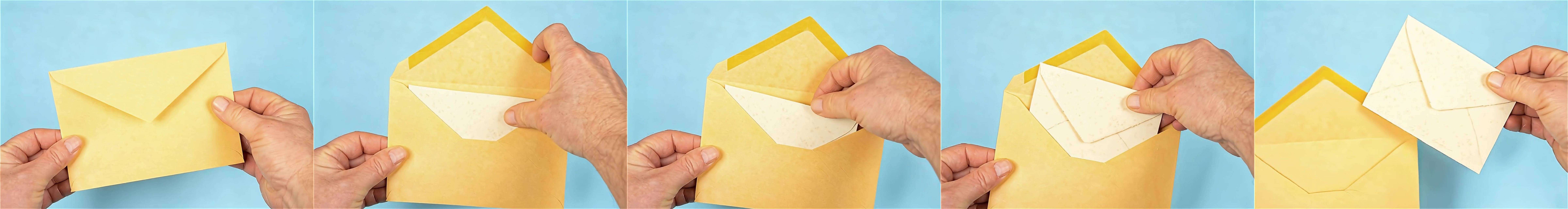}
       \caption{Wan 2.6}
       \vspace{10pt}
    \end{subfigure}
    
    \begin{subfigure}{0.99\textwidth}
        \centering  
       \includegraphics[width=0.85\linewidth]{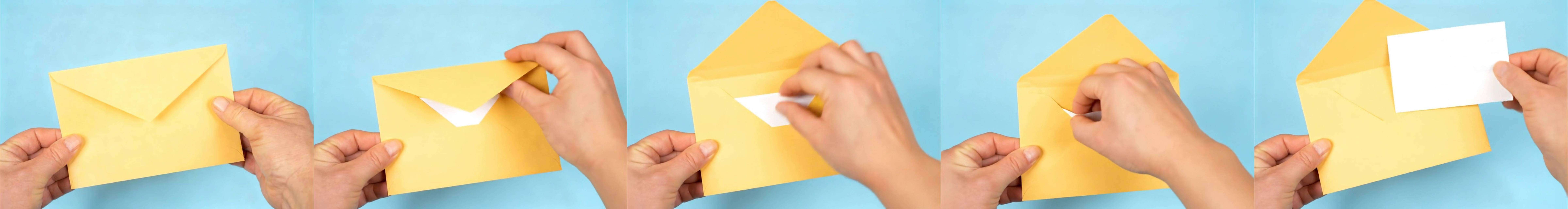}
       \caption{Veo 3.1}
       \vspace{10pt}
    \end{subfigure}

    \label{fig:case_appendix_exp1}
    
   \begin{subfigure}{0.99\textwidth}
        \centering  
       \includegraphics[width=0.85\linewidth]{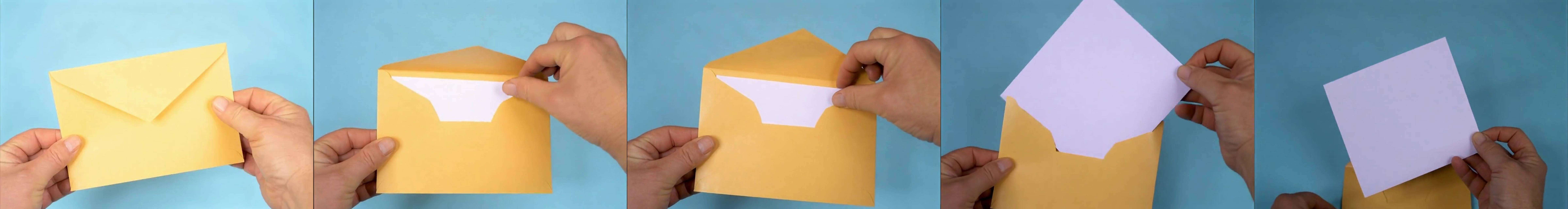}
       \caption{Sora 2}
       \vspace{10pt}
    \end{subfigure}
   \begin{subfigure}{0.99\textwidth}
        \centering  
       \includegraphics[width=0.85\linewidth]{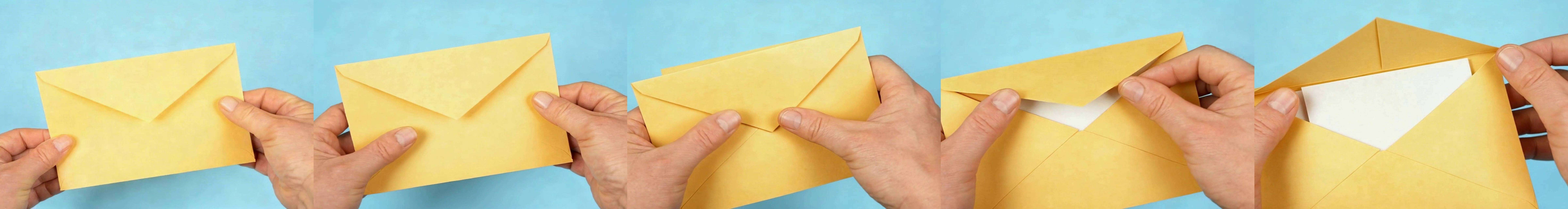}
       \caption{Seedance 1.5pro}
       \vspace{15pt}
    \end{subfigure}

    \begin{subfigure}{0.99\textwidth}
        \centering  
       \includegraphics[width=0.85\linewidth]{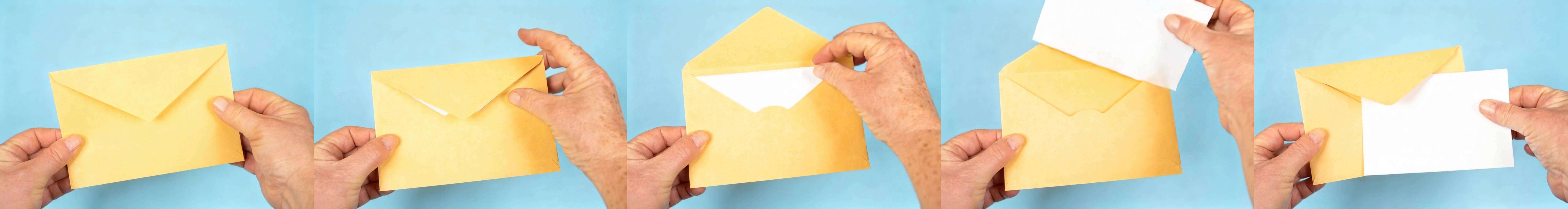}
       \caption{Kling 2.6}
       \vspace{15pt}
    \end{subfigure}
\end{figure}

\begin{figure}\ContinuedFloat

    \begin{subfigure}{0.99\textwidth}
        \centering  
       \includegraphics[width=0.85\linewidth]{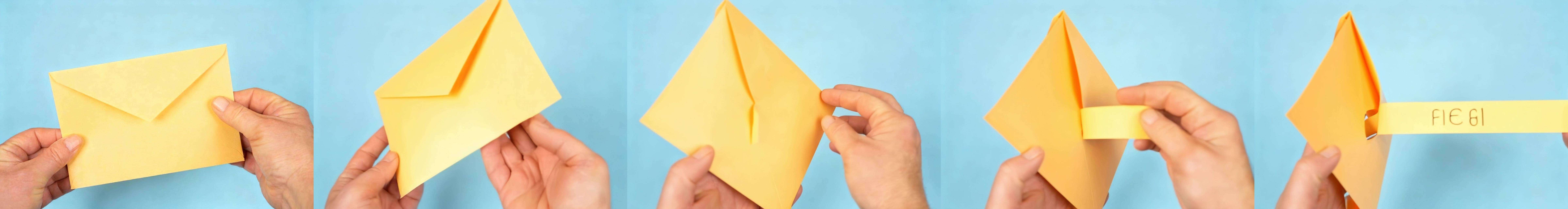}
       \caption{Wan 2.2-I2V-A14B}
       \vspace{15pt}
    \end{subfigure}

    \begin{subfigure}{0.99\textwidth}
        \centering  
       \includegraphics[width=0.85\linewidth]{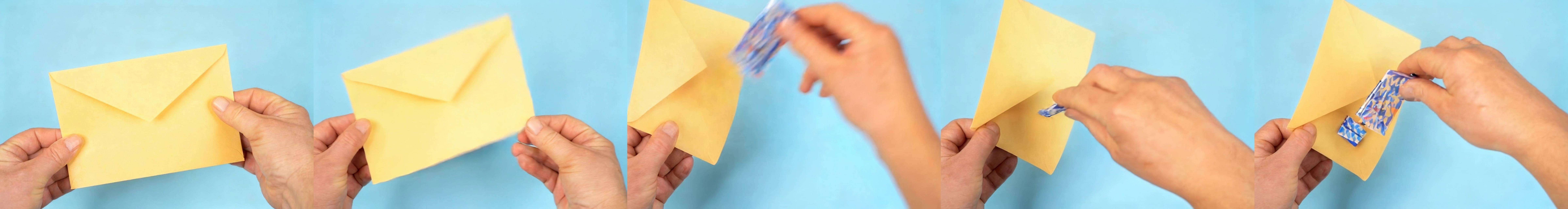}
       \caption{Wan 2.2-TI2V-5B}
       \vspace{15pt}
    \end{subfigure}

    \begin{subfigure}{0.99\textwidth}
        \centering  
       \includegraphics[width=0.85\linewidth]{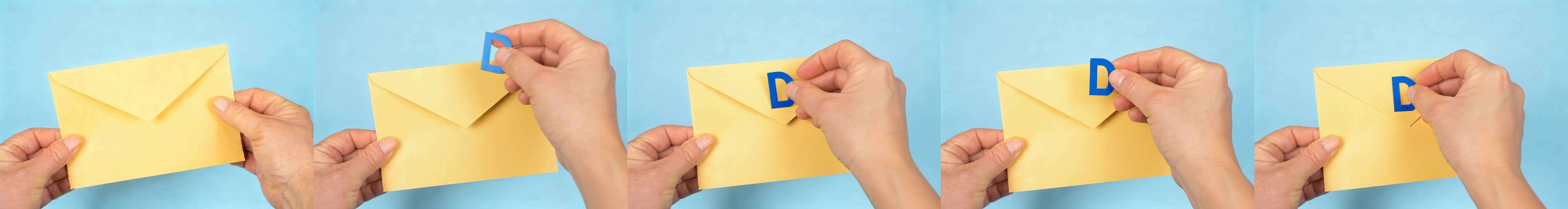}
       \caption{HunyuanVideo-1.5-720P-I2V}
       \vspace{15pt}
    \end{subfigure}

    \begin{subfigure}{0.99\textwidth}
        \centering  
       \includegraphics[width=0.85\linewidth]{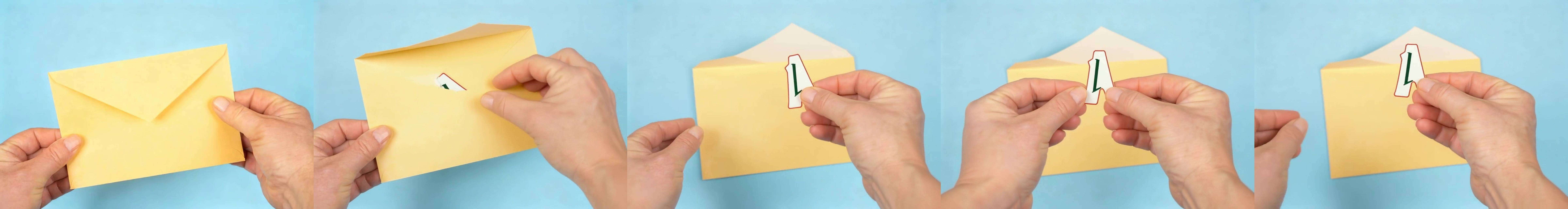}
       \caption{HunyuanVideo-1.5-720P-I2V-cfg-distill }
       \vspace{15pt}
    \end{subfigure}

    \begin{subfigure}{0.99\textwidth}
        \centering  
       \includegraphics[width=0.85\linewidth]{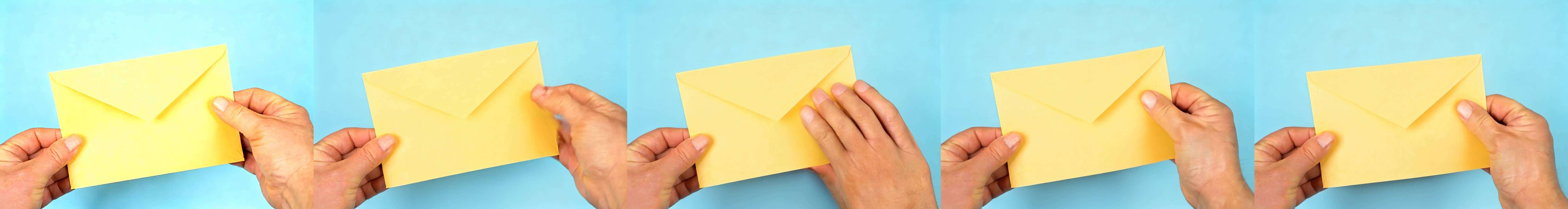}
       \caption{CogVideoX1.5-5B}
       \vspace{15pt}
    \end{subfigure}
    \label{fig:case_appendix_exp}
    \vspace{-10pt}
    \caption{Qualitative examples of generation results from different TI2V models on \textbf{Experiential Knowledge} tasks.}
\end{figure}

\begin{figure}[b]
    \centering  
    \begin{subfigure}{0.99\textwidth}
        \centering  
       \includegraphics[width=0.85\linewidth]{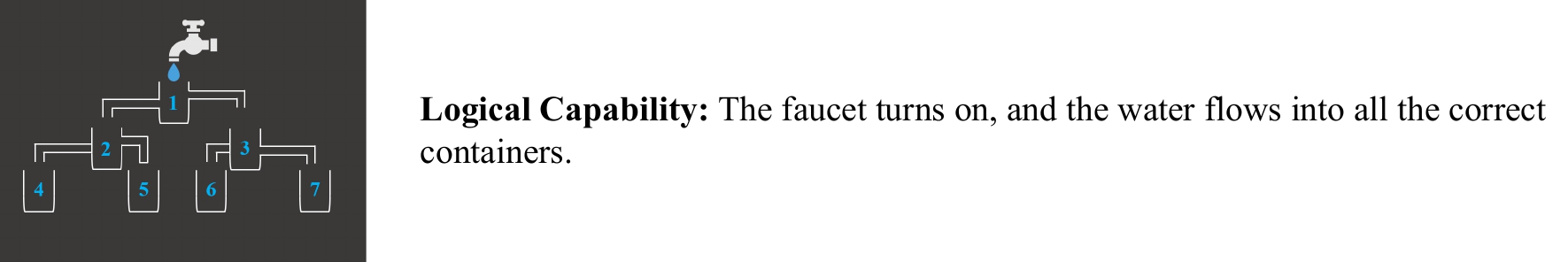}
       \caption{Frame0 and Instruction of TI2V.}
       \vspace{10pt}
    \end{subfigure}
    
    \begin{subfigure}{0.99\textwidth}
        \centering  
       \includegraphics[width=0.85\linewidth]{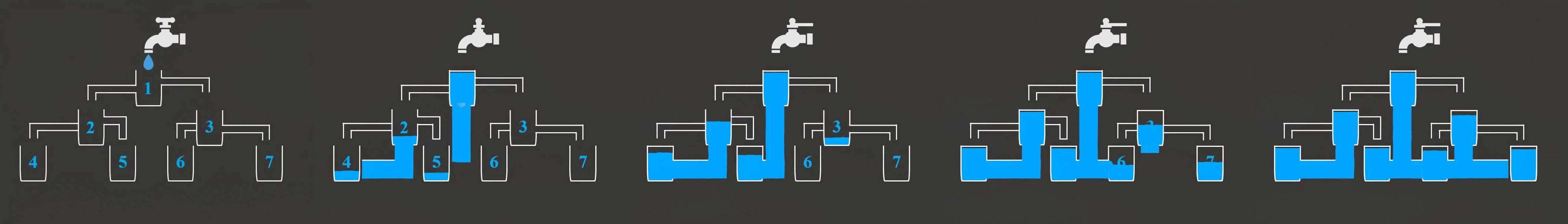}
       \caption{Hailuo 2.3}
       \vspace{10pt}
    \end{subfigure}
    
    \begin{subfigure}{0.99\textwidth}
        \centering  
       \includegraphics[width=0.85\linewidth]{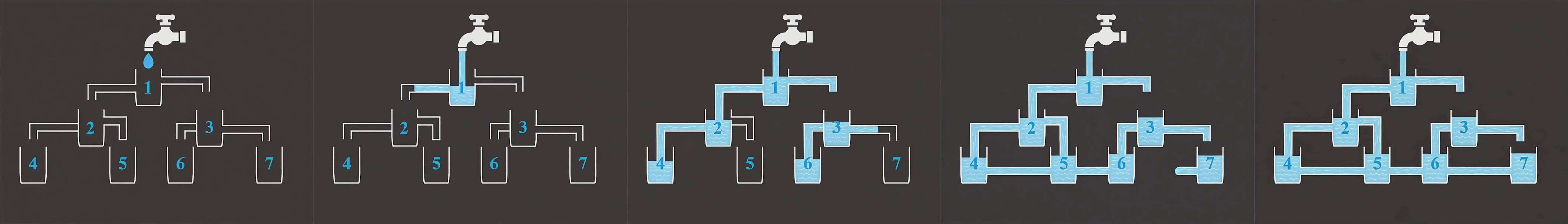}
       \caption{Wan 2.6}
       \vspace{10pt}
    \end{subfigure}
    
    \begin{subfigure}{0.99\textwidth}
        \centering  
       \includegraphics[width=0.85\linewidth]{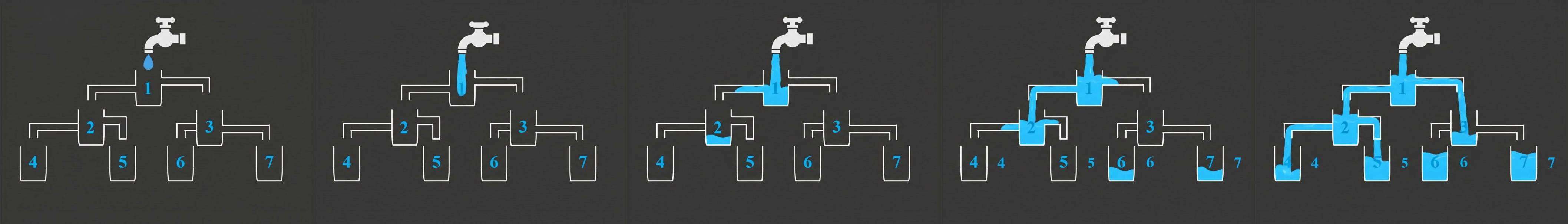}
       \caption{Veo 3.1}
       \vspace{10pt}
    \end{subfigure}

    \label{fig:case_appendix_log1}
    
   \begin{subfigure}{0.99\textwidth}
        \centering  
       \includegraphics[width=0.85\linewidth]{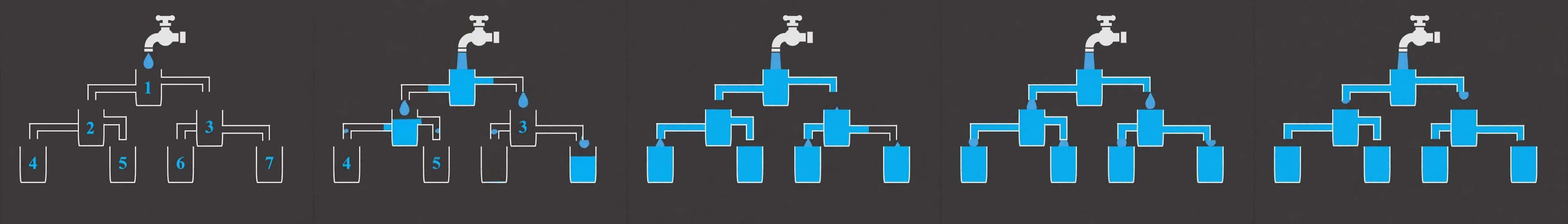}
       \caption{Sora 2}
       \vspace{10pt}
    \end{subfigure}
   \begin{subfigure}{0.99\textwidth}
        \centering  
       \includegraphics[width=0.85\linewidth]{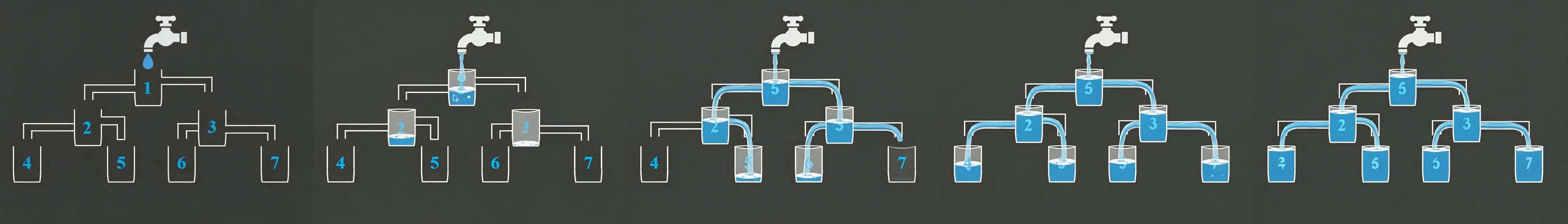}
       \caption{Seedance 1.5pro}
       \vspace{15pt}
    \end{subfigure}

    \begin{subfigure}{0.99\textwidth}
        \centering  
       \includegraphics[width=0.85\linewidth]{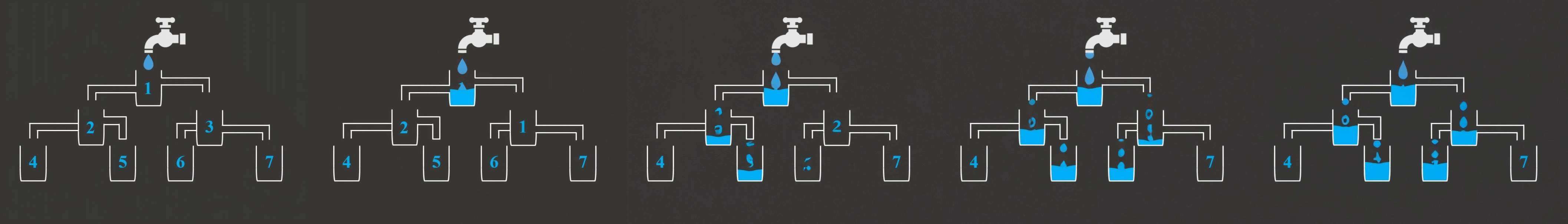}
       \caption{Kling 2.6}
       \vspace{15pt}
    \end{subfigure}
\end{figure}

\begin{figure}\ContinuedFloat

    \begin{subfigure}{0.99\textwidth}
        \centering  
       \includegraphics[width=0.85\linewidth]{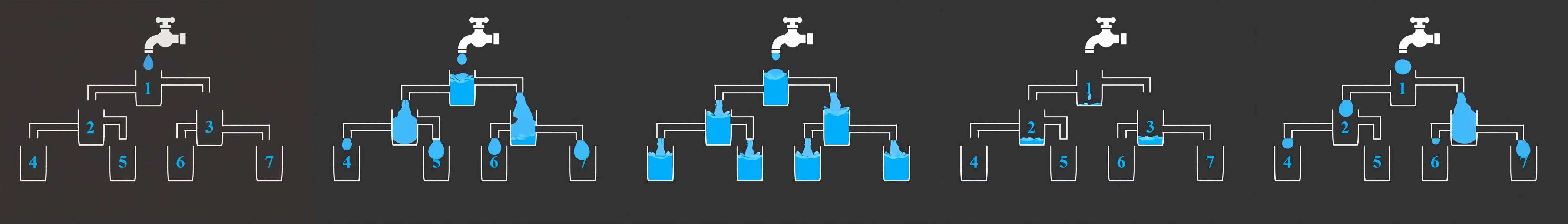}
       \caption{Wan 2.2-I2V-A14B}
       \vspace{15pt}
    \end{subfigure}

    \begin{subfigure}{0.99\textwidth}
        \centering  
       \includegraphics[width=0.85\linewidth]{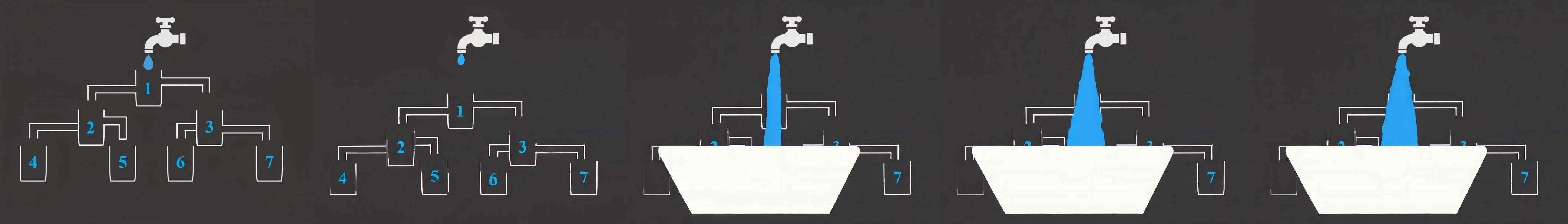}
       \caption{Wan 2.2-TI2V-5B}
       \vspace{15pt}
    \end{subfigure}

    \begin{subfigure}{0.99\textwidth}
        \centering  
       \includegraphics[width=0.85\linewidth]{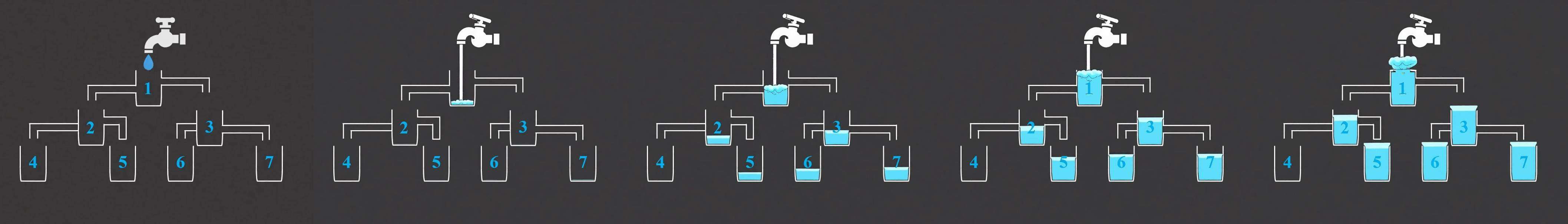}
       \caption{HunyuanVideo-1.5-720P-I2V}
       \vspace{15pt}
    \end{subfigure}

    \begin{subfigure}{0.99\textwidth}
        \centering  
       \includegraphics[width=0.85\linewidth]{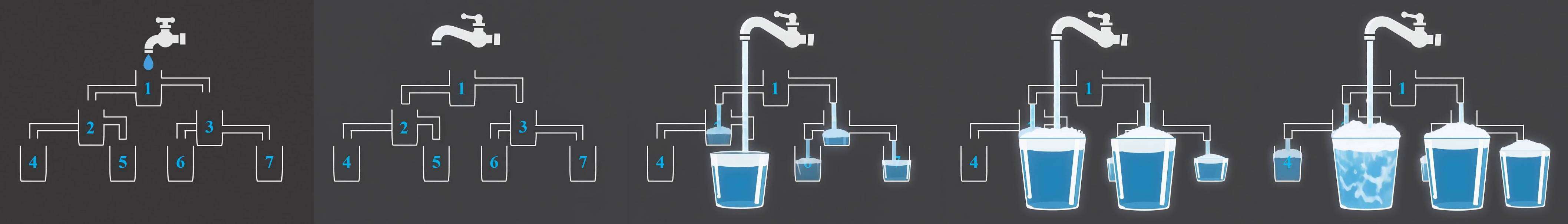}
       \caption{HunyuanVideo-1.5-720P-I2V-cfg-distill }
       \vspace{15pt}
    \end{subfigure}

    \begin{subfigure}{0.99\textwidth}
        \centering  
       \includegraphics[width=0.85\linewidth]{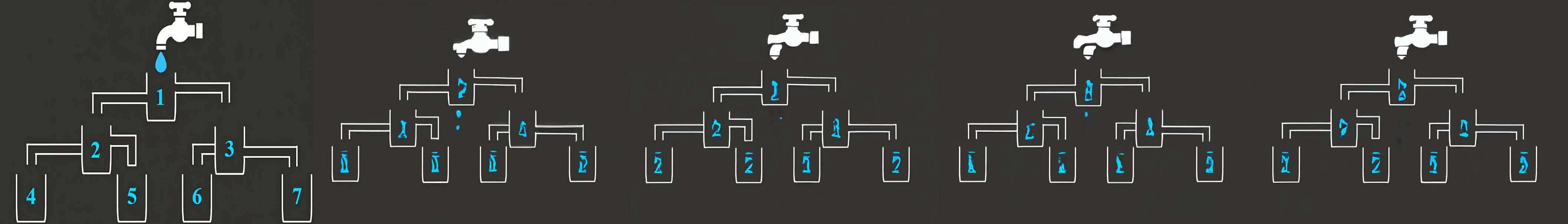}
       \caption{CogVideoX1.5-5B}
       \vspace{15pt}
    \end{subfigure}
    \label{fig:case_appendix_log}
    \vspace{-10pt}
    \caption{Qualitative examples of generation results from different TI2V models on \textbf{Logical Capability} tasks.}
\end{figure}

\begin{figure}[b]
    \centering  
    \begin{subfigure}{0.99\textwidth}
        \centering  
       \includegraphics[width=0.85\linewidth]{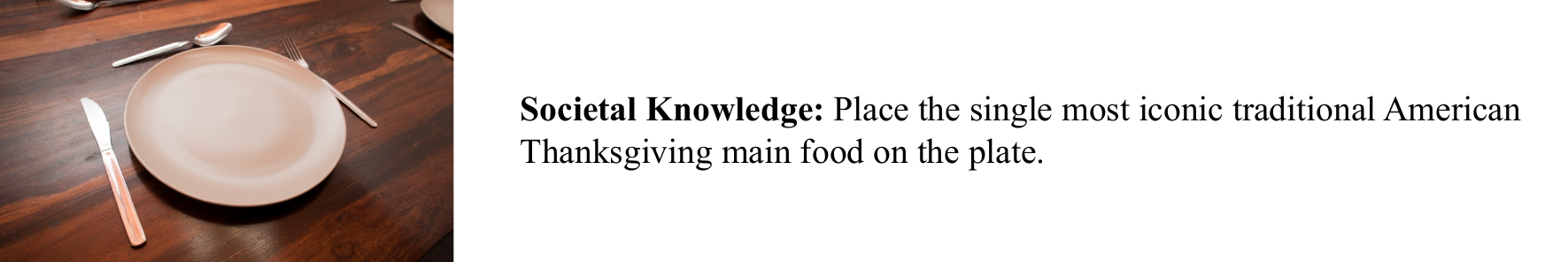}
       \caption{Frame0 and Instruction of TI2V.}
       \vspace{10pt}
    \end{subfigure}
    
    \begin{subfigure}{0.99\textwidth}
        \centering  
       \includegraphics[width=0.85\linewidth]{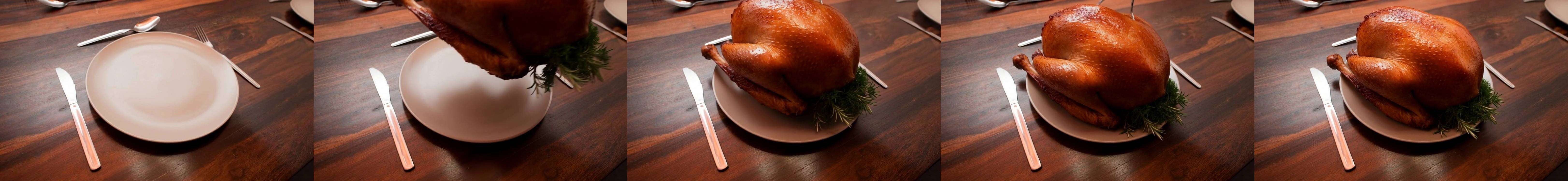}
       \caption{Hailuo 2.3}
       \vspace{10pt}
    \end{subfigure}
    
    \begin{subfigure}{0.99\textwidth}
        \centering  
       \includegraphics[width=0.85\linewidth]{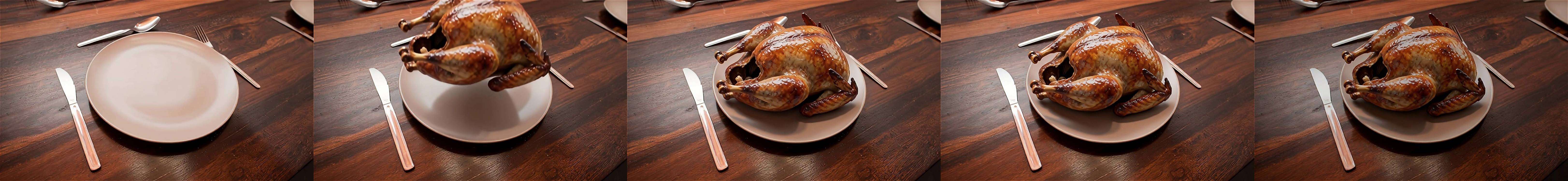}
       \caption{Wan 2.6}
       \vspace{10pt}
    \end{subfigure}
    
    \begin{subfigure}{0.99\textwidth}
        \centering  
       \includegraphics[width=0.85\linewidth]{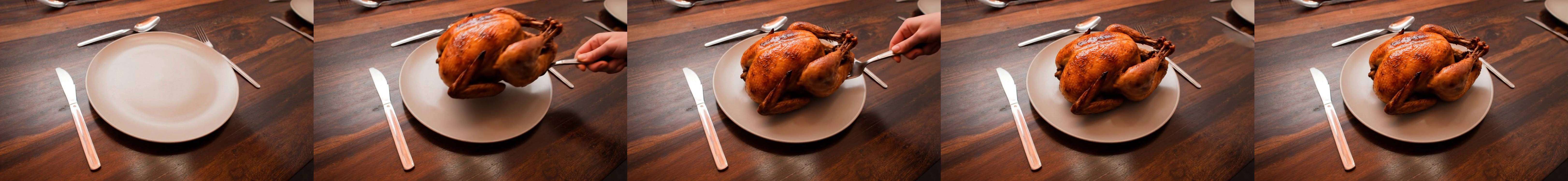}
       \caption{Veo 3.1}
       \vspace{10pt}
    \end{subfigure}

    \label{fig:case_appendix_soc1}
    
   \begin{subfigure}{0.99\textwidth}
        \centering  
       \includegraphics[width=0.85\linewidth]{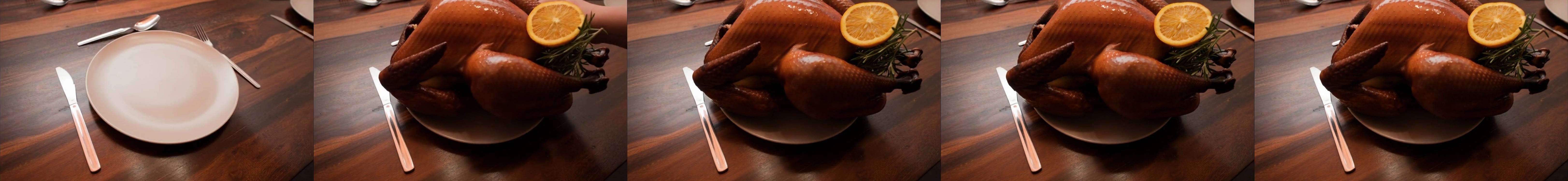}
       \caption{Sora 2}
       \vspace{10pt}
    \end{subfigure}
   \begin{subfigure}{0.99\textwidth}
        \centering  
       \includegraphics[width=0.85\linewidth]{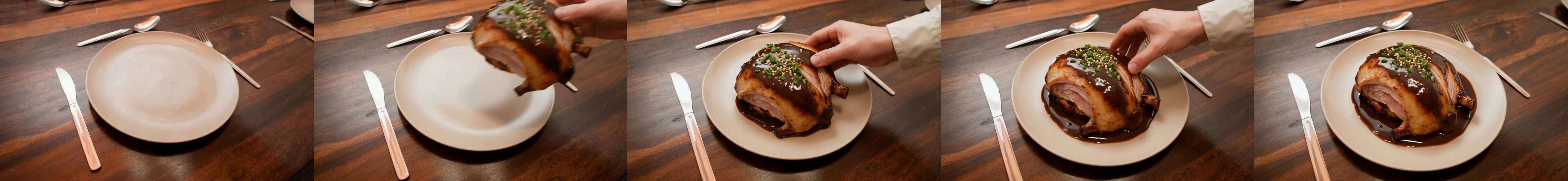}
       \caption{Seedance 1.5pro}
       \vspace{15pt}
    \end{subfigure}

    \begin{subfigure}{0.99\textwidth}
        \centering  
       \includegraphics[width=0.85\linewidth]{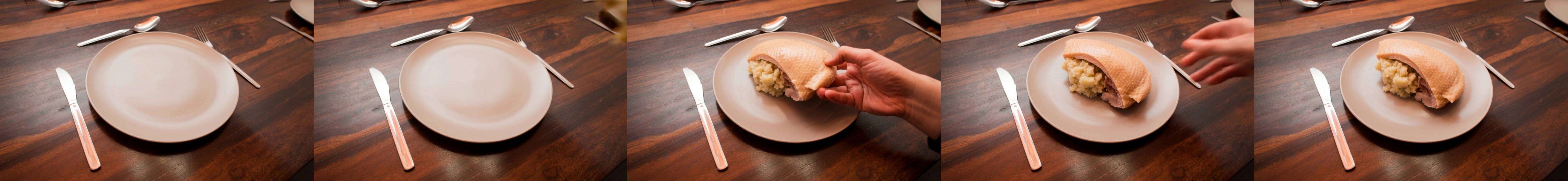}
       \caption{Kling 2.6}
       \vspace{15pt}
    \end{subfigure}
\end{figure}

\begin{figure}\ContinuedFloat

    \begin{subfigure}{0.99\textwidth}
        \centering  
       \includegraphics[width=0.85\linewidth]{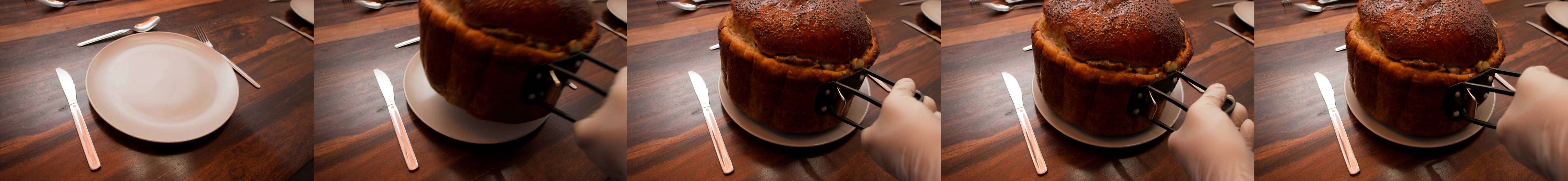}
       \caption{Wan 2.2-I2V-A14B}
       \vspace{15pt}
    \end{subfigure}

    \begin{subfigure}{0.99\textwidth}
        \centering  
       \includegraphics[width=0.85\linewidth]{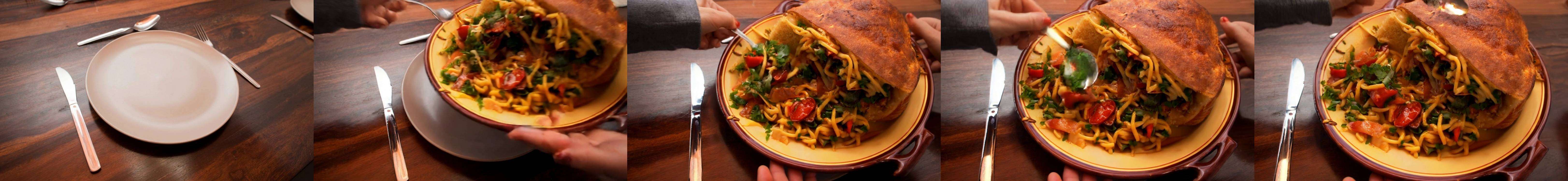}
       \caption{Wan 2.2-TI2V-5B}
       \vspace{15pt}
    \end{subfigure}

    \begin{subfigure}{0.99\textwidth}
        \centering  
       \includegraphics[width=0.85\linewidth]{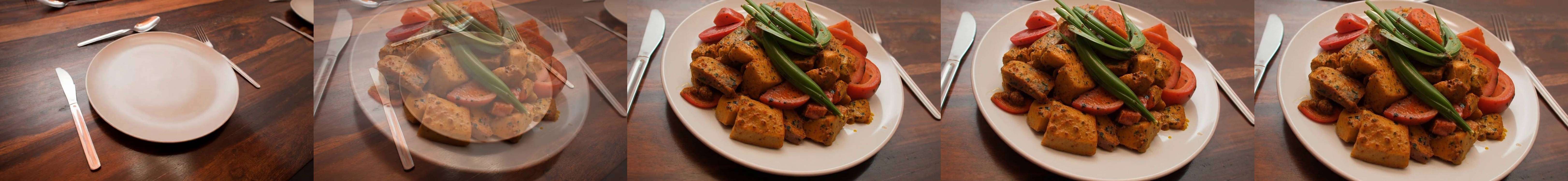}
       \caption{HunyuanVideo-1.5-720P-I2V}
       \vspace{15pt}
    \end{subfigure}

    \begin{subfigure}{0.99\textwidth}
        \centering  
       \includegraphics[width=0.85\linewidth]{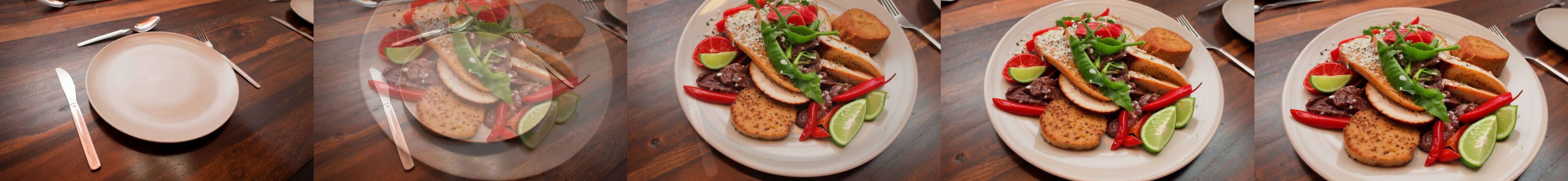}
       \caption{HunyuanVideo-1.5-720P-I2V-cfg-distill }
       \vspace{15pt}
    \end{subfigure}

    \begin{subfigure}{0.99\textwidth}
        \centering  
       \includegraphics[width=0.85\linewidth]{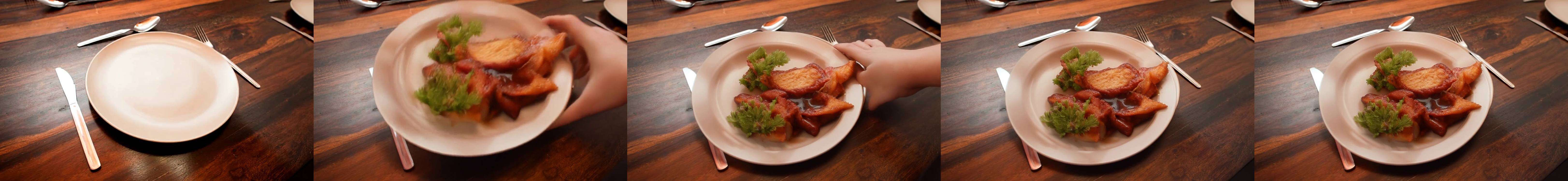}
       \caption{CogVideoX1.5-5B}
       \vspace{15pt}
    \end{subfigure}
    \label{fig:case_appendix_soc}
    \vspace{-10pt}
    \caption{Qualitative examples of generation results from different TI2V models on \textbf{Societal Knowledge} tasks.}
\end{figure}

\begin{figure}[b]
    \centering  
    \begin{subfigure}{0.99\textwidth}
        \centering  
       \includegraphics[width=0.85\linewidth]{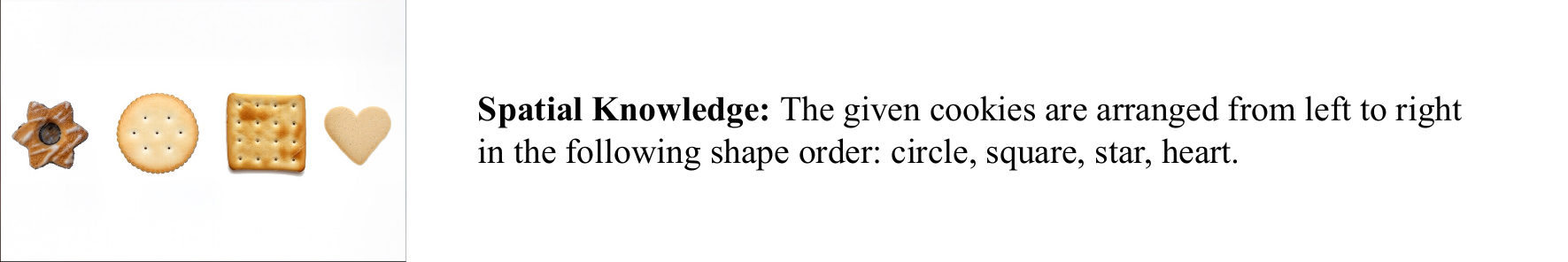}
       \caption{Frame0 and Instruction of TI2V.}
       \vspace{10pt}
    \end{subfigure}
    
    \begin{subfigure}{0.99\textwidth}
        \centering  
       \includegraphics[width=0.85\linewidth]{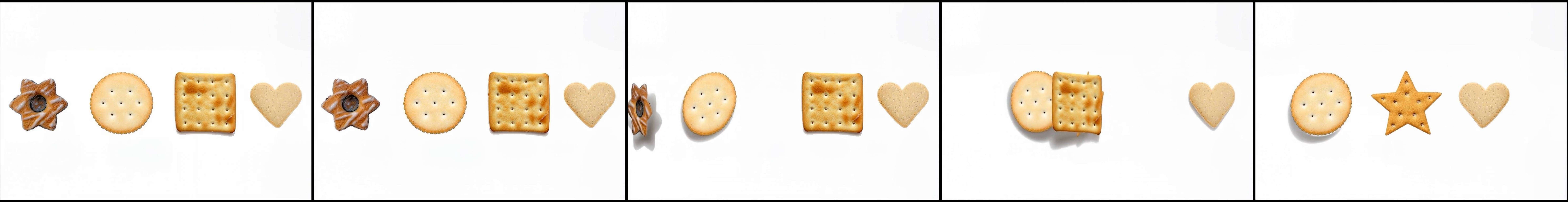}
       \caption{Hailuo 2.3}
       \vspace{10pt}
    \end{subfigure}
    
    \begin{subfigure}{0.99\textwidth}
        \centering  
       \includegraphics[width=0.85\linewidth]{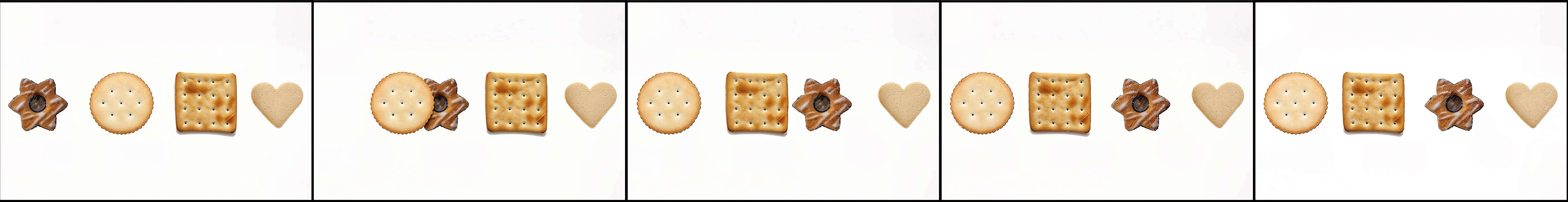}
       \caption{Wan 2.6}
       \vspace{10pt}
    \end{subfigure}
    
    \begin{subfigure}{0.99\textwidth}
        \centering  
       \includegraphics[width=0.85\linewidth]{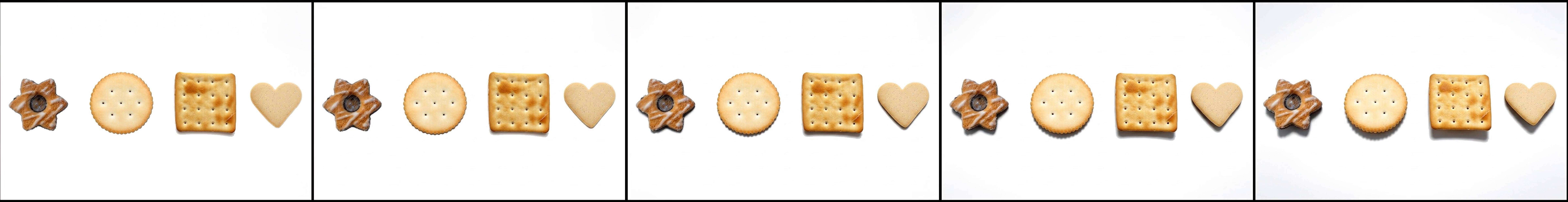}
       \caption{Veo 3.1}
       \vspace{10pt}
    \end{subfigure}

    \label{fig:case_appendix_spa1}
    
   \begin{subfigure}{0.99\textwidth}
        \centering  
       \includegraphics[width=0.85\linewidth]{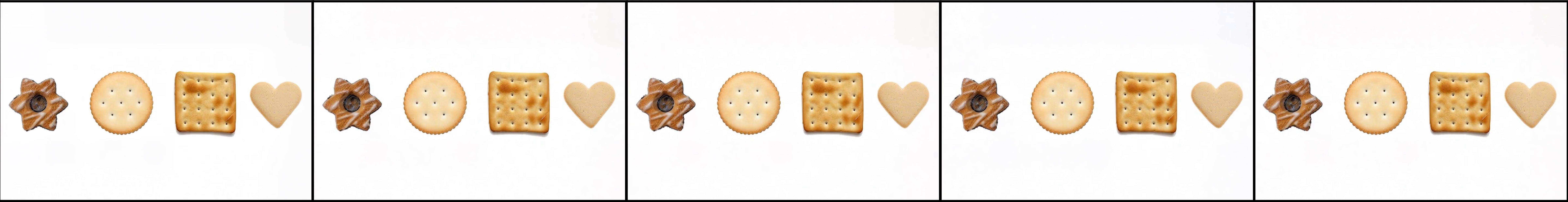}
       \caption{Sora 2}
       \vspace{10pt}
    \end{subfigure}
   \begin{subfigure}{0.99\textwidth}
        \centering  
       \includegraphics[width=0.85\linewidth]{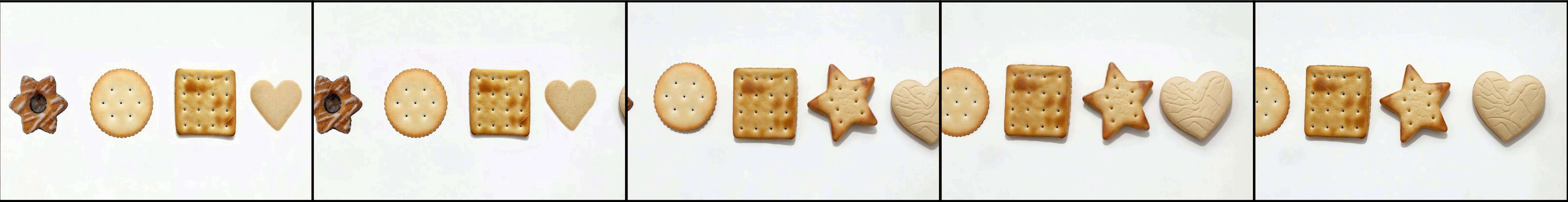}
       \caption{Seedance 1.5pro}
       \vspace{15pt}
    \end{subfigure}

    \begin{subfigure}{0.99\textwidth}
        \centering  
       \includegraphics[width=0.85\linewidth]{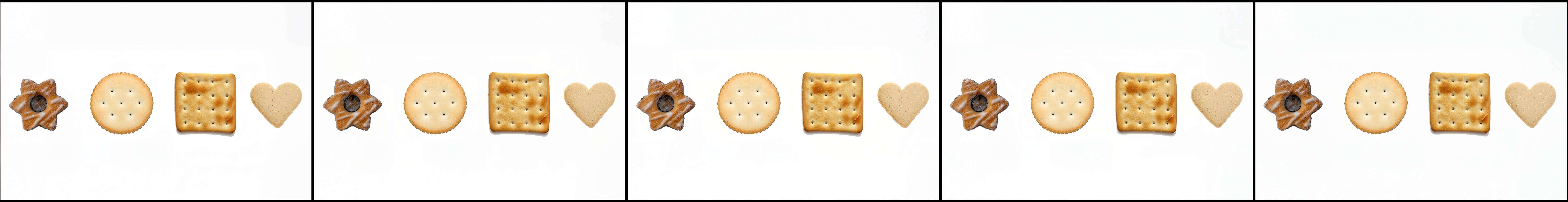}
       \caption{Kling 2.6}
       \vspace{15pt}
    \end{subfigure}
\end{figure}

\begin{figure}\ContinuedFloat

    \begin{subfigure}{0.99\textwidth}
        \centering  
       \includegraphics[width=0.85\linewidth]{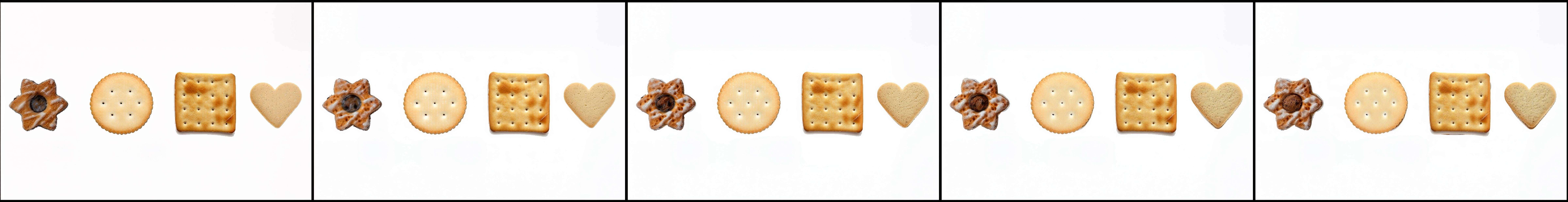}
       \caption{Wan 2.2-I2V-A14B}
       \vspace{15pt}
    \end{subfigure}

    \begin{subfigure}{0.99\textwidth}
        \centering  
       \includegraphics[width=0.85\linewidth]{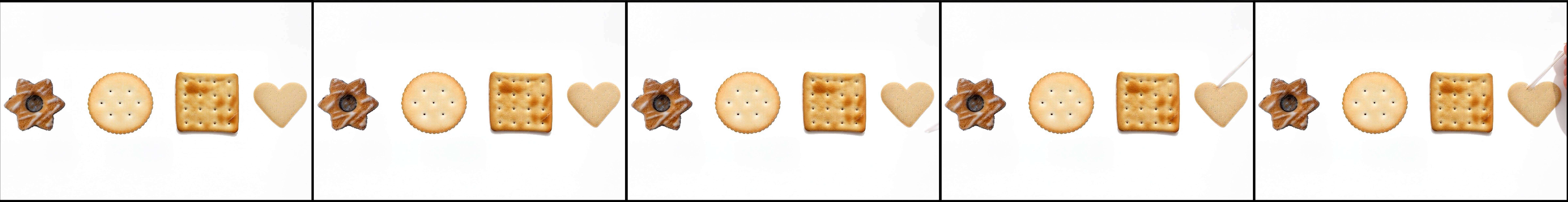}
       \caption{Wan 2.2-TI2V-5B}
       \vspace{15pt}
    \end{subfigure}

    \begin{subfigure}{0.99\textwidth}
        \centering  
       \includegraphics[width=0.85\linewidth]{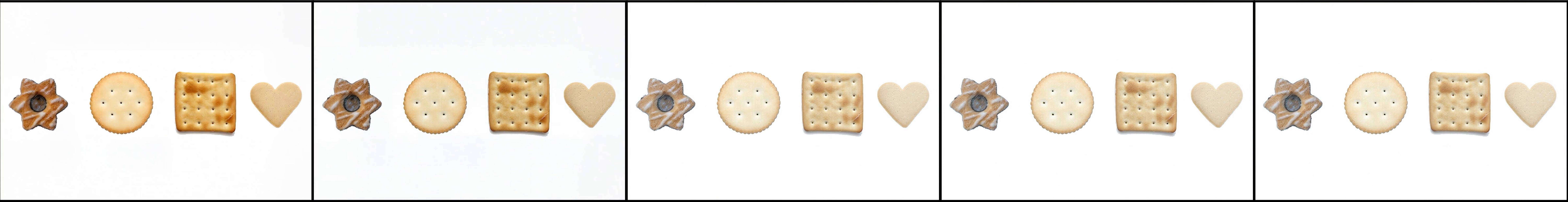}
       \caption{HunyuanVideo-1.5-720P-I2V}
       \vspace{15pt}
    \end{subfigure}

    \begin{subfigure}{0.99\textwidth}
        \centering  
       \includegraphics[width=0.85\linewidth]{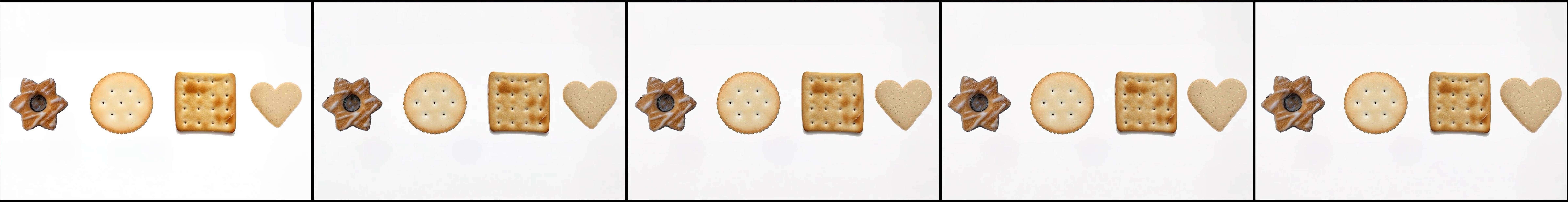}
       \caption{HunyuanVideo-1.5-720P-I2V-cfg-distill }
       \vspace{15pt}
    \end{subfigure}

    \begin{subfigure}{0.99\textwidth}
        \centering  
       \includegraphics[width=0.85\linewidth]{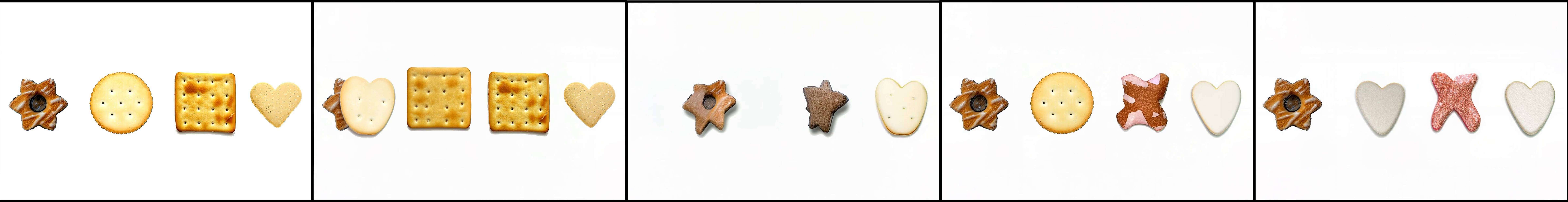}
       \caption{CogVideoX1.5-5B}
       \vspace{15pt}
    \end{subfigure}
    \label{fig:case_appendix_spa}
    \vspace{-10pt}
    \caption{Qualitative examples of generation results from different TI2V models on \textbf{Spatial Knowledge} tasks.}
\end{figure}

\begin{figure}[b]
    \centering  
    \begin{subfigure}{0.99\textwidth}
        \centering  
       \includegraphics[width=0.85\linewidth]{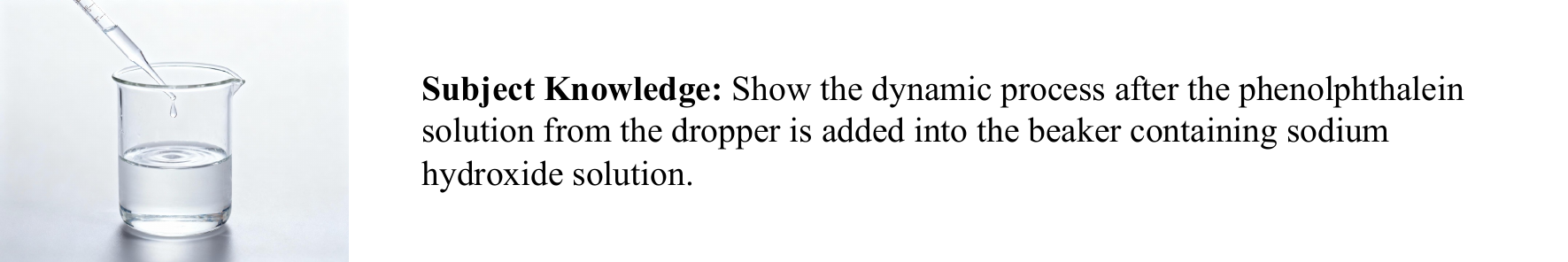}
       \caption{Frame0 and Instruction of TI2V.}
       \vspace{10pt}
    \end{subfigure}
    
    \begin{subfigure}{0.99\textwidth}
        \centering  
       \includegraphics[width=0.85\linewidth]{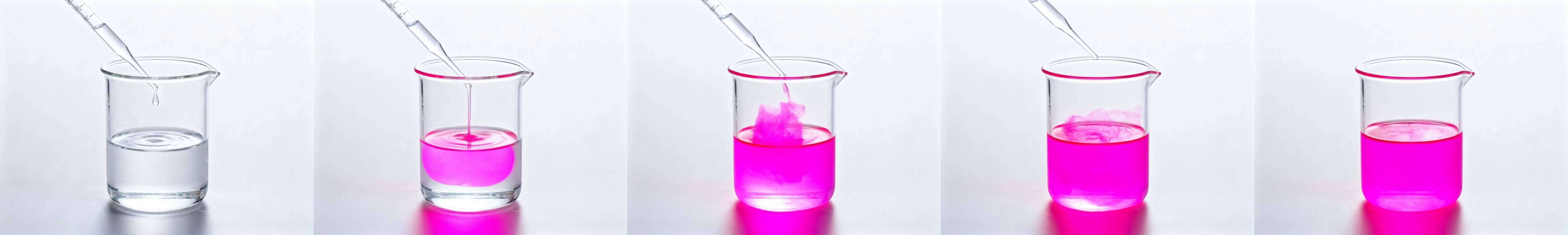}
       \caption{Hailuo 2.3}
       \vspace{10pt}
    \end{subfigure}
    
    \begin{subfigure}{0.99\textwidth}
        \centering  
       \includegraphics[width=0.85\linewidth]{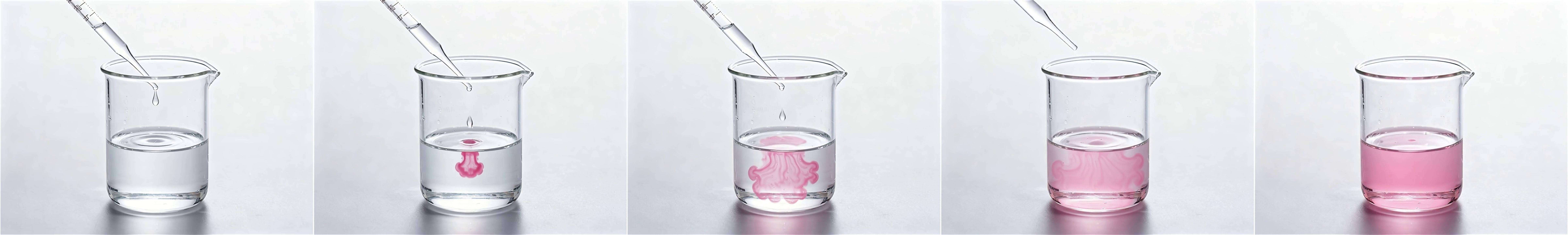}
       \caption{Wan 2.6}
       \vspace{10pt}
    \end{subfigure}
    
    \begin{subfigure}{0.99\textwidth}
        \centering  
       \includegraphics[width=0.85\linewidth]{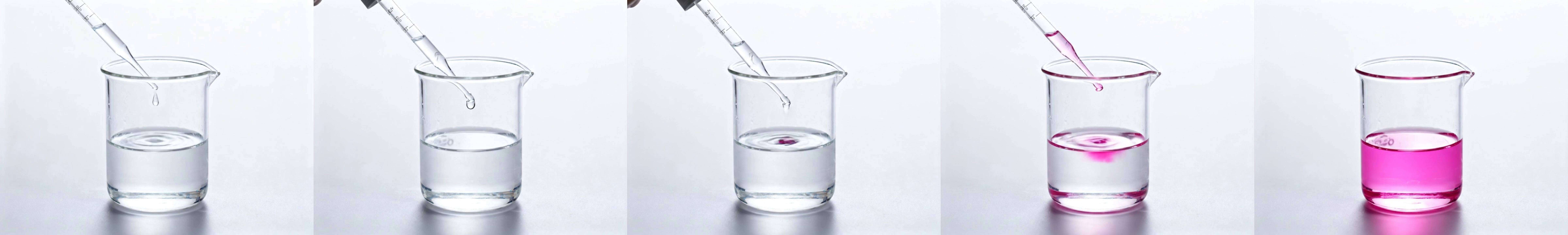}
       \caption{Veo 3.1}
       \vspace{10pt}
    \end{subfigure}

    \label{fig:case_appendix_dom1}
    
   \begin{subfigure}{0.99\textwidth}
        \centering  
       \includegraphics[width=0.85\linewidth]{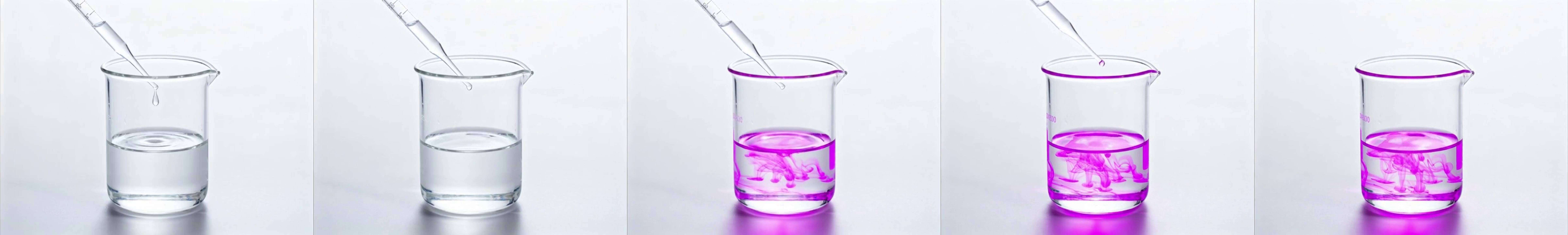}
       \caption{Sora 2}
       \vspace{10pt}
    \end{subfigure}
   \begin{subfigure}{0.99\textwidth}
        \centering  
       \includegraphics[width=0.85\linewidth]{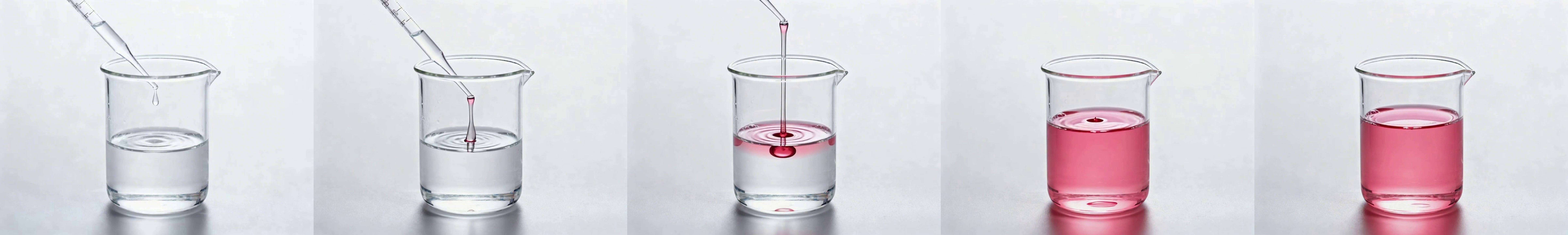}
       \caption{Seedance 1.5pro}
       \vspace{15pt}
    \end{subfigure}

    \begin{subfigure}{0.99\textwidth}
        \centering  
       \includegraphics[width=0.85\linewidth]{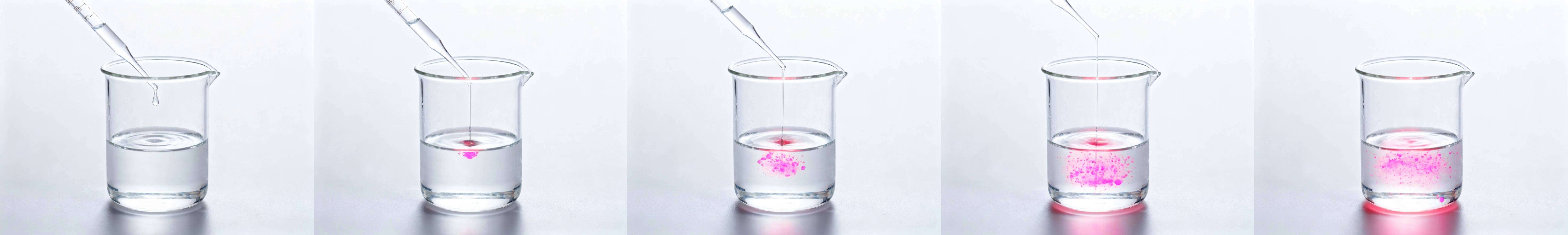}
       \caption{Kling 2.6}
       \vspace{15pt}
    \end{subfigure}
\end{figure}

\begin{figure}\ContinuedFloat

    \begin{subfigure}{0.99\textwidth}
        \centering  
       \includegraphics[width=0.85\linewidth]{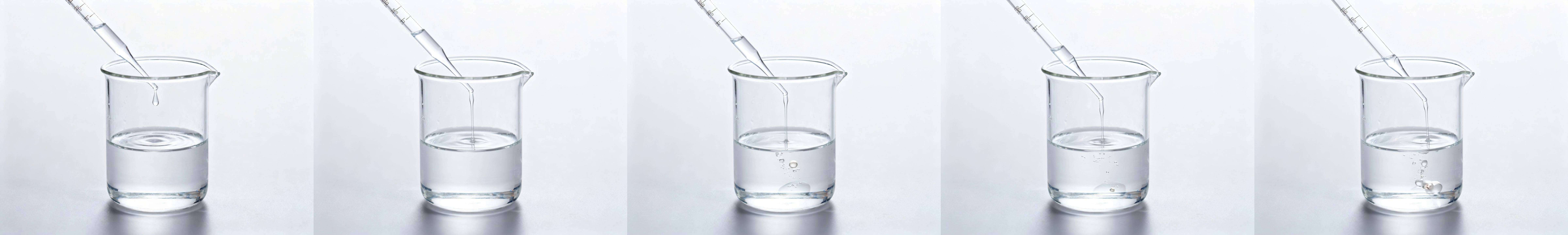}
       \caption{Wan 2.2-I2V-A14B}
       \vspace{15pt}
    \end{subfigure}

    \begin{subfigure}{0.99\textwidth}
        \centering  
       \includegraphics[width=0.85\linewidth]{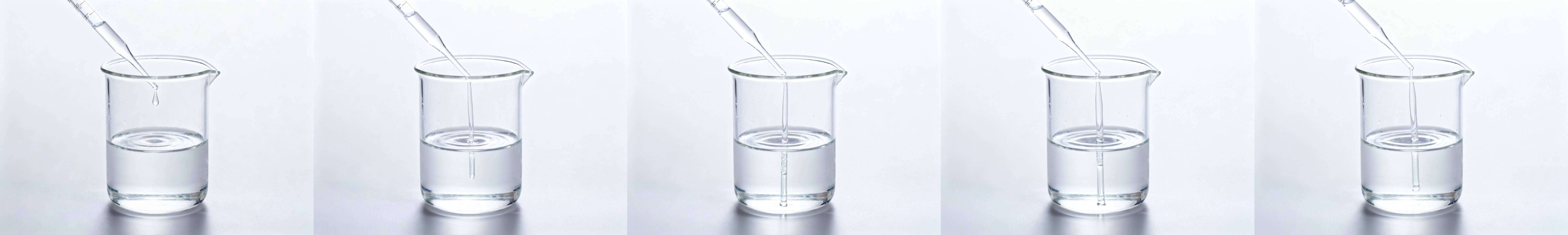}
       \caption{Wan 2.2-TI2V-5B}
       \vspace{15pt}
    \end{subfigure}

    \begin{subfigure}{0.99\textwidth}
        \centering  
       \includegraphics[width=0.85\linewidth]{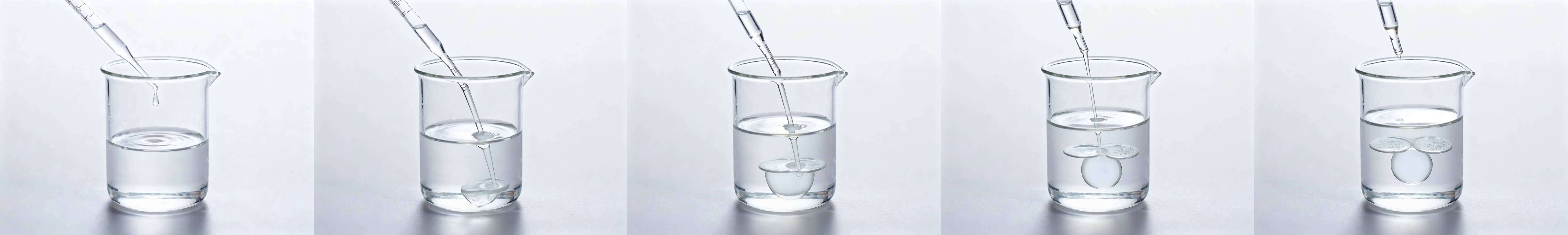}
       \caption{HunyuanVideo-1.5-720P-I2V}
       \vspace{15pt}
    \end{subfigure}

    \begin{subfigure}{0.99\textwidth}
        \centering  
       \includegraphics[width=0.85\linewidth]{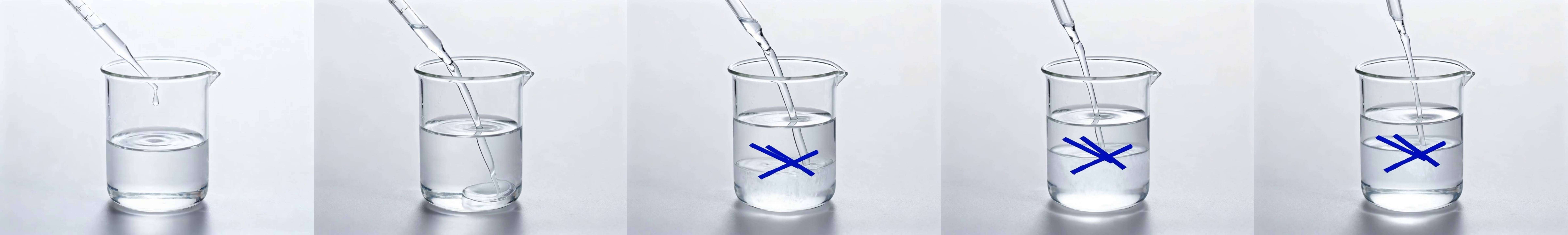}
       \caption{HunyuanVideo-1.5-720P-I2V-cfg-distill }
       \vspace{15pt}
    \end{subfigure}

    \begin{subfigure}{0.99\textwidth}
        \centering  
       \includegraphics[width=0.85\linewidth]{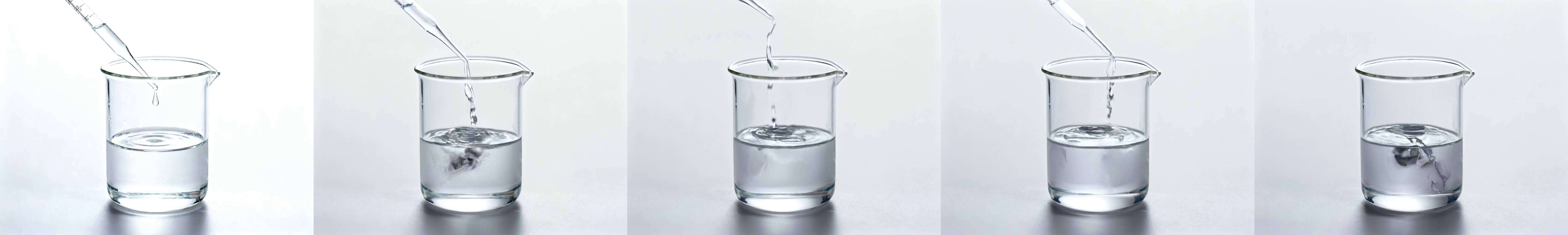}
       \caption{CogVideoX1.5-5B}
       \vspace{15pt}
    \end{subfigure}
    \label{fig:case_appendix_dom}
    \vspace{-10pt}
    \caption{Qualitative examples of generation results from different TI2V models on \textbf{Subject Knowledge} tasks.}
\end{figure}